\documentclass[conference,compsoc]{IEEEtran}
\ifCLASSOPTIONcompsoc
  \usepackage[nocompress]{cite}
\else
  \usepackage{cite}
\fi
\ifCLASSINFOpdf
\else
\fi

\usepackage{pdflscape} % to rotate the page to landscape
\usepackage{adjustbox} % for resizing

\usepackage{amsmath}
\usepackage{pifont}% http://ctan.org/pkg/pifont
\usepackage{todonotes}
\usepackage[htt]{hyphenat}
\usepackage[aboveskip=-2pt]{subcaption}
\usepackage{graphics}
\usepackage{graphicx}
\usepackage{url}

\usepackage{breakurl}
\usepackage{multirow}
\usepackage{booktabs}
\usepackage[utf8]{inputenc}
\UseRawInputEncoding
\usepackage{float}

\usepackage{tikz}
\usepackage{amsmath}
\usepackage{textcomp}
\usepackage{bmpsize}
\usepackage{xcolor}
\usepackage{lipsum}
\usepackage[ruled,linesnumbered]{algorithm2e}
\usepackage{dblfloatfix}
\usepackage[english]{babel}
\usepackage{multirow}
\usepackage{amssymb}% http://ctan.org/pkg/amssymb
\usepackage{pifont}% http://ctan.org/pkg/pifont
\usepackage{amsmath,array,graphicx}
\usepackage{todonotes}
\usepackage{float}
\usepackage{booktabs}
\newcommand{\cmmnt}[1]{}
\usepackage[most]{tcolorbox}
\usepackage{cite}
\usepackage[font=small,labelfont=bf]{caption}
\usepackage{titlesec}

\usepackage[utf8]{inputenc}
\usepackage[T1]{fontenc}

\usepackage{multirow}
\usepackage{booktabs}
\usepackage{amssymb} % for checkmark symbols
\usepackage{pifont} % for additional symbols

\usepackage{algorithmic}
\usepackage{graphicx}
\usepackage{textcomp}
\usepackage{xcolor}
\usepackage{booktabs}
\usepackage[most]{tcolorbox}
\usepackage{mathrsfs}
\usepackage[ruled,linesnumbered]{algorithm2e}
\usepackage{amsmath}
\usepackage{hyperref}

\usepackage{pdflscape}
\usepackage{afterpage}  % To apply commands after a page break
\usepackage{graphicx} % For \rotatebox

\usepackage{multirow}
\usepackage[aboveskip=-2pt]{subcaption}
\usepackage{makecell} % Add in the preamble

\usepackage{etoolbox}  % for patching commands

\newcommand{\appsection}[2][]{%
  \refstepcounter{section}%
  \section*{Appendix \Alph{section}. #2}%
  \phantomsection%
  \addcontentsline{toc}{section}{Appendix \Alph{section}. #2}%
  \ifstrempty{#1}{}{\label{#1}}%
  \markboth{Appendix \Alph{section}. #2}{}%
}

\begin{document}
\title{On the Effectiveness of Adversarial Training on Malware Classifiers}
% \title{\vm{\texttt{Onion:} Re-thinking/Assessing the Complexity Layers of Adversarial Training for Malware Classifiers }}

% \author{\IEEEauthorblockN{Hamid Bostani}
% \IEEEauthorblockA{University of Luxembourg\\
% Luxembourg}
% \and
% \IEEEauthorblockN{Jacopo Cortellazzi}
% \IEEEauthorblockA{King's College London\\
% United Kingdom}
% \and
% \IEEEauthorblockN{Daniel Arp}
% \IEEEauthorblockA{TU Wien\\
% Austria}
% \and
% \IEEEauthorblockN{Fabio Pierazzi}
% \IEEEauthorblockA{University College London\\
% United Kingdom}
% \and 
% \IEEEauthorblockN{Veelasha Moonsamy}
% \IEEEauthorblockA{Ruhr University Bochum\\
% Germany}
% \and
% \IEEEauthorblockN{Lorenzo Cavallaro}
% \IEEEauthorblockA{University College London\\
% United Kingdom}}

\author{%
\begin{tabular}{@{}c@{\hspace{1.5em}}c@{\hspace{1.5em}}c@{}}
% ---- Row 1 (three independent blocks) ----
\begin{minipage}[t]{0.30\linewidth}\centering
Hamid Bostani$^{*}$\thanks{First author’s footnote text goes here.}\\
University of Luxembourg\\
Luxembourg
\end{minipage} &
\begin{minipage}[t]{0.30\linewidth}\centering
Jacopo Cortellazzi\\
King's College London\\
United Kingdom
\end{minipage} &
\begin{minipage}[t]{0.30\linewidth}\centering
Daniel Arp\\
TU Wien\\
Austria
\end{minipage} \\[3.5em]
% ---- Row 2 (three independent blocks) ----
\begin{minipage}[t]{0.30\linewidth}\centering
Fabio Pierazzi\\
University College London\\
United Kingdom
\end{minipage} &
\begin{minipage}[t]{0.30\linewidth}\centering
Veelasha Moonsamy\\
Ruhr University Bochum\\
Germany
\end{minipage} &
\begin{minipage}[t]{0.30\linewidth}\centering
Lorenzo Cavallaro\\
University College London\\
United Kingdom
\end{minipage}
\end{tabular}%
} % end \author


% compsoc conferences compared to traditional conferences), use this
%Telephone: (800) 555--1212, Fax: (888) 555--1212}


\maketitle

\makeatletter
\renewcommand{\@makefntext}[1]{#1} % remove dot and special formatting
\makeatother

\footnotetext{${}^{*}$ This work was primarily conducted during the author's PhD at Radboud University, Nijmegen, The Netherlands.}
\begin{abstract}
Adversarial Training (AT) is a key defense against Machine Learning evasion attacks, but its effectiveness for real-world malware detection remains poorly understood. This uncertainty stems from a critical disconnect in prior research: studies often overlook the inherent nature of malware and are fragmented, examining diverse variables like realism or confidence of adversarial examples in isolation, or relying on weak evaluations that yield non-generalizable insights.
To address this, we introduce \textit{Rubik}, a framework for the systematic, multi-dimensional evaluation of AT in the malware domain. This framework defines diverse key factors across essential dimensions, including data, feature representations, classifiers, and robust optimization settings, for a comprehensive exploration of the interplay of influential AT's variables through reliable evaluation practices, such as realistic evasion attacks. We instantiate Rubik on Android malware, empirically analyzing how this interplay shapes robustness. Our findings challenge prior beliefs---showing, for instance, that realizable adversarial examples offer only conditional robustness benefits---and reveal new insights, such as the critical role of model architecture and feature-space structure in determining AT's success. From this analysis, we distill four key insights, expose four common evaluation misconceptions, and offer practical recommendations to guide the development of truly robust malware classifiers.
\end{abstract}


% \fi
\IEEEpeerreviewmaketitle

\section{Introduction}
\vspace{-0.4em}
% Adversarial Training (AT) is widely regarded as a cornerstone of robust machine learning, yet its development and evaluation have been almost entirely dominated by the image domain ~\cite{Goodfellow2014Explaining, madry2017towards, carlini2019evaluating, zimmermann2022increasing, zhang2019limitations, wang2018mixtrain}. Applying these image-centric lessons to malware detection creates a fundamental mismatch. 
Adversarial training (AT)~\cite{madry2017towards} is a primary defense for hardening malware classifiers against evasion attacks. However, its standard methodology is largely disconnected from the distinctive challenges of the malware domain.
Malware classifiers operate in discrete, structured spaces, where the alignment between a program's logic and its feature abstraction is not inherent~\cite{li2021arms}. Unlike image classification models~\cite{Goodfellow2014Explaining, madry2017towards, carlini2019evaluating, zimmermann2022increasing, zhang2019limitations, wang2018mixtrain}, which largely rely on deep neural networks, malware detection employs a diverse range of classifiers---from linear models to deep architectures---each may interact with AT in distinct ways. These challenges are further compounded by the critical constraint of \textit{realizability}: \textit{adversarial examples} (AEs) generated by evasion attacks must preserve malware's malicious functionality, remain robust to preprocessing, and be plausible in practice~\cite{Pierazzi2020Intriguing}. Prior efforts to apply AT to malware have been fragmented. They often isolate single variables~\cite{li2023pad,doan2023feature} such as a realizability of AEs~\cite{dyrmishi2023empirical, bostani2022domain, cortellazzi2024intriguingpropertiesadversarialml}, rely on basic adversarial re-training~\cite{dyrmishi2023empirical,wang2021advandmal,chen2021using,wang2020mdea,rathore2021robust,huang2019malware,anderson2018learning,yang2017malware,grosse2017adversarial} rather than rigorous robust optimization, and evaluate against weak or unrealistic benchmarks~\cite{grosse2017adversarial,xu2023ofei,dyrmishi2023empirical,li2023pad,doan2023feature,chen2021using}. This disjointed landscape hinders a comprehensive understanding of AT's efficacy, leaving the field with critical unknowns, such as true role of AE realizability and its impact on the robustness-accuracy trade-off. This raises a fundamental question: \\

\noindent\textit{What are the factors contributing to the success and failure of AT in the malware domain?}\\
% \textit{Why does adversarial training succeed or fail in the malware domain?}

To bridge this gap, we introduce \textit{Rubik}, a unified framework that treats AT as a multidimensional puzzle, enabling a systematic study of how its components interact in malware detection.
% we introduce \hb{\textit{Rubik}}, a unified framework that systematically rethinks AT for malware detection.
Rubik provides a structured methodology to integrate and control the key dimensions of AT---from data and feature representations to classifiers and robust optimization settings---within a coherent system.
% , \hb{highlighting} how these variables jointly determine AT's effectiveness. 
Specifically, it defines interlocking factors---capturing variations in data richness, representational structure, model expressiveness, and optimization tuning---that form the core faces of the puzzle our framework solves, demonstrating how local adjustments in one variable induce global shifts in the robustness landscape.
% it uncovers ten critical factors---such as feature-space dimensionality and model flexibility---that shape robustness outcomes, 
Rubik also highlights evaluation aspects like adversarial realism that give a truer picture of malware classifiers performance. Applying Rubik reveals that AT's effectiveness is far from absolute, but instead governed by the intricate interplay of these variables.

We then instantiate Rubik for static malware detection in the Android ecosystem---a richly instrumented platform offering an extensive set of timestamped datasets and interpretable intermediate representations, making it particularly suitable for exploring AT's multidimensional effects.
% , a well-studied platform that provides a solid foundation for our analysis. 
While Rubik is generalizable across platforms, exploring all variable combinations (e.g., feature representations and classifiers) is computationally intractable (see Appendix~\ref{app:exhaustive_exploration}). We therefore focus on well-established configurations that allow a feasible yet representative evaluation, including, for example, datasets with different distributions, diverse classifier types, and realistic evasion attacks. Feature representations are chosen for which realistic evasion attacks are available, as constructing such attacks for arbitrary representations remains an open challenge. By systematically varying datasets, representations, and classifier types across robust optimization settings, we provide a comprehensive, actionable view of AT's performance and the conditions under which it succeeds or fails.

Our exploration refines key insights from prior work: while realizable AEs help preserve clean accuracy~\cite{cortellazzi2024intriguingpropertiesadversarialml,lucas2023adversarial}, their robustness benefits~\cite{dyrmishi2023empirical,cortellazzi2024intriguingpropertiesadversarialml} are conditional. We further challenge~\cite{li2023pad} by showing that high-confidence AEs are not always required, and contest observations in~\cite{dyrmishi2023empirical,lucas2023adversarial} by demonstrating that AT can meaningfully benefit from increasing the number of AEs. Moreover, our evaluation reveals novel synergies between model architecture and feature-space structure, revealing that their interplay---mediated by robust optimization settings---shapes how effectively AT hardens Android malware classifiers. Specifically
% Through this exploration, 
we derive four key insights and highlight four prevalent misconceptions in robustness evaluation, each paired with practical recommendations. 
Our findings show that (i) model architecture dictates how AT adapts or fails, (ii) the ratio of adversarial to clean samples governs the robustness-accuracy balance, (iii) dense, low-dimensional feature spaces enable more effective AT when models and attacks are aligned, and (iv) robustness in discrete spaces depends primarily on feature alignment rather than decision-boundary smoothness. 
% Furthermore, we reveal flawed evaluation practices, such as relying solely on pure robustness, 
Finally, Rubik enables us to empirically expose flawed evaluation practices---such as relying solely on absolute robustness rather than relative robustness accounting for a model's baseline robustness, assuming perfect-knowledge attacks represent the worst case, or overlooking true constraints when assessing realizability---and to offer actionable guidance to address them.
Our \textbf{contributions} can be summarized as follows:

\begin{itemize}
    \item We introduce \textit{Rubik} (\S\ref{sub:unified_evaluation_framework}), a unified framework that defines key factors---such as feature dimensionality and AE realism---across multiple dimensions, analyzing how their interplay affects AT effectiveness in the malware domain. To ensure reproducibility, the code is publicly available at \url{https://anonymous.4open.science/r/robust-optimization-malware-detection-C295}.
    \item Our evaluation on Android malware classifiers demonstrates that clean and robust performance derived from AT depends on the interaction of these factors, challenging prior influential studies~\cite{dyrmishi2023empirical,lucas2023adversarial,doan2023feature,cortellazzi2024intriguingpropertiesadversarialml,li2023pad}, by showing that key assumptions in robust optimization---such as the universal benefits of realizable and high-confidence AEs, and the minimal impact of adversarial budget---are highly conditional on the interplay of data, feature representations, and classifiers.
    % and showing that AT's success is not absolute but requires careful tuning of parameters alongside variations in data, feature representations, and classifiers.
    \item Extensive experiments reveal four key novel insights (\S\ref{takeaway1}) and four common flawed evaluation assumptions (\S\ref{hypothesis1}), each accompanied by practical recommendations to improve AT methodology and understanding.
\end{itemize}

\section{Background}
\vspace{-0.5em}
\label{sec:background}
In this section, we briefly describe evasion attacks and Machine Learning (ML) hardening.

%\subsection{ML for Static Analysis}
%Static analysis serves as a primary method for detecting malicious programs, categorizing them based on their source code without executing them. In recent years, ML has garnered significant attention in both academic and industrial settings for malware detection. ML algorithms excel in statically analyzing complex datasets of program codes to identify patterns that distinguish malicious software from benign counterparts. By training binary classifiers on extensive datasets, ML models hold promise in detecting zero-day threats, surpassing traditional signature-based methods~\cite{Jordaney2017Transcend}. However, ML-based static analysis encounters challenges such as evasion attacks, where alterations are made to source codes to evade detection without altering their functionality.
% However, ML-based static analysis faces challenges such as evasion attacks, where code is altered to evade detection without changing its functionality. Additionally, high-dimensional features can cause inefficiencies and overfitting, impacting model performance \cite{Papernot2016Limitations}.
 \vspace{-0.5em}
 \subsection{Evasion Attacks}
 \vspace{-1em}
% ML-based static analysis faces challenges such as evasion attacks, where the code is altered to evade detection without changing its functionality. The goal of an adversary in evasion attacks is to perform targeted or untargeted attacks to change the predicted class assigned by the classifier. In targeted attacks, the objective is to alter the prediction to a desired, predefined class; however, in untargeted attacks, any change in the predicted label from its original classification suffices. Evasion attacks can be categorized into feature-space attacks, which modify the input features, and problem-space attacks, which directly manipulate domain-specific objects~\cite{Goodfellow2014Explaining, Papernot2016Limitations, biggio2018wild}, such as Android apps. The transformation between feature space and problem space is neither differentiable nor invertible, complicating the adversary's task in these dimensions.
ML-based static analysis encounters challenges such as evasion attacks, where code is modified to avoid detection without altering its function. An adversary's goal in these attacks is to change the classifier's predicted class, either through targeted attacks aiming for a specific class or untargeted attacks where any misclassification is sufficient. Evasion attacks fall into two categories: feature-space attacks that modify input features, and problem-space attacks that directly manipulate domain-specific objects like Android apps.
% ~\cite{Goodfellow2014Explaining, Papernot2016Limitations, biggio2018wild}. 
The non-differentiable and non-invertible nature of the transformation between these two spaces complicates the adversary's task~\cite{Pierazzi2020Intriguing}.\\
\noindent{\textbf{Feature-space evasion attacks.}} 
%These adversarial attacks operate directly on the feature vector representing the input data to the ML model. By making subtle modifications to the feature values, attackers can deceive the model into making incorrect classifications. This type of attack is particularly insidious because it requires knowledge of the model's features and how they are processed but does not necessarily require direct access to the model itself. The simplicity and effectiveness of feature-space attacks highlight the vulnerabilities inherent in relying solely on ML for security~\cite{Biggio2013Evasion}. Feature-space evasion attacks can be categorized into constrained and unconstrained attacks. Constrained attacks target classification datasets by considering inherent domain-specific constraints (e.g., feature immutability or non-linear relationships between features), while unconstrained feature-space attacks ignore these constraints.
These adversarial attacks manipulate the feature vector that serves as input to the ML model. By subtly altering feature values, attackers can mislead the model into making incorrect classifications. Such attacks are especially insidious because they require knowledge of the model's features and processing, but not necessarily direct access to the model itself. Their simplicity and effectiveness underscore the vulnerabilities of relying solely on ML for security~\cite{Biggio2013Evasion}. Feature-space evasion attacks are categorized as constrained or unconstrained. Constrained attacks target classification datasets while respecting domain-specific constraints like feature immutability, whereas unconstrained attacks ignore them.\\
\noindent{\textbf{Problem-space evasion attacks.}} These adversarial attacks manipulate input data directly, such as altering an Android malware app without affecting its malicious function. They are often complex, requiring a deep understanding of how such changes impact the data's representation in feature space. Problem-space attacks are considered more practical than feature-space attacks because they craft real-world malware that is inherently difficult for models to classify correctly~\cite{Pierazzi2020Intriguing}.\\ \indent
Evasion attacks are also broadly categorized as \textit{realistic} or \textit{unrealistic}, regardless of their origin~\cite{bostani2022domain}. Realistic attacks produce realizable AEs that adhere to domain constraints, making them resemble legitimate programs. Unrealistic attacks, by contrast, may generate unrealizable AEs that do not resemble legitimate software.

\subsection{ML Hardening}
\vspace{-0.5em}
% To defend against evasion attacks, robust optimization techniques~\cite{madry2017towards} have become a pivotal focus. These techniques are designed to enhance the resilience of ML models against adversarial attacks, which manipulate input data to cause misclassification. Among these techniques, adversarially robust optimization, also known as adversarial training, plays a crucial role in fortifying models by exposing them to AEs during the training phase. This exposure aims to improve the model's ability to generalize from adversarial perturbations it might encounter in real-world scenarios~\cite{madry2017towards}. AT involves incorporating AEs into the training process, thereby enabling the model to learn from these perturbations and make more robust predictions. Specifically, the most established strategy~\cite{madry2017towards} is to iteratively generate high-confidence AEs--worst-case AEs that cause the highest loss-- and update the model parameters to minimize the classification error on these examples. This method has been shown to significantly enhance the model's resilience against certain types of attacks, though it may not guarantee protection against all possible adversarial inputs~\cite{Goodfellow2014Explaining}. In addition to AT, adversarial retraining~\cite{szegedy2013intriguing} follows a similar approach but is based on a simpler idea. This defense strategy involves enriching the training dataset from scratch with AEs generated by an evasion attack.
A primary defense against evasion attacks is robust optimization, which builds model resilience directly into the training process. A cornerstone technique for robust optimization is adversarial training~\cite{madry2017towards}, which exposes models to AEs during the training phase. This improves their ability to generalize against adversarial perturbations. The established method involves iteratively generating high-confidence AEs---perturbed inputs that cause high loss---and updating the model to correctly classify them. While this method can significantly boost resilience against certain attacks, it does not guarantee protection against all adversarial inputs~\cite{Goodfellow2014Explaining}. A related but simpler strategy is adversarial re-training, which involves enriching the initial training dataset with AEs generated by an evasion attack~\cite{szegedy2013intriguing}.
\section{Methodology}
% This study highlights the intertwined factors crucial for effective AT. We first formulate the problem and define key dimensions, such as feature representations and classifiers that must be explored to understand AT's effectiveness. Then, we present a unified framework that introduces diverse training and evaluation factors to explore AT from various perspectives. Finally, we outline a systematic evaluation to understand how these factors impact adversarial robustness and malware classifier performance on clean data. Exploring AT through different training factors is vital, as practitioners must select appropriate configurations that will inherently influence their detectors' performance in real-world scenarios. 
This work investigates the core, interconnected factors that govern effective AT. We begin by formulating the problem and defining the essential dimensions---such as feature representations and classifiers---that shape AT's success. To navigate this complex landscape, we introduce Rubik, a unified framework for exploring AT through diverse training and evaluation perspectives. We then detail a systematic evaluation to precisely quantify how these factors trade off between adversarial robustness and performance on clean data. This multi-faceted analysis provides practitioners with the critical insights needed to configure robust detectors that perform reliably in the real world.

\subsection{Problem Definition}
\label{sub:problem_definition}
% The main objective of this work is to understand how to increase the robustness of malware classifiers while maintaining high performance on clean data (referred to as clean performance in this study), by defining the elements and characteristics that influence the process. More formally,
Suppose $\mathcal{Z}$ represents the problem space encompassing all potential objects (e.g., Android apps or Windows programs). Additionally, $\mathcal{F}$ denotes the feature representation that characterizes the dimensions of the feature space $\mathcal{X}$. For utilizing ML in malware detection, each object $z \in \mathcal{Z}$ is first mapped to $x \in \mathcal{X}$ through a mapping function $\psi: \mathcal{Z} \rightarrow \mathcal{X}$. Malware detection is then performed by a binary classifier $f: \mathcal{X} \rightarrow \mathcal{Y}$ with a discriminant function $g: \mathcal{X} \times \mathcal{Y} \rightarrow \mathcal{R}$, where $f(x) = \arg\max_{i \in \mathcal{Y}} g_i(x)$ determines the label of $x \in \mathcal{X}$ from the label space $\mathcal{Y} = \{0, 1\}$. Specifically, $x$ is classified as malware if $f(x) = 1$, and as benign, otherwise.\\ \indent
Evasion attacks can transform a malware sample $x \in \mathcal{X}$, which is correctly classified (i.e., $f(x) = 1$), into an AE by finding an adversarial perturbation $\delta$ that changes the prediction to $f(x + \delta) = 0$ when added to $x$. Mathematically, identifying $\delta$ corresponds to solving the following optimization problem:
% \begin{equation} \label{eq_evasion_attack}
% \begin{aligned}
% \arg\min_{\delta \vDash \Omega} \quad & g_{0}(x+ \delta)\\
% \end{aligned}
% \end{equation}

\begin{equation} \label{eq_evasion_attack}
\begin{aligned}
x^* = \arg\max_{x' \in \mathcal{N}(x)} \quad & \mathcal{L} (g_{y}(x'),\theta,y)\\
\end{aligned}
\end{equation}

\noindent where $x'=x+\delta$, and $\mathcal{L}$ and $\theta$ denote the loss function and parameters of the classifier $f$, respectively. Moreover, $\mathcal{N}(x)$ represents the set of allowed perturbations. The common constraint in finding $\delta$, which represents the allowed perturbations, is a norm bound $\|x - x'\|_p \leq \epsilon$, where $\epsilon$ signifies the magnitude of the maximum allowed changes.
% however, most existing studies consider simple norm bounds when finding $\delta$. 
To harden $f$ against adversarial perturbations, robust optimization aims to adjust the parameters of $f$ by incorporating eq.~\ref{eq_evasion_attack} in the training process and solving the following min-max optimization problem:

\begin{equation}
    \begin{split}
    \min_{\theta} \mathbb{E}_{(x_i,y_i) \sim \mathcal{D}} \left[ \max_{\|x_i - x'_i\|_p \leq \epsilon} \mathcal{L} (g_{y_i}(x'_i),\theta,y_i) \right]
    \end{split}
    \label{eq:robust_optimization}
\end{equation}

\noindent where $\mathbb{E}$ denotes the expected value of the inner maximization problem, considering that $(x_i, y_i) \sim \mathcal{D}$ are training data samples drawn from the distribution $\mathcal{D}$. It is important to note that adversarial robustness implies that $f$ can accurately classify any variations of $x_i$, demonstrated by $x'_i = x_i + \delta$, as long as $x'_i \in \mathcal{N}(x_i)$. In other words, $f$ exhibits adversarial robustness under $\|x_i - x'_i\|_p \leq \epsilon$ if $f(x_i) = f(x'_i)$.  

\subsection{Characterizing the Effectiveness of AT}
\label{subsec:characterizing_the_effectiveness_of_at}
% To assess the robustness of malware detectors derived from AT, it is crucial to initially identify the factors that might significantly impact the effectiveness of AT. Subsequently, comprehending how these factors influence both clean performance and the adversarial robustness of hardened malware detectors is essential. According to eq.~\ref{eq:robust_optimization}, taking the following key dimensions into account is vital for identifying factors that are likely to significantly influence the performance of AT:
To properly assess the robustness of malware detectors derived from AT, we must first identify the most impactful factors. We then need to understand how these factors influence both the clean performance and adversarial robustness of the hardened detectors. As shown in eq.~\ref{eq:robust_optimization}, the following key dimensions are vital for identifying these influential factors:

\begin{figure}[t!]
    \centering
    \includegraphics[width=0.6\columnwidth]{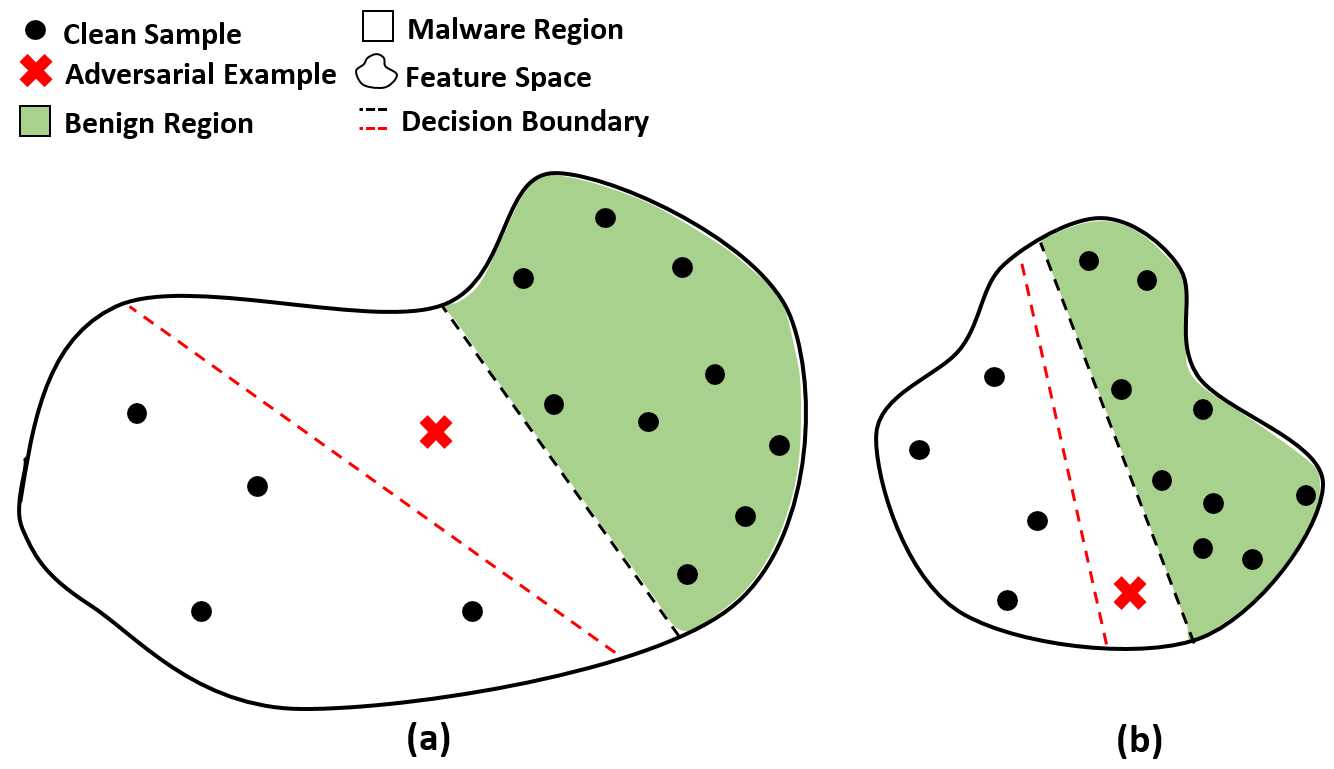}
    \caption{The impact of feature representation on data distribution and blind spot coverage, showing potentially (a) larger, more vulnerable regions and (b) smaller, less vulnerable ones.
    % Illustrating the impact of feature representation on altering the data distribution and covering blind spots (i.e., the vulnerable regions between the decision boundary and the benign region).
    }
    \vspace{-1.5em}
    \label{fig:feature_representation}
\end{figure}
\begin{figure}[b!]
    \centering
    \includegraphics[width=0.7\columnwidth]{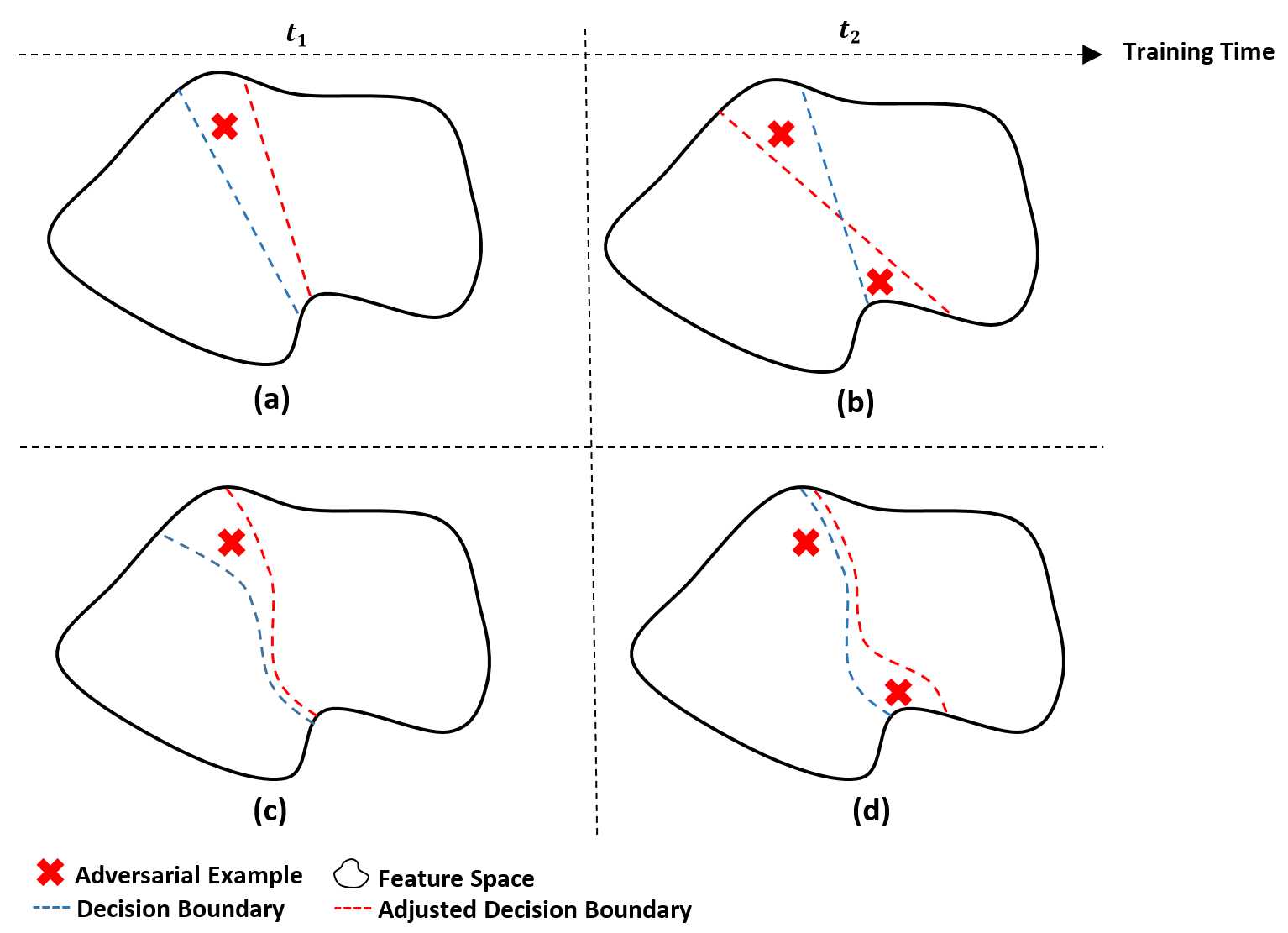}
    \caption{Demonstrating the impact of classifiers on AT: (a, b) A linear classifier may forget earlier adjustments (at $t$) when facing new AEs (at $t+1$); (c, d) A non-linear classifier better adapts to new AEs.
    % Demonstrating the impact of classifiers on AT: (a) and (b) show that a linear classifier, due to its low flexibility, may lose the adjustments made to its decision boundary at time $t$ based on an AE when encountering a new AE at time $t+1$. In contrast, (c) and (d) illustrate that a non-linear classifier is more adaptable when encountering new AE.
    }
    \vspace{-1.5em}
    \label{fig:classifier}
\end{figure}
\begin{figure}[t!]
    \centering
    \includegraphics[width=0.8\columnwidth]{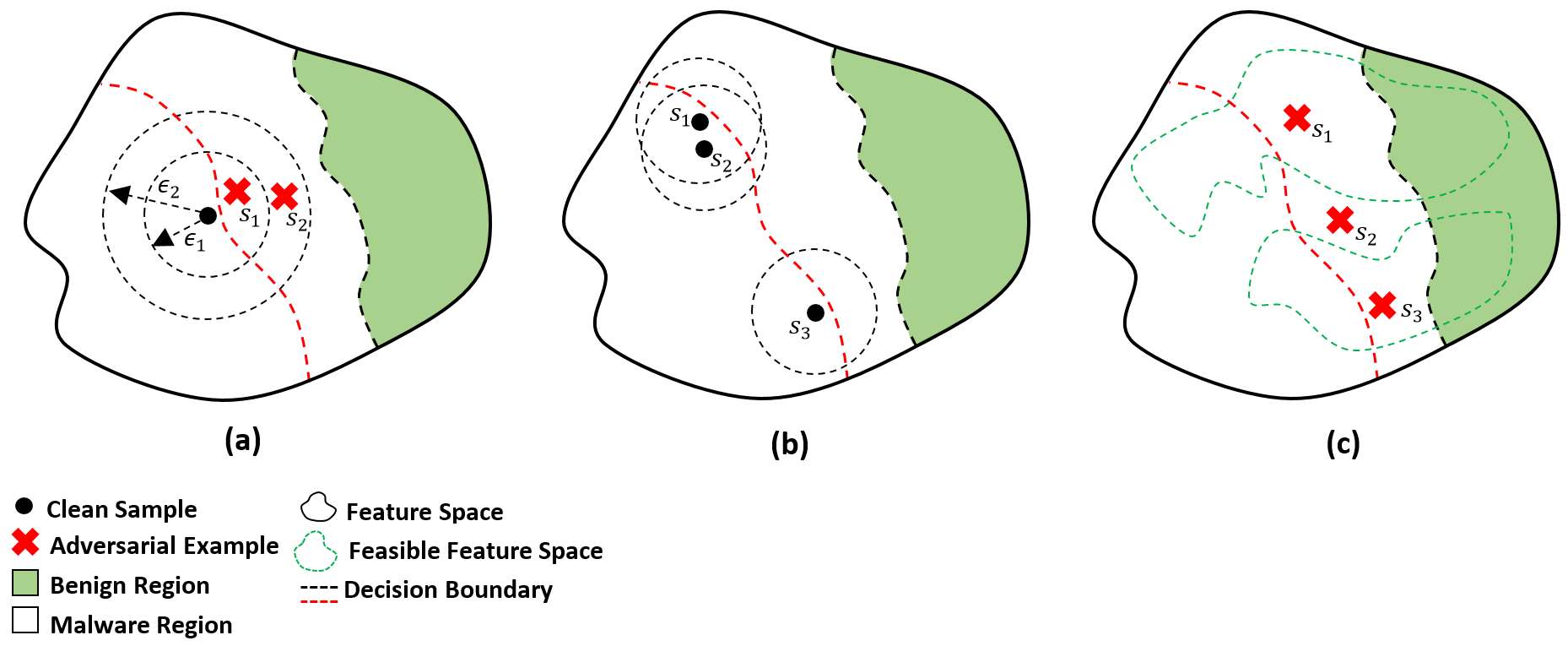}
    \caption{Illustrating how different settings affect robust optimization: (a) larger perturbation bounds (e.g., $\epsilon_2$) and high-confidence AEs (e.g., $s_2$) can reveal more blind spots; (b) varying malware sets lead to different blind spot coverage, e.g., $s_1$, $s_2$, and $s_3$ being more effective than $s_1$, $s_2$; (c) targeting blind spots within the feasible space is sufficient, as only realizable AEs (e.g., $s_1$, $s_3$) represent practical threats.
    % Illustrating the influence of different settings on the performance of robust optimization: (a) demonstrates that a large perturbation bound (e.g., $\epsilon_2$), along with a high-confidence AE (e.g., $s_2$) can potentially reveal more blind spots. (b) shows that using different sets of malware samples results in varying effects on uncovering blind spots, e.g., $s_1$, $s_2$, and $s_3$ being more effective than $s1$ and $s_2$. (c) indicates that in AT, uncovering only those blind spots within the feasible feature space is sufficient, as realistic evasion attacks target these regions (e.g., $s_1$ and $s_3$ are realizable AEs, whereas $s_2$ is not realizable).
    }
    \vspace{-1.5em}
    \label{fig:robust_optimization_settings}
\end{figure}

\noindent\textbf{Data and Feature Representations.}
%Eq.~\ref{eq:robust_optimization} shows that the training set is crucial for AT as robust optimization is performed in it. When the empirical distribution of the training set diverges from the true data distribution, AT may become ineffective since adversaries can generate AEs that fall outside the empirical distribution of the training set\cite{zhang2019limitations}. 
% One of the primary dimensions significantly affecting the distribution of the training samples and the coverage of blind spots is feature representation. As shown in Figure~\ref{fig:feature_representation}, the distribution of training samples and the size of the blind spots can be altered by using different feature representations. Specifically, in discrete feature spaces, using low-dimensional feature representations can reduce the vulnerable region compared to high-dimensional feature representations, intuitively impacting the capabilities of uncovering blind spots by AT. This can be extended on continuous feature space with domain constraints that would \textit{discretize} the space. Our experiments, especially the analysis in \S\ref{subsubsec:perturbation_bound_and_the_confidence_of_aes}, tend to provide empirical insights into this matter. 
Eq.~\ref{eq:robust_optimization} establishes the training set's foundational role in AT, as robust optimization is performed directly on it. A divergence between this set's distribution and the true data distribution can cause AT to fail, as adversaries can generate AEs outside its scope~\cite{zhang2019limitations}. Feature representation is a key dimension governing this distribution and the coverage of blind spots. As shown in Figure~\ref{fig:feature_representation}, different representations alter the sample distribution and the size of these blind spots. For example, in discrete spaces, low-dimensional features shrink the vulnerable region compared to high-dimensional ones, directly influencing AT's ability to uncover blind spots. Our analysis in \S\ref{subsubsec:perturbation_bound_and_the_confidence_of_aes} tends to provide empirical evidence for this matter.

\noindent\textbf{Classifiers.} 
% Our work aims at understanding the role that different learning algorithms, especially linear and non-linear classifiers, play in model hardening through AT. The intuition is that low-flexible classifiers, such as linear SVM, might be more susceptible to adversarial instability compared to high-flexible classifiers~\cite{fawzi2018analysis}, such as non-linear classifiers, resulting in varying levels of adversarial robustness. Specifically, as shown in Figure~\ref{fig:classifier}, linear classifiers might start to \textit{forget} patterns of adversarial inputs encountered earlier in the training process due to their limited flexibility~\cite{labaca2021realizable}. 
Eq.~\ref{eq:robust_optimization} highlights the importance of examining how classifier flexibility, such as linear versus non-linear models, affects hardening with AT. The intuition is that simpler, low-flexibility classifiers (e.g., linear SVM) may be more vulnerable to adversarial instability than complex, high-flexibility ones~\cite{fawzi2018analysis}, impacting robustness. As Figure~\ref{fig:classifier} shows, the limited capacity of linear classifiers can cause them to \textit{forget} earlier adversarial patterns during training~\cite{labaca2021realizable}.

\noindent\textbf{Robust Optimization Settings.} Eq.~\ref{eq:robust_optimization} indicates that the following hyperparameters might have a considerable influence on the performance of AT:
\\\noindent \textbf{(i) Perturbation bounds for identifying AEs and confidence levels of AEs.} 
Figure~\ref{fig:robust_optimization_settings}-a suggests that a larger perturbation bound $\epsilon$ should intuitively achieve higher robustness by uncovering more blind spots, specifically by finding adversarial examples $x'$ in the inner maximization that are misclassified with high confidence.\\
% As shown in Figure~\ref{fig:robust_optimization_settings}a, intuitively a larger perturbation bound $\epsilon$ should uncover more blind spots when the adversarial example $x'$ found in the inner maximization is misclassified with high confidence, achieving higher robustness.\\
\noindent \textbf{ (ii) Adversarial fractions, indicating the proportion of AEs used in AT.}
% The robust optimization in eq.~\ref{eq:robust_optimization} utilizes malware samples, underscoring the critical importance of the number of malware samples used for AT. Increasing the number of AEs can potentially enhance AT's ability to uncover more blind spots. However, as illustrated in Figure~\ref{fig:robust_optimization_settings}b, this is more likely to occur if the original malware used for AT covers a broader feature space rather than just specific narrow areas.
The robust optimization in eq.~\ref{eq:robust_optimization} relies on malware samples, highlighting the importance of their quantity for AT. While generating more AEs can uncover more blind spots, Figure~\ref{fig:robust_optimization_settings}-b shows that this might be more effective when the original malware samples cover a broad feature space, not just narrow areas.
\\\noindent \textbf{(iii) Domain constraints.}
% Realistic adversarial attacks target specific regions in the feature space to compromise ML-based malware detectors~\cite{bostani2022domain}. As shown in Figure~\ref{fig:robust_optimization_settings}, this suggests that it is sufficient for AT to uncover only those vulnerable regions that include the feature representations of realizable AEs. These vulnerable regions can be exposed by considering domain constraints during AE generation, e.g., creating realizable AEs that satisfy the domain constraints specified in the problem space~\cite{Pierazzi2020Intriguing}.
Since realistic adversarial attacks target feasible feature-space regions to compromise ML-based malware detectors~\cite{bostani2022domain}, AT need only uncover these relevant, vulnerable regions. As Figure~\ref{fig:robust_optimization_settings} suggests, these regions can be exposed by generating realizable AEs that adhere to problem-space domain constraints~\cite{Pierazzi2020Intriguing}.

% \vspace{-0.5em}
\subsection{Our Unified Evaluation Framework}
\label{sub:unified_evaluation_framework}
% This framework provides a comprehensive environment for researchers and practitioners to conduct controlled investigations into the robustness of adversarial training (AT) in malware detection. Structured as a modular and interactive tool, it enables the systematic testing of hypotheses and debugging of AT configurations by isolating and manipulating various factors.

% Each module within the framework offers diagnostic capabilities to identify sources of robustness weaknesses and optimization challenges. Here’s a breakdown of how users can leverage each module effectively to diagnose and refine adversarial training strategies:

\begin{figure}[t!]
    \centering
    \includegraphics[width=0.8\columnwidth]{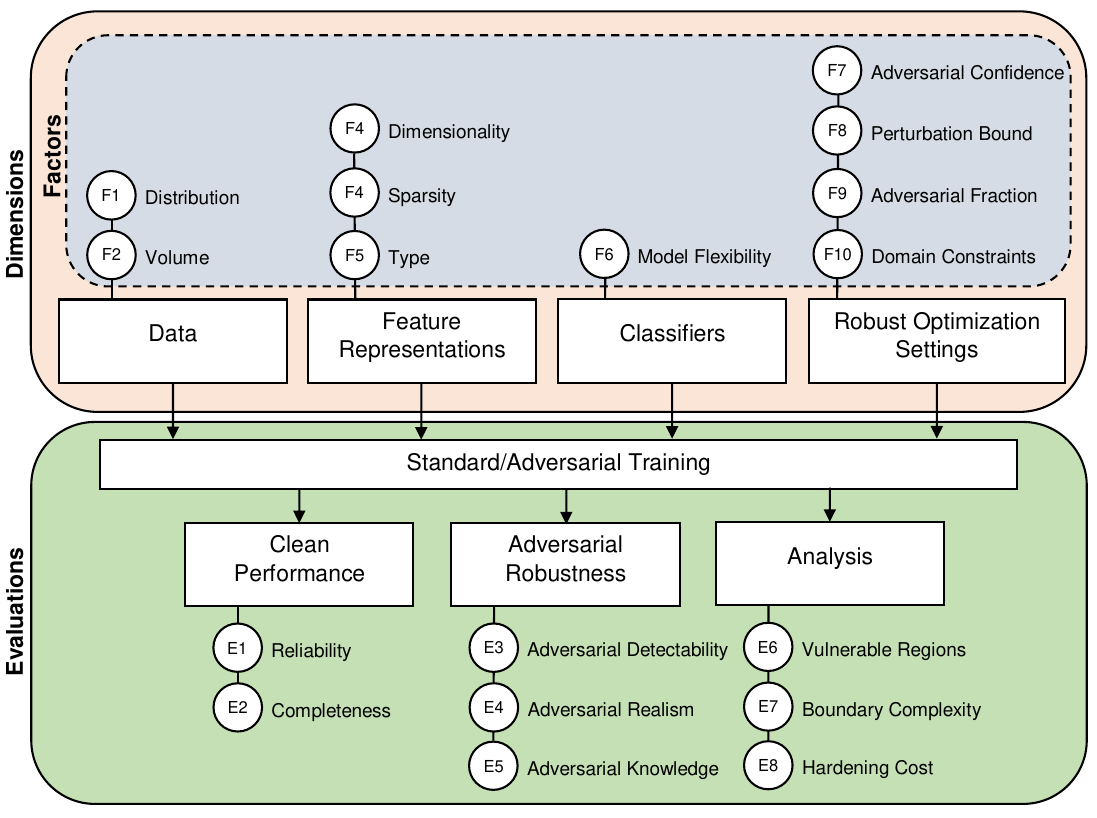}
    \caption{Illustration of our unified framework proposed to investigate the influence of various key dimensions on the performance of malware classifiers.}
    \vspace{-1.5em}
    \label{fig:framework}
\end{figure}

% To thoroughly investigate AT, we propose a unified framework named Rubik, shown in Figure~\ref{fig:framework}, that helps identify impactful factors in AT. It allows for various controlled evaluations necessary to clarify the impact of the training factors within the key dimensions defined in \S\ref{subsec:characterizing_the_effectiveness_of_at} on the success of robust optimization.
% The framework defines these dimensions for AT and essential evaluations to assess vanilla and robust models, allowing systematic hypothesis testing and debugging of AT configurations through controlled factor adjustments.
This paper proposes Rubik, a unified framework (Figure~\ref{fig:framework}) for investigating AT in the malware domain by disentangling the interplay of its key factors. Rubik enables controlled evaluations to clarify how factors within the key dimensions from \S\ref{subsec:characterizing_the_effectiveness_of_at} impact robust optimization. Defining these dimensions and essential evaluations for both vanilla and robust models facilitates systematic hypothesis testing and debugging of AT configurations.
% \vspace{-0.5em}
\subsubsection{Dimensions and Their Relevant Factors}
Rubik facilitates investigating the effectiveness of AT based on various factors across the following key dimensions:

\noindent\textbf{Data}. 
% \textit{Distribution} and \textit{volume} of data are shown to be two essential training factors in AT~\cite{schmidt2018adversarially}. Our framework allows us to explore the impacts of the variations of these factors on the effectiveness of AT. Specifically, we can import various datasets of different sizes that include real-world objects like Android Packages (APKs), with diverse distributions based on variables such as source and release date. Additionally, we can specify the proportion of samples in the training, validation, and test sets.
The \textit{distribution} and \textit{volume} of data are established as essential factors in AT~\cite{schmidt2018adversarially}. Rubik enables exploring how variations in these factors impact AT's efficacy. Specifically, it allows the import of datasets of varying sizes---including real-world objects like Android Packages (APKs)---with diverse distributions based on attributes like source and release date. It also provides control over the proportion of samples allocated to training, validation, and test sets.

\noindent\textbf{Feature Representations}. 
% Our framework enables us to explore \textit{dimensionality}, \textit{sparsity}, and \textit{types} of feature representations, as they seem to influence both model performance and computational efficiency. For instance, high-dimensional feature spaces can hinder generalization because increased features lead to sparser data points, which may cause models to capture incidental correlations instead of meaningful patterns~\cite{domingos2012few}. It is important to note that by elucidating each supported feature representation within the framework, we ensure that all real-world objects in the training and test sets are represented in the feature space according to the specified representations.
Rubik enables exploring how \textit{dimensionality}, \textit{sparsity}, and \textit{type} of feature representations influence model performance and computational efficiency. For instance, high-dimensional spaces can hinder generalization, as increased features lead to sparser data, potentially causing models to capture incidental correlations instead of meaningful patterns~\cite{domingos2012few}. By elucidating each supported representation within the framework, we ensure all real-world objects in the training and test sets are accurately mapped into the feature space.

\noindent\textbf{Classifiers.} The framework supports building various malware detectors by employing a range of learning algorithms. With support for both linear and non-linear classifiers, this key dimension facilitates a thorough investigation of how the \textit{model flexibility} influences AT.
% , which indicates the flexibility of classifiers, influences AT.

\noindent\textbf{Robust Optimization Settings.}
% \textit{Adversarial confidence}, \textit{perturbation bound}, \textit{adversarial fraction}, and \textit{domain constraints} are adjustable factors specific to AT. The framework helps us understand how variations in these factors, along with other discussed factors, contribute to strengthening malware classifiers. In the AT process, we can specify the perturbation bound for generating AEs and select different evasion attacks, either unrealistic feature-space attacks or realistic problem-space attacks, to solve the inner maximization problem in AT. The former is used to explore the influence of adversarial confidence, as different unrealistic feature-space attacks produce AEs with varying levels of misclassification confidence, while the latter is used to examine the impact of domain constraints since realistic problem-space attacks can generate realizable AEs that meet these constraints.
\textit{Adversarial confidence}, \textit{perturbation bound}, \textit{adversarial fraction}, and \textit{domain constraints} are adjustable factors specific to AT. Rubik helps understand how variations in these and other discussed factors contribute to strengthening malware classifiers. During the AT, we can specify the perturbation bound for generating AEs and select different evasion attacks---either unrealistic feature-space attacks or realistic problem-space attacks---to solve the inner maximization. The former can explores the influence of adversarial confidence, as different unrealistic attacks can meaningfully produce AEs with varying misclassification confidence, while the latter examines the impact of domain constraints by generating realizable AEs that satisfy them.

\subsubsection{Evaluations}
% The proposed framework provides options for building either vanilla or robust malware detectors through standard or adversarial training. Exploring robust optimization settings is solely crucial for AT, while the remaining factors in the key dimensions are relevant for both types of training. Once ML models are built according to the configurations outlined in the framework's key dimensions, we can assess their performance using the following evaluation aspects. Please note that in \S\ref{sec:discussion}, we discuss some questionable evaluation assumption that might occur when exploring evaluation factors.
The proposed framework allows building either vanilla or robust malware detectors via standard or adversarial training. Exploring robust optimization settings is especially important for AT, while the other key dimensions apply to both types of training. Once ML models are configured according to the framework's key dimensions, we can evaluate their performance using the following aspects. Note that in \S\ref{sec:discussion}, we highlight some potentially questionable evaluation assumptions that may arise when exploring these factors.

\noindent\textbf{Clean Performance.}
% Two important evaluation factors in specifying the performance of malware classifiers in the absence of adversarial attacks are \textit{reliability} and \textit{completeness}, offering practical insight into their impact beyond formal definitions. Reliability ensures the classifier accurately detects malware without mislabeling goodware, while completeness ensures it identifies all malicious instances. The framework includes metrics like F1-score to clarify these aspects.
Two key evaluation factors for malware classifiers in the absence of adversarial attacks are \textit{reliability} and \textit{completeness}, providing practical insight beyond formal definitions. Reliability ensures the classifier detects malware without mislabeling goodware, while completeness ensures all malicious instances are identified. Rubik includes metrics such as F1-score to capture these aspects.

\noindent\textbf{Adversarial Robustness.}
% Our proposed framework supports three key evaluation factors for assessing adversarial robustness: \textit{adversarial detectability}, \textit{adversarial realism}, and \textit{adversarial knowledge}. While adversarial detectability measures the classifier's ability to detect AEs, adversarial realism ensures that only realistic evasion attacks are considered. Additionally, robustness is assessed against attacks with varying levels of knowledge about the classifier.
Rubik supports three key factors for evaluating adversarial robustness: \textit{adversarial detectability}, \textit{adversarial realism}, and \textit{adversarial knowledge}. Adversarial detectability measures the classifier's ability to detect AEs, adversarial realism ensures only realistic evasion attacks are considered, and adversarial knowledge emphasizes that robustness should be evaluated against attacks with varying levels of knowledge about the classifier.

\noindent\textbf{Analysis.} 
\label{subsec:analysis}
The framework provides a collection of tools for plotting, such as t-SNE visualization~\cite{van2008visualizing}. Besides regular plots, two following tools are supported to further interpret the functionality of AT:

% \noindent \textbf{(a)
\noindent \textbf{(i) Joint Feature Importance}. 
\label{subsec:joint_feature_importance}
% This tool is designed to assess how \textit{vulnerable regions} in the feature space are protected by AT. Specifically, to evaluate which features are important for both AT and a realistic evasion attack, we proposed a technique utilizing the Joint Distribution Plot (JDP). To this end, we first determine the frequency of alteration for each feature within the feature space during AT and attacking, as this frequency can be seen as a metric indicating the importance of features for either AT or the realistic attack. For example, if $f_1 \in \mathcal{F}$ is involved in transforming $x_1 \in \mathcal{X}$ and $x_2 \in \mathcal{X}$ into AEs during AT, its frequency is 2. We then use the JDP to visualize the overlap in the importance of features between AT and the evasion attack. The JDP is based on the Probability Distribution Function (PDF) and is plotted by estimating the values of two random variables: one representing the importance of features for AT and the other for the evasion attack. Using the PDF, the distribution of every feature is displayed along the x-axis and y-axis according to the frequencies determined by AT and the evasion attack, respectively. Refer to $\S\ref{subsubsec:properties_of_aes}$ to see an example for JPD.
This tool is designed to assess how \textit{vulnerable regions} in the feature space are protected through AT. Specifically, to identify which features are important for both AT and a realistic evasion attack, we propose a technique based on the Joint Distribution Plot (JDP). We first measure how often each feature is altered during AT and the attack, as this frequency reflects the feature's importance to each process. For example, if $f_1 \in \mathcal{F}$ is involved in transforming $x_1, x_2 \in \mathcal{X}$ into AEs during AT, its frequency is 2. The JDP then visualizes the overlap in feature importance between AT and the evasion attack. It is derived from the Probability Distribution Function (PDF) and is plotted by estimating two random variables: one representing feature importance in AT and the other in the evasion attack. Using the PDF, the distribution of each feature is shown along the x- and y-axes according to the frequencies determined by AT and the attack, respectively. Refer to $\S\ref{subsubsec:properties_of_aes}$ for an example of the JDP.
 
 % \noindent\textbf{(b)
 \noindent\textbf{(ii) Decision-function Roughness}. \label{subsec:decision_function_roughness}
%  To investigate whether the adversarial vulnerability is related to \textit{boundary complexity,} which indicates the shape of the decision surface learned by the models, we adapt the technique proposed in~\cite{eghbal2024rethinking} for use in our framework. This method estimates a model's prediction-change risk $r$ by comparing the predictions of synthetic samples, uniformly drawn within the $\epsilon$-bound of a training sample $x \in \mathcal{X}$, with the prediction of $x$. A larger $r$ indicates a rougher decision function. For more details about this measurement, refer to~\cite{eghbal2024rethinking}. 
% Besides the above analysis factors, our framework includes the \textit{hardening cost}, which captures the computational resources needed for AT, helping assess its practical feasibility.
To examine whether adversarial vulnerability is linked to \textit{boundary complexity}---the shape of the decision learned by the models---we adapt the technique proposed in~\cite{eghbal2024rethinking} for use in our framework. This method estimates a model's prediction-change risk $r$ by comparing the predictions of synthetic samples, uniformly drawn within the $\epsilon$-bound of a training sample $x \in \mathcal{X}$, with the prediction of $x$. A higher $r$ value indicates a rougher decision function. For more details, see~\cite{eghbal2024rethinking}.

In addition to the above analysis factors, Rubik considers the \textit{hardening cost}, which captures the computational resources required for AT, helping evaluate its practical feasibility.

\subsection{Systematic Evaluation}
\label{sub:systematic_evaluation}
To evaluate the influence of the key factors outlined in $\S\ref{subsec:characterizing_the_effectiveness_of_at}$, we design a wide range of experiments that enable controlled assessments. Specifically, these experiments are crafted to explore the following RQs:\\
\\

% To characterize the resilience of malware classifiers steamed from AT, understanding the effects of different factors on their clean performance and adversarial robustness is paramount. To this end, we explore the following research questions:\\

% \hb{We must include some images showing why exploring the following research questions makes sense.}\\
\vspace{-0.4cm}
\noindent\textbf{\hypertarget{RQ1}{RQ1}:} What factors significantly affect the effectiveness of AT?\\

To address this research question, we conduct a systematic set of experiments that vary the factors identified across our key dimensions to examine their influence on the effectiveness of AT. Specifically, our experimental setup considers the following aspects:
\begin{itemize}
\item \textbf{Data:} We use multiple datasets comprising raw software objects, denoted as $\mathcal{D}_1$, $\mathcal{D}_2$, etc.
\item \textbf{Feature Representations:} We evaluate different feature representations, denoted as $\mathcal{F}_1$, $\mathcal{F}_2$, etc.
\item \textbf{Classifiers:} We train and evaluate a variety of classifiers, denoted as $\mathcal{C}_1$, $\mathcal{C}_2$, etc.
\item \textbf{Perturbation Bounds:} We vary the perturbation bound $\scalebox{1.5}{$\varepsilon$}$ across a range of values, e.g., $[0, 5, 10, ..., 100]$.
\item \textbf{Adversarial Fractions:} We adjust the fraction of AEs used for AT, $\mathcal{A} = [10\%, 20\%, ..., 100\%]$, where each $\alpha_i \in \mathcal{A}$ specifies the percentage of malware samples per epoch available for AT.
\item \textbf{Evasion Attacks:} We employ several evasion attacks, denoted as $\mathcal{A}ttack_1$, $\mathcal{A}ttack_2$, etc., to generate AEs for both training and evaluation. Our attack set includes feature-space attacks for analyzing the influence of both AE confidence and AE fraction on AT, and problem-space attacks for assessing the effect of domain constraints. The latter yield realizable AEs that satisfy domain constraints and are also used to evaluate adversarial robustness.
\end{itemize}

% For each combination of data, feature representation, and classifier, we conduct a series of experiments by varying the perturbation bounds, the adversarial fractions, and evasion attacks. For instance, we first combine data $\mathcal{D}_1$, feature representation $\mathcal{F}_1$, and classifier $\mathcal{C}_1$, and evaluate the performance at each $5 \in \scalebox{1.5}{$\varepsilon$}$ and $50\% \in \mathcal{A}$ using different $\mathcal{A}ttack$. We then continue in this manner for all combinations.\\
We systematically conduct experiments for each combination of data, feature representation, and classifier by varying the perturbation bounds, adversarial fractions, and evasion attacks. For example, we combine dataset $\mathcal{D}_1$, feature representation $\mathcal{F}_1$, and classifier $\mathcal{C}_1$, and evaluate their performance across different values of $\scalebox{1.5}{$\varepsilon$}$ (e.g., 5) and $\mathcal{A}$ (e.g., 50\%) using various $\mathcal{A}ttacks$. This process is repeated for all possible combinations.\\
\\
 
\vspace{-0.4cm}
\noindent\textbf{\hypertarget{RQ2}{RQ2}:} What properties of the generated AEs influence the outcomes of the hardening process?\\

% Different evasion attacks generate AEs by leveraging distinct logical approaches, leading to varying levels of adversarial robustness. Therefore, each method uniquely influences the training process, resulting in different coverage of blind spots (i.e., vulnerable regions in the decision space of classifiers) and decision boundaries. To investigate this RQ, we examine the coverage of blind spots and decision boundaries using analysis tools prepared in the framework, particularly Joint Feature Importance and Decision-function Roughness.
Different evasion attacks generate AEs using distinct underlying strategies, which in turn lead to varying degrees of adversarial robustness. Consequently, each attack type influences the training process differently, affecting how well blind spots (i.e., vulnerable regions in the classifier's decision space) and decision boundaries are covered. To explore this RQ, we analyze the coverage of blind spots and the shape of decision boundaries using the framework's analysis tools---specifically, Joint Feature Importance and Decision-function Roughness.

\section{Experiments and Evaluations}
% \subsection{Experimental Settings}
% \label{sec:settings}

% \noindent{\bf Scope of Analysis.} 
\subsection{Scope of Analysis}
AT is platform-independent, operating on feature representations rather than raw problem-space objects. However, to avoid dataset bias and enable comprehensive exploration, it is crucial to collect a large and diverse set of malware and goodware samples. The Android ecosystem is ideal for this, as it supports many interpretable intermediate representations and provides extensive timestamped APKs through repositories like AndroZoo~\cite{Allix:AndroZoo}, ensuring the volume and variety required for reliable analysis.

Building on this foundation, this paper emphasize the need for practitioners to fine-tune key parameters (e.g., adversarial fraction) in robust optimization to ensure its effectiveness. Our framework systematically illustrates how these parameters interact with variations in data, feature representations, and classifiers. Given the practical challenges of selecting optimal representations and classifiers, we rely on well-established solutions to stay focused on our main research objectives. Specifically, to explore training factors such as \textit{feature dimensionality} and \textit{model flexibility}, we employ multiple datasets (DREBIN20~\cite{Pierazzi2020Intriguing} and APIGraph~\cite{zhang2020enhancing}), feature representations (DREBIN~\cite{Arp:Drebin} and RAMDA~\cite{li2021robust}), and classifiers (linear Support Vector Machine (SVM), Decision Tree (DT), and Deep Neural Network (DNN)). To study evaluation factors like \textit{adversarial realism} and \textit{adversarial knowledge}, we consider both unrealistic (PGD~\cite{madry2017towards}, JSMA~\cite{Papernot2016Limitations}) and realistic (PK-Greedy~\cite{Pierazzi2020Intriguing}, EvadeDroid~\cite{bostani2024evadedroid}) attacks, generating AEs from 1K randomly selected clean malware samples from the test set (true positives). Full details of our experimental settings and implementation appear in Appendices~\ref{app:settings} and~\ref{app:implementation_details}.
\subsection{Systematic Evaluations}
This section examines \hyperlink{RQ1}{RQ1} in $\S\ref{subsubsec:perturbation_bound_and_the_confidence_of_aes}$, $\S\ref{subsubsec:varying_fraction_of_aes}$, and $\S\ref{subsubsec:domain_constraints}$, as well as \hyperlink{RQ2}{RQ2} in $\S\ref{subsubsec:properties_of_aes}$, to analyze the interacting effects of key factors on the effectiveness of AT and identify which AE properties influence its performance. As outlined in $\S\ref{sub:systematic_evaluation}$, our systematic evaluations are structured around a series of experiments, each employing various combinations of data, feature representations, and classifiers.
% classifiers and feature representations.
% experimental design is based on  classifiers and feature representations are two essential factors that influence other variables in robust optimization settings. 
% Therefore, utilizing various classifiers and feature representations is the standard configuration throughout all explorations.

\subsubsection{Standard Evaluation Configuration} 
\label{susbsec:standard_evaluation_Configuration}
% PGD and JSMA are two feature-space attacks within the optimization module of our proposed framework. They are used to harden baseline malware detectors in most of our evaluations by generating adversarial examples (AEs) with varying confidence levels (see Appendix~\ref{app:confidence_of_AEs}). Additionally, given that both DREBIN and RAMDA are binary feature representations, we consider the $\ell_0$-norm (i.e., the number of changes) in eq.~\ref{eq:robust_optimization}, because it is the common perturbation bound for binary feature representations~\cite{carlini2019evaluating}. 
% Moreover, to assess adversarial robustness, we evaluate the models' resistance to bounded PK-Greedy and EvadeDroid attacks, ensuring that their attack bounds match those used in the evasion attacks during AT. Considering the attack bounds enables a meaningful comparison of defense approaches~\cite{nasr2025evaluating}. Additionally, it is imperative to confirm that the observed adversarial robustness against evasion attacks is attributable to AT. To this end, we subtract the baseline robustness of the vanilla model $\mathcal{M}_v$ against bounded attacks from that of the hardened model $\mathcal{M}_h$ as follows:
PGD and JSMA are two feature-space attacks within the optimization module of our proposed framework. They harden baseline malware detectors in most evaluations by generating AEs with varying confidence levels (see Appendix~\ref{app:confidence_of_AEs}). Since both DREBIN and RAMDA use binary feature representations, we consider the $\ell_0$-norm (i.e., the number of changes) in eq.~\ref{eq:robust_optimization}, as it is the standard perturbation bound for binary features~\cite{carlini2019evaluating}.

To assess adversarial robustness, we evaluate models against bounded PK-Greedy and EvadeDroid attacks, ensuring their attack bounds match those used during AT. Considering these bounds allows for meaningful comparisons of defenses~\cite{nasr2025evaluating}. Moreover, it is crucial to verify that observed robustness against evasion attacks stems from AT. To this end, we subtract the baseline robustness of the vanilla model $\mathcal{M}_v$ from that of the hardened model $\mathcal{M}_h$ as follows:

\begin{equation}
\label{eq:real_robustness}
\text{$\mathcal{R}$}_{\text{AT}} = \mathcal{R}_h -\mathcal{R}_{v}
\end{equation}
% \vspace{-1.5em}

\noindent where $\mathcal{R}_h$ represents the robust accuracy of $\mathcal{M}_h$ in response to the $\epsilon$-bounded $\mathcal{A}ttack$, and $\mathcal{R}_{v}$ denotes the robust accuracy of $\mathcal{M}_v$ under the same $\epsilon$-bounded attack. While $\mathcal{R}_{\text{AT}}$ isolates the robustness directly attributed to AT, comparing its effectiveness across different settings (e.g., \textit{attack bounds}) requires the following normalization. We define the relative robustness gained by AT as:

% \begin{equation} \label{eq:relative_robustness} \mathcal{R}_{\text{rel}} = \frac{\mathcal{R}_{\text{AT}}}{100 - (\mathcal{R}_{v_1}-\mathcal{R}_{v_2})}\times 100 \end{equation}

\begin{equation} \label{eq:relative_robustness} \mathcal{R}_{\text{rel}} = \frac{\mathcal{R}_{\text{AT}}}{100 -\mathcal{R}_{v}}\times 100 \end{equation}

% \noindent where $100 - (\mathcal{R}{v_1} - \mathcal{R}{v_2})$ 

\noindent where $100 - \mathcal{R}{v}$ represents the portion of robustness that is still available for improvement by AT in $\mathcal{M}_h$ under the $\epsilon$-bounded $\mathcal{A}ttack$. Here, 100 represents the theoretical maximum robustness a model can achieve. This normalization ensures a fair comparison across different settings by measuring how effectively AT improves robustness relative to the remaining improvable portion.

\begin{figure}
    \centering
    \vspace{0.3em}
    \begin{subfigure}[b]{0.200\textwidth}
        \centering        \includegraphics[width=\textwidth]{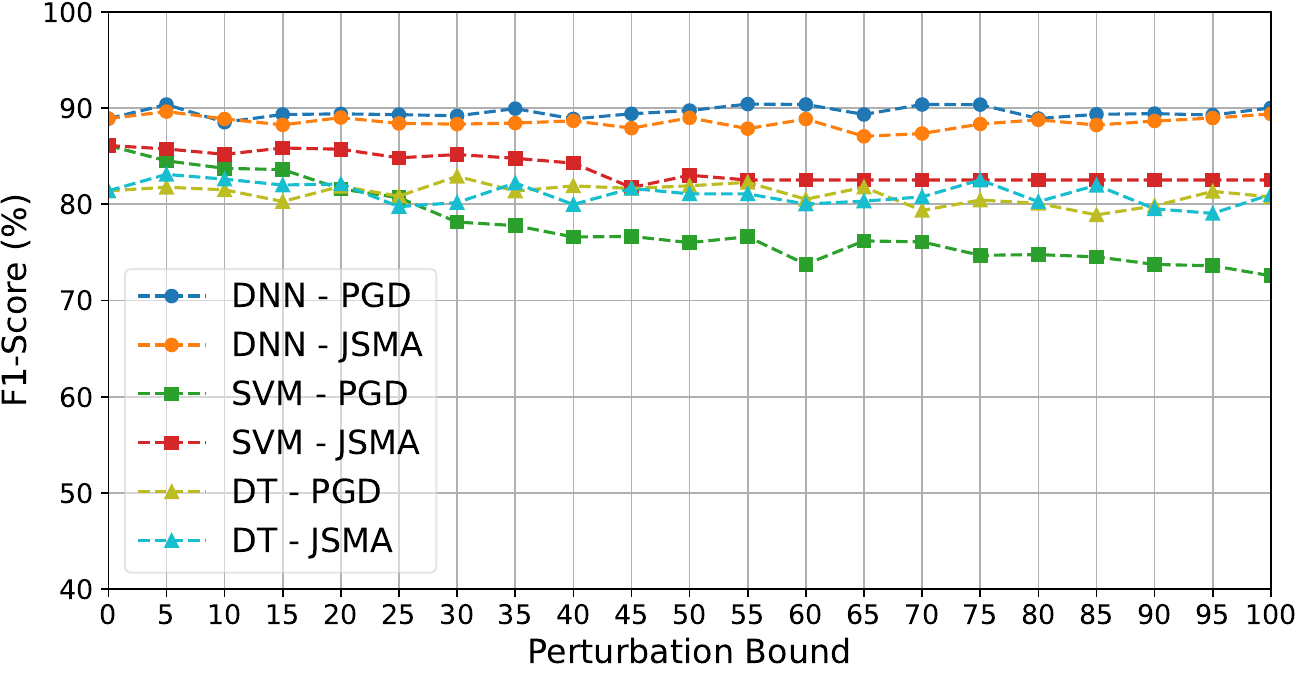} % Replace with your image file
        \vspace{-0.9em} % Adjust vertical space as needed
        \caption{DREBIN (DREBIN20)}
        \label{fig:image1}
    \end{subfigure}
    % \hfill
    \hspace{0.5em} 
    \begin{subfigure}[b]{0.200\textwidth}
        \centering
        \includegraphics[width=\textwidth]{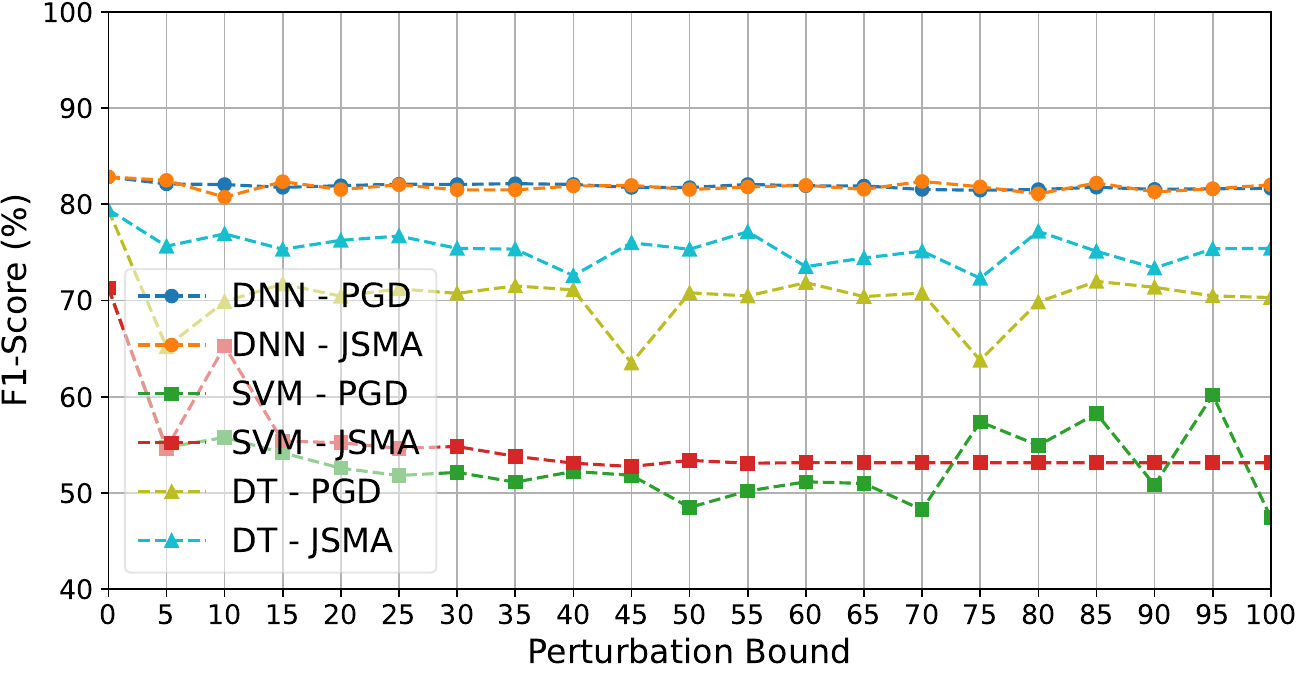} % Replace with your image file
        \vspace{-0.9em} % Adjust vertical space as needed
        \caption{RAMDA (DREBIN20)}
        \label{fig:image2}
    \end{subfigure}
     
    \vspace{0.3em}
    
    \begin{subfigure}[b]{0.200\textwidth}
        \centering
        \includegraphics[width=\textwidth]{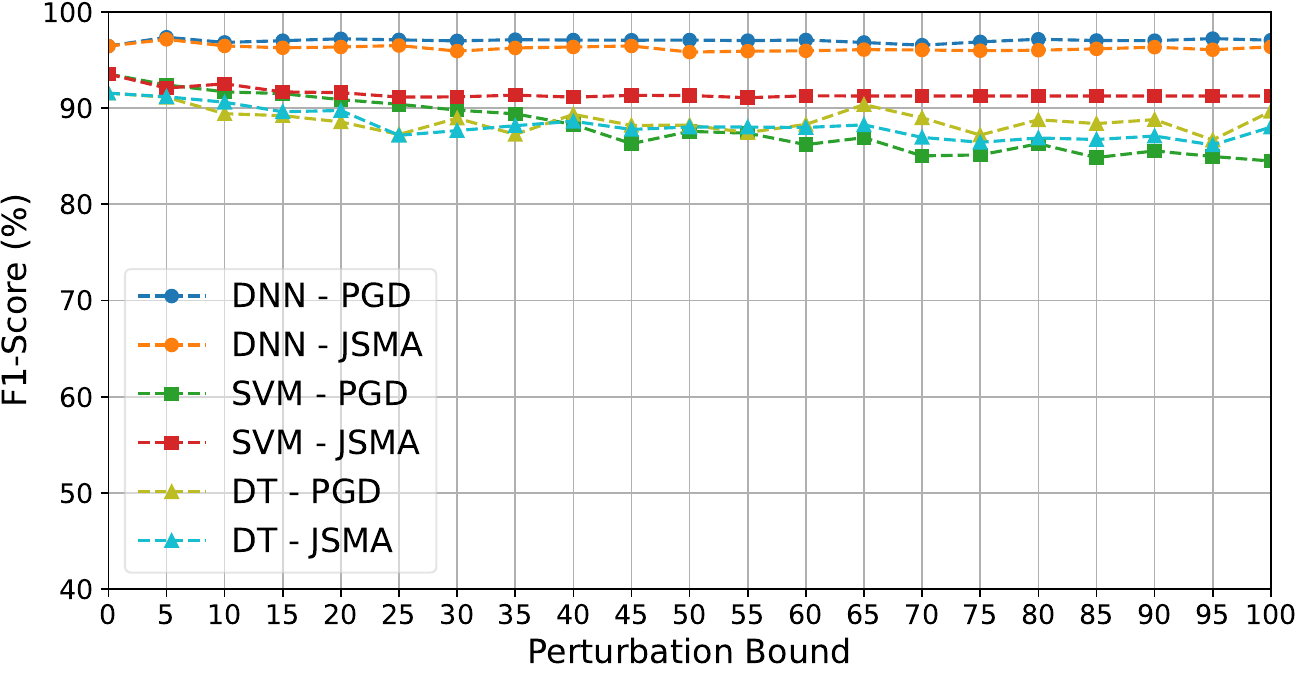} % Replace with your image file
        \vspace{-0.9em} % Adjust vertical space as needed
        \caption{DREBIN (APIGraph)}
        \label{fig:image1}
    \end{subfigure}
    % \hfill
    \hspace{0.5em} 
    \begin{subfigure}[b]{0.200\textwidth}
        \centering
        \includegraphics[width=\textwidth]{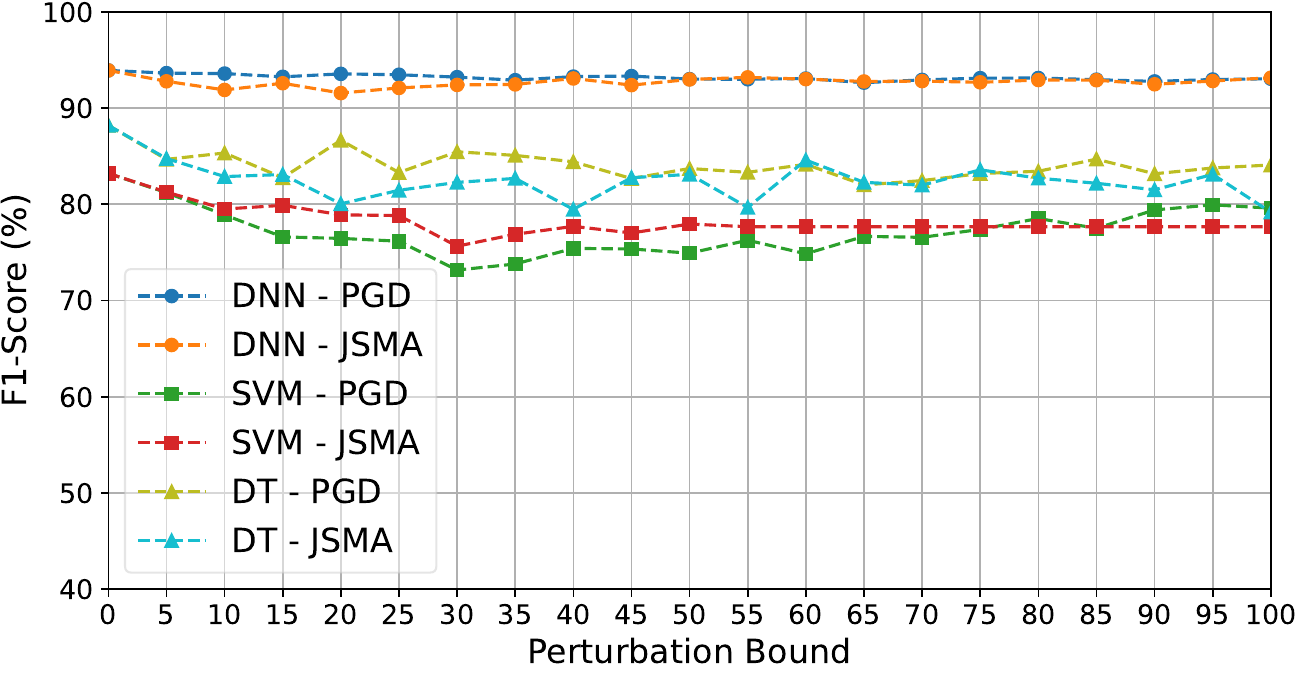} % Replace with your image file
        \vspace{-0.9em} % Adjust vertical space as needed
        \caption{RAMDA (APIGraph)}
        \label{fig:image2}
    \end{subfigure}
    \caption{Clean performance of various models trained on (a) the DREBIN and (b) RAMDA representations of the DREBIN20 dataset, and (c) the DREBIN and (d) RAMDA representations of the APIGraph dataset, measured in terms of F1 score. The models are strengthened using either PGD or JSMA with different perturbation bounds. The F-Scores of different vanilla models are displayed with a perturbation bound of 0.}
    \vspace{-1.5em}
    \label{fig:clean_acc_perturbation_budgets_vs_robustness}
\end{figure}

\subsubsection{Robust Optimization Settings: Variation of Perturbation Bounds and AE Confidence Levels}
\label{subsubsec:perturbation_bound_and_the_confidence_of_aes}
\begin{figure}[t!]
    \centering
    \vspace{0.3em} 
    \begin{subfigure}[b]{0.200\textwidth}
        \centering        \includegraphics[width=\textwidth]{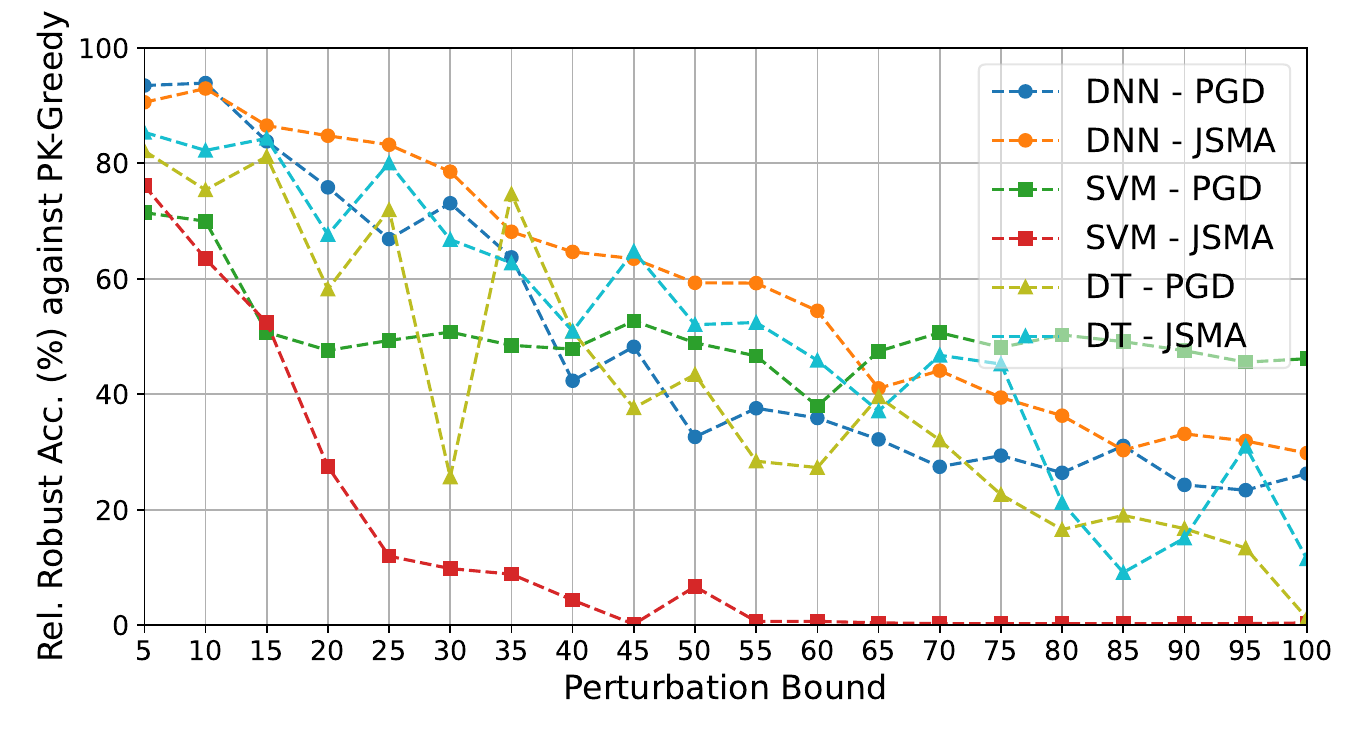} % Replace with your image file
        \vspace{-01.2em} % Adjust vertical space as needed
        \caption{DREBIN (DREBIN20)}
        \label{fig:image1}
    \end{subfigure}
    % \hfill
    \hspace{0.5em} 
    \begin{subfigure}[b]{0.200\textwidth}
        \centering
        \includegraphics[width=\textwidth]{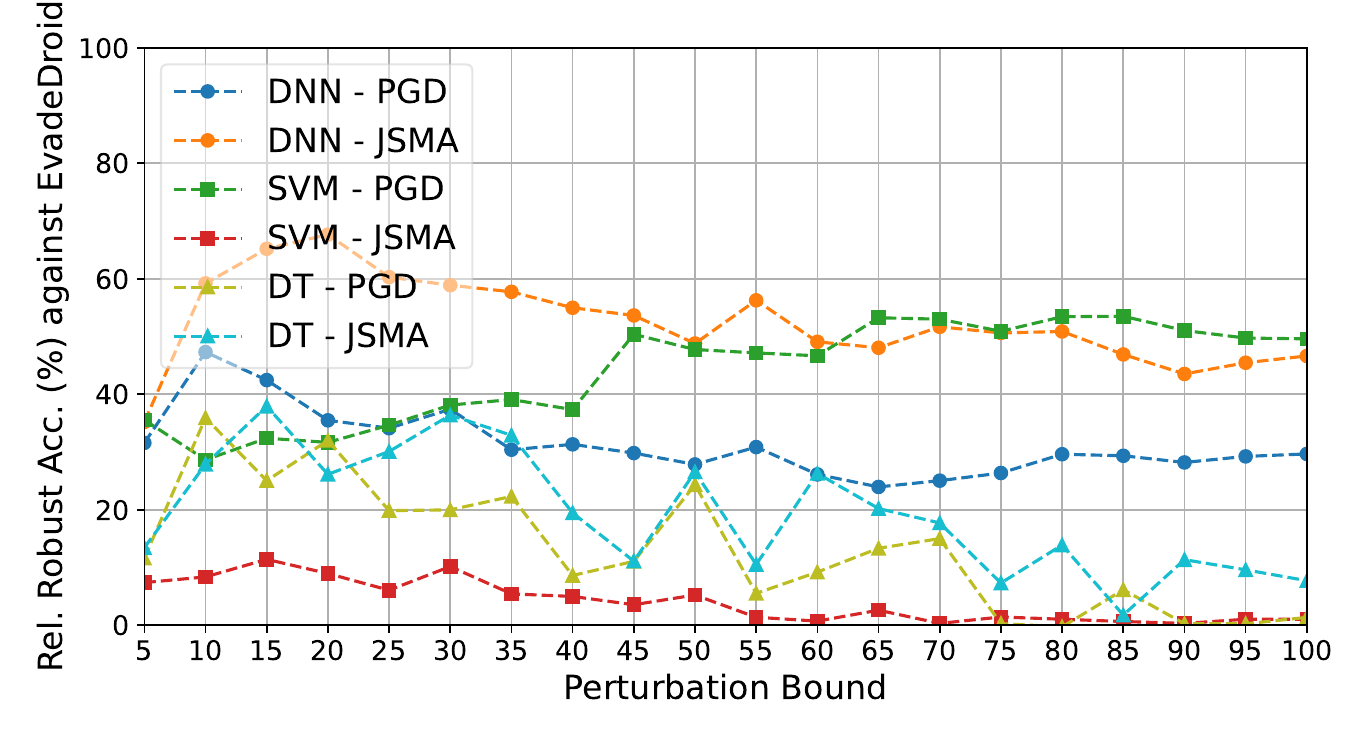} % Replace with your image file
        \vspace{-01.2em} % Adjust vertical space as needed
        \caption{DREBIN (DREBIN20)}
        \label{fig:image2}
    \end{subfigure}     
    \vspace{0.3em}    
    \begin{subfigure}[b]{0.200\textwidth}
        \centering
        \includegraphics[width=\textwidth]{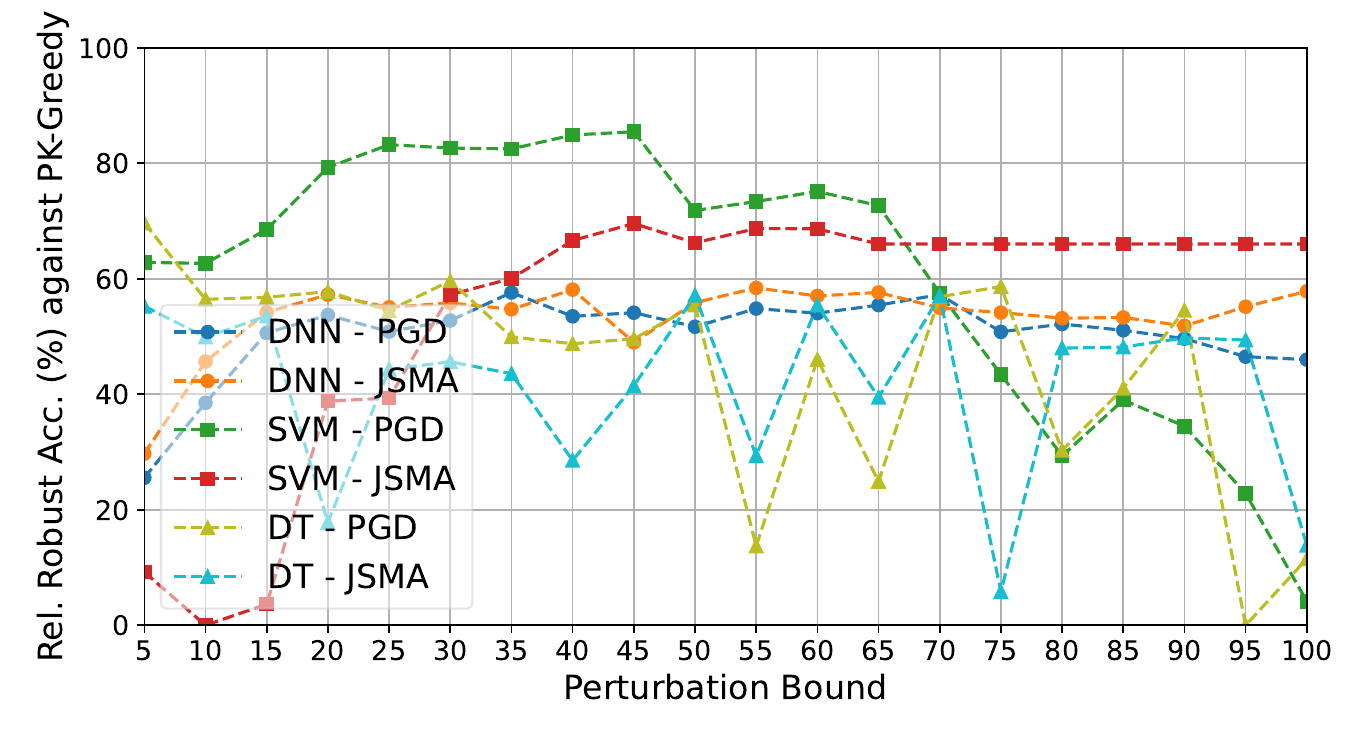} % Replace with your image file
        \vspace{-1.2em} % Adjust vertical space as needed
        \caption{RAMDA (DREBIN20)}
        \label{fig:image1}
    \end{subfigure}
    % \hfill
    \hspace{0.5em} 
    \begin{subfigure}[b]{0.200\textwidth}
        \centering
        \includegraphics[width=\textwidth]{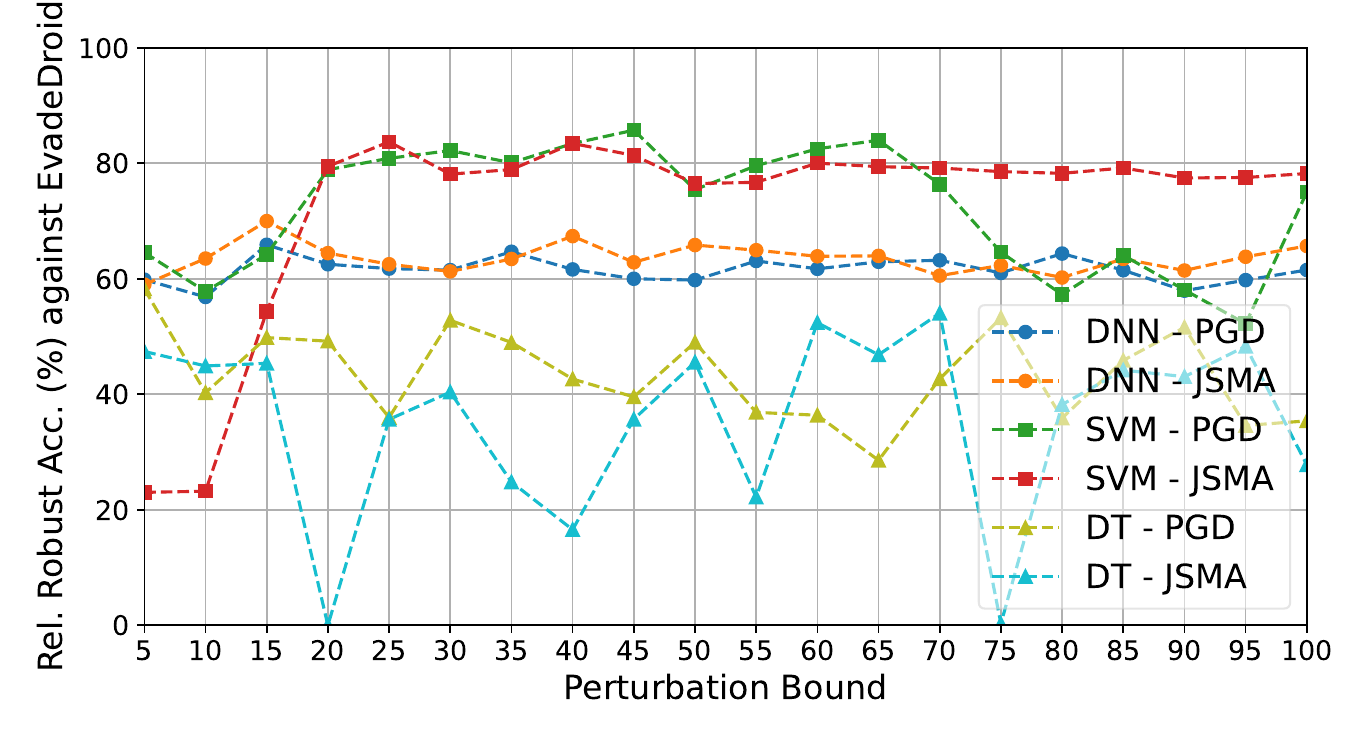} % Replace with your image file
        \vspace{-1.2em} % Adjust vertical space as needed
        \caption{RAMDA (DREBIN20)}
        \label{fig:image2}
    \end{subfigure}

    \begin{subfigure}[b]{0.200\textwidth}
        \centering        \includegraphics[width=\textwidth]{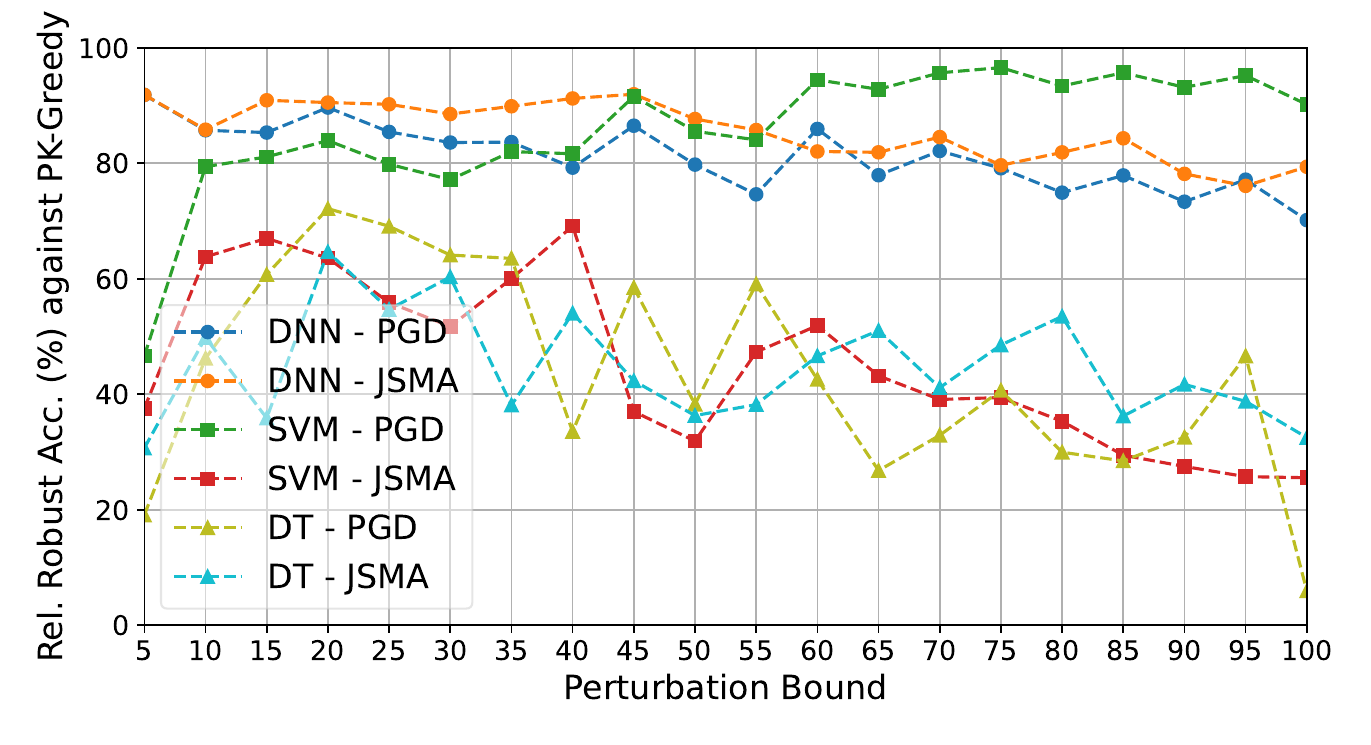} % Replace with your image file
        \vspace{-1.2em} % Adjust vertical space as needed
        \caption{DREBIN (APIGraph)}
        \label{fig:image1}
    \end{subfigure}
    % \hfill
    \hspace{0.5em} 
    \begin{subfigure}[b]{0.200\textwidth}
        \centering
        \includegraphics[width=\textwidth]{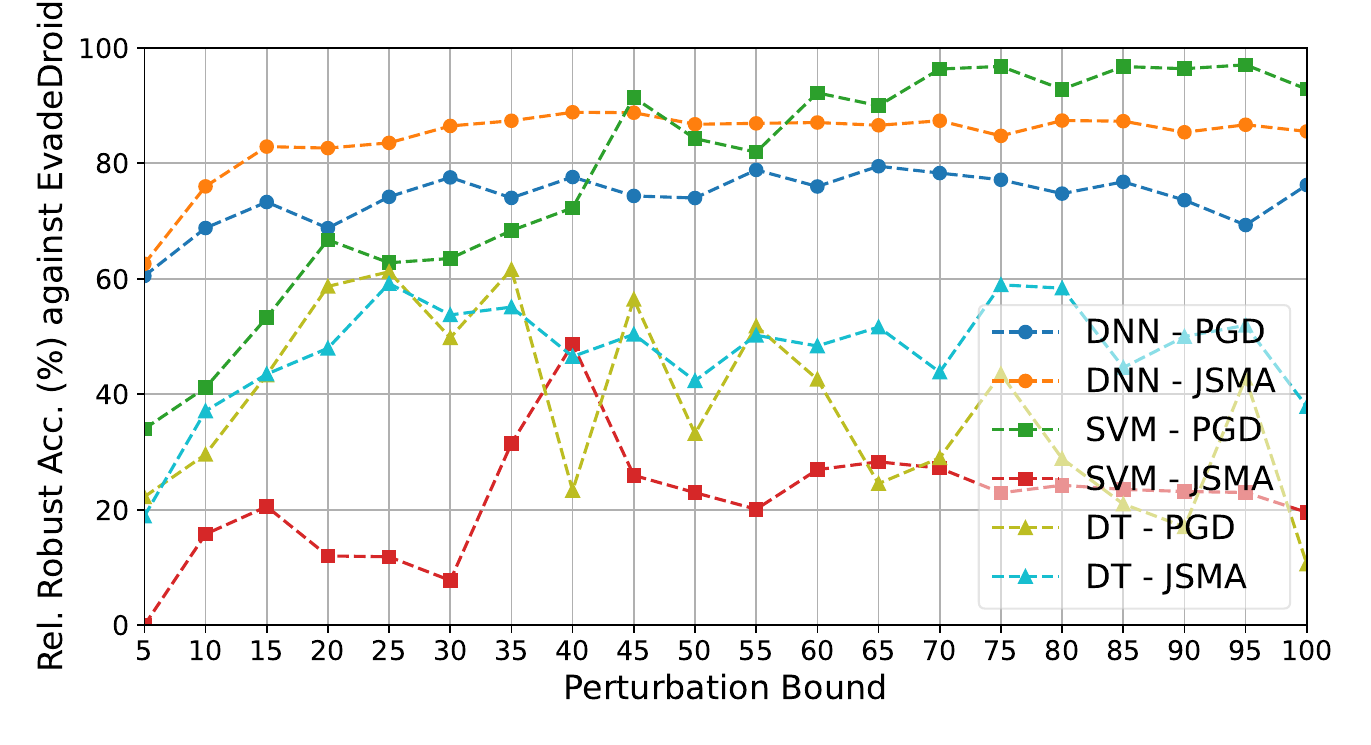} % Replace with your image file
        \vspace{-1.2em} % Adjust vertical space as needed
        \caption{DREBIN (APIGraph)}
        \label{fig:image2}
    \end{subfigure}     
    \vspace{0.3em}    
    \begin{subfigure}[b]{0.200\textwidth}
        \centering
        \includegraphics[width=\textwidth]{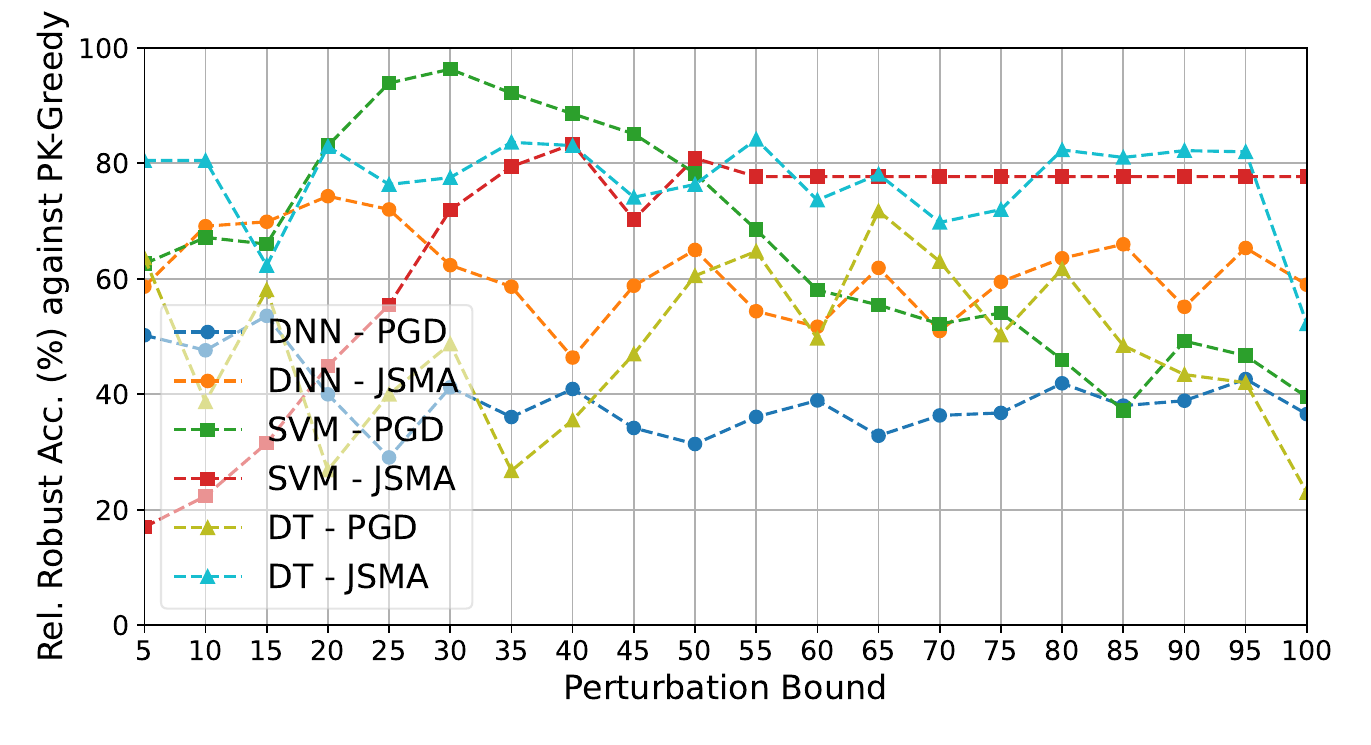} % Replace with your image file
        \vspace{-1.2em} % Adjust vertical space as needed
        \caption{RAMDA (APIGraph)}
        \label{fig:image1}
    \end{subfigure}
    % \hfill
    \hspace{0.5em} 
    \begin{subfigure}[b]{0.200\textwidth}
        \centering
        \includegraphics[width=\textwidth]{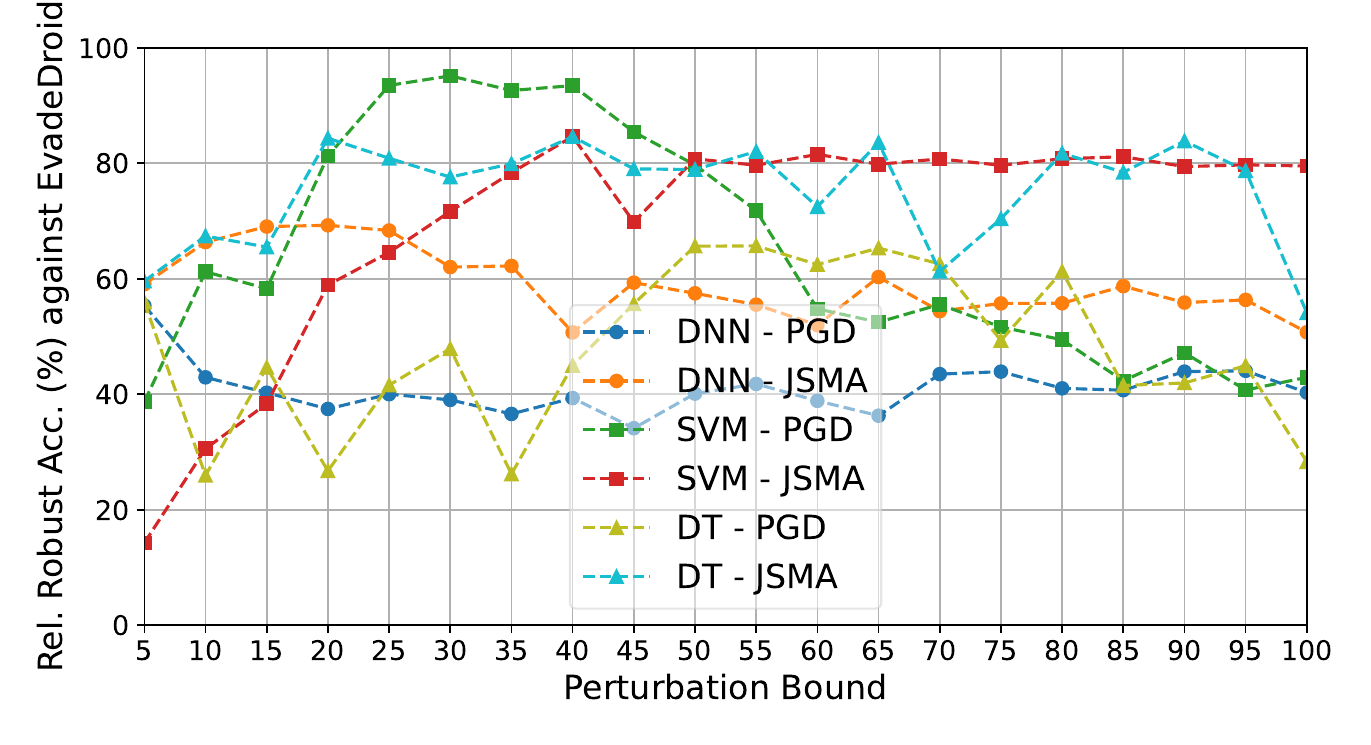} % Replace with your image file
        \vspace{-1.2em} % Adjust vertical space as needed
        \caption{RAMDA (APIGraph)}
        \label{fig:image2}
    \end{subfigure}  
    
    \caption{Relative robust accuracy, gained from AT, of hardened models trained on the DREBIN representation of the DREBIN20 dataset (a and b) and the APIGraph dataset (e and f), and the RAMDA representation of the DREBIN20 dataset (c and d) and the APIGraph dataset (g and h)
    % (a and b) DREBIN representation and (c and d) RAMDA representation of the DREBIN20 dataset 
    against PK-Greedy and EvadeDroid. The models are strengthened using either PGD or JSMA with different perturbation bounds.}
\label{fig:real_robust_acc_perturbation_budgets_vs_robustness_bounded_attack}
 \vspace{-1.0em}
\end{figure}

To investigate the impact of these two variables on AT, we utilized PGD and JSMA to generate AEs with perturbation bounds from 5 to 100 in increments of 5 and set the fraction of AEs to $\alpha = 50\%$. Figure~\ref{fig:clean_acc_perturbation_budgets_vs_robustness} shows the clean performance of different models via F1 score. As seen, classifiers exhibit varying sensitivities to AT with changing perturbation bound $\epsilon$ and AE confidence. Increasing $\epsilon$ can enable the inner optimization to find higher-confidence AEs, provided the evasion attack in AT can generate them. Figure~\ref{fig:clean_acc_perturbation_budgets_vs_robustness} shows that AT for hardening linear SVM and DT often compromises clean performance, with greater sacrifice as $\epsilon$ increases, especially for linear SVM with PGD. In contrast, DNN is more adaptable. For example, Figure~\ref{fig:clean_acc_perturbation_budgets_vs_robustness}-a shows in DREBIN (DREBIN20) the F1 Score of linear SVM drops from 86.1\% to 72.6\% with PGD-based AT at $\epsilon=100$. JSMA in AT affects clean performance less than PGD, particularly for linear SVM, as it may not generate high-confidence AEs even with higher $\epsilon$. For instance, Figure~\ref{fig:clean_acc_perturbation_budgets_vs_robustness}b shows that, while the F1 score of the linear SVM on RAMDA drops with JSMA at $\epsilon=5$, it remains stable at higher $\epsilon$ values because, as Figure~\ref{fig:conf_DNN_without} in Appendix~\ref{app:confidence_of_AEs} shows, the confidence of JSMA-generated AEs does not change significantly with increasing perturbation bounds.\\ \indent
% Figure~\ref{fig:real_robust_acc_perturbation_budgets_vs_robustness_bounded_attack} illustrates the relative robust accuracy of different models trained on DREBIN and RAMDA against PK-Greedy and EvadeDroid on DREBIN20 and APIGraph datasets. 
% Although Figure~\ref{fig:real_robust_acc_perturbation_budgets_vs_robustness_bounded_attack} shows variations in the adversarial robustness of models when hardened with different perturbation bounds, our observations reveal a few interesting results. First, models trained on RAMDA (dense, low-dimensional discrete feature space) often exhibit higher adversarial robustness compared to those trained on DREBIN (sparse, high-dimensional discrete feature space), as AT can potentially uncover more vulnerable regions in the lower-dimensional space. For example, Figure~\ref{fig:real_robust_acc_perturbation_budgets_vs_robustness_bounded_attack}-a shows that 12 out of 20 DNN-JSMA models trained on DREBIN achieve a robust accuracy greater than 50\% against PK-Greedy, whereas as can be seen in Figure~\ref{fig:real_robust_acc_perturbation_budgets_vs_robustness_bounded_attack}-c, 18 models achieve this benchmark when trained on RAMDA.
Figure~\ref{fig:real_robust_acc_perturbation_budgets_vs_robustness_bounded_attack} illustrates the relative robust accuracy of different models trained on DREBIN and RAMDA against PK-Greedy and EvadeDroid on DREBIN20 and APIGraph datasets. While the figure shows variations in adversarial robustness across perturbation bounds, our observations reveal several interesting results. First, models trained on RAMDA (dense, low-dimensional discrete feature space) often exhibit higher robustness than those trained on DREBIN (sparse, high-dimensional discrete feature space), as AT can uncover more vulnerable regions in the lower-dimensional space. For example, Figure~\ref{fig:real_robust_acc_perturbation_budgets_vs_robustness_bounded_attack}-a shows that 12 of 20 DNN-JSMA models trained on DREBIN achieve robust accuracy above 50\% against PK-Greedy, whereas Figure~\ref{fig:real_robust_acc_perturbation_budgets_vs_robustness_bounded_attack}-c shows 17 models achieve this benchmark when trained on RAMDA.\\ \indent
% Second, low-confidence AEs like those from JSMA often enhance the robustness of linear SVM on RAMDA as the perturbation bound increases (Figure~\ref{fig:real_robust_acc_perturbation_budgets_vs_robustness_bounded_attack}-c, d, g, and h). This is likely because, in dense, low-dimensional spaces, AEs within smaller $\epsilon$-bounded regions have limited impact on the decision boundary. Larger $\epsilon$ enables greater shifts, but at the cost of clean performance. A similar pattern is observed for linear SVM on DREBIN, a sparse, high-dimensional space, when high-confidence AEs like those from PGD are used in AT (Figure~\ref{fig:real_robust_acc_perturbation_budgets_vs_robustness_bounded_attack}-b, e, and f).
Second, low-confidence AEs, such as those from JSMA, often improve the robustness of linear SVM on RAMDA as the perturbation bound increases (Figure~\ref{fig:real_robust_acc_perturbation_budgets_vs_robustness_bounded_attack}-c, d, g, and h). In dense, low-dimensional spaces, AEs within smaller $\epsilon$-bounded regions have limited impact on the decision boundary. Therefore, larger $\epsilon$ allows greater shifts, but comes at the cost of clean performance. A similar trend occurs for linear SVM on DREBIN, a sparse, high-dimensional space, when high-confidence AEs like PGD are used in AT (Figure~\ref{fig:real_robust_acc_perturbation_budgets_vs_robustness_bounded_attack}-b, e, and f).\\ \indent
Thirdly, using high-confidence AEs in AT does not reliably enhance robustness. For example, Figure~\ref{fig:real_robust_acc_perturbation_budgets_vs_robustness_bounded_attack}-a to d reveals that among 240 models hardened with JSMA, 120 prove more robust against realistic attacks than those hardened with PGD. Similarly, Figure~\ref{fig:real_robust_acc_perturbation_budgets_vs_robustness_bounded_attack}-e to h shows that using JSMA (low-confidence AEs) to harden DNNs on both DREBIN and RAMDA representations often yields superior robustness. Conversely, for linear SVMs---especially on DREBIN---AT with high-confidence AEs generated by PGD enhances robustness more effectively than JSMA. This likely correlates with the linear decision boundary, which can be substantially shifted by high-confidence AEs, exposing larger vulnerable regions. However, as Figure~\ref{fig:clean_acc_perturbation_budgets_vs_robustness} illustrates, these significant adjustments incur a notable cost in clean performance. We also note that Appendix~\ref{app:big_perturbations} examines larger perturbation bounds for DREBIN, as current bounds may be insufficient for its high dimensionality. There, we observe that using very perturbation bounds in AT provides minimal to no benefit, or even a negative effect, across all models.

% \subsubsection{Robust Optimization Settings: Varying Fraction of AEs}
% \vspace{-0.5em}

\begin{figure}[b!]
        \centering
    \vspace{0.3em}
    \begin{subfigure}[b]{0.200\textwidth}
        \centering        \includegraphics[width=\textwidth]{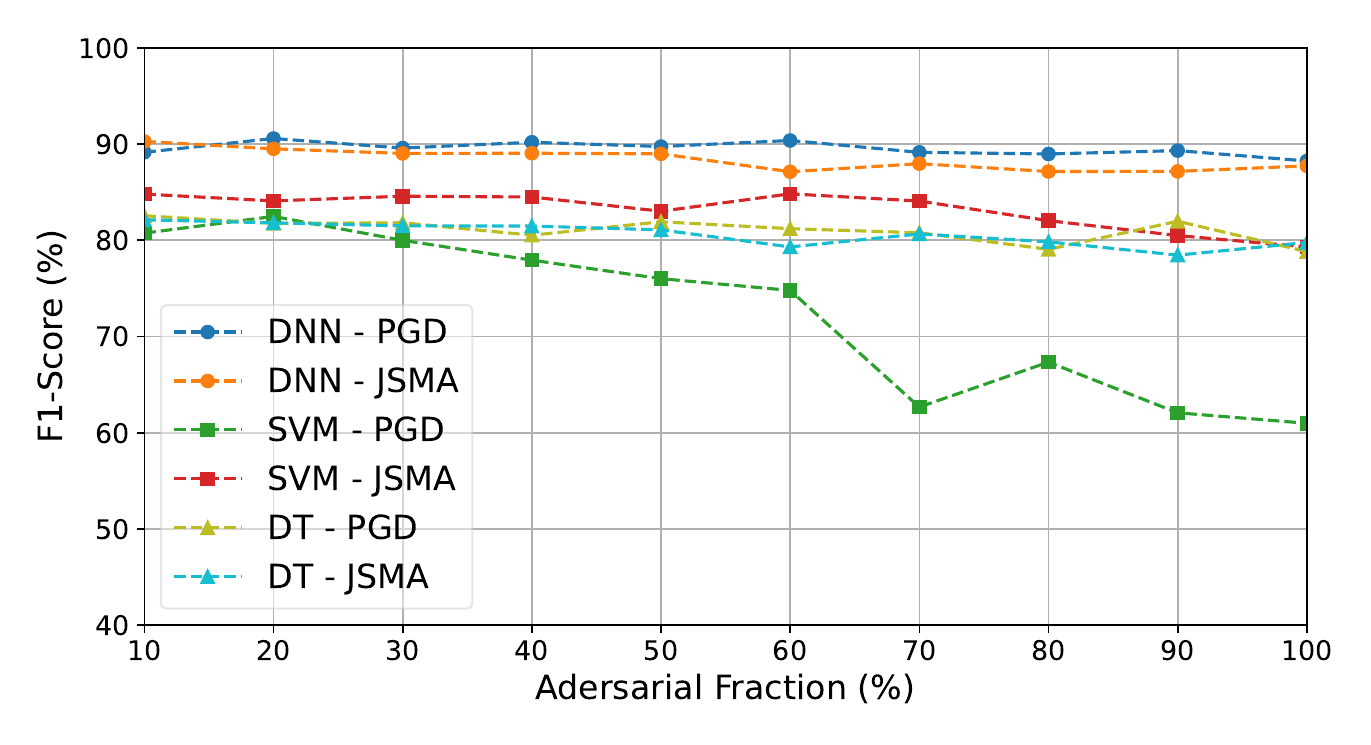} % Replace with your image file
        \vspace{-0.9em} % Adjust vertical space as needed
        \caption{DREBIN (DREBIN20)}
        \label{fig:image1}
    \end{subfigure}
    % \hfill
    \hspace{0.5em} 
    \begin{subfigure}[b]{0.200\textwidth}
        \centering
        \includegraphics[width=\textwidth]{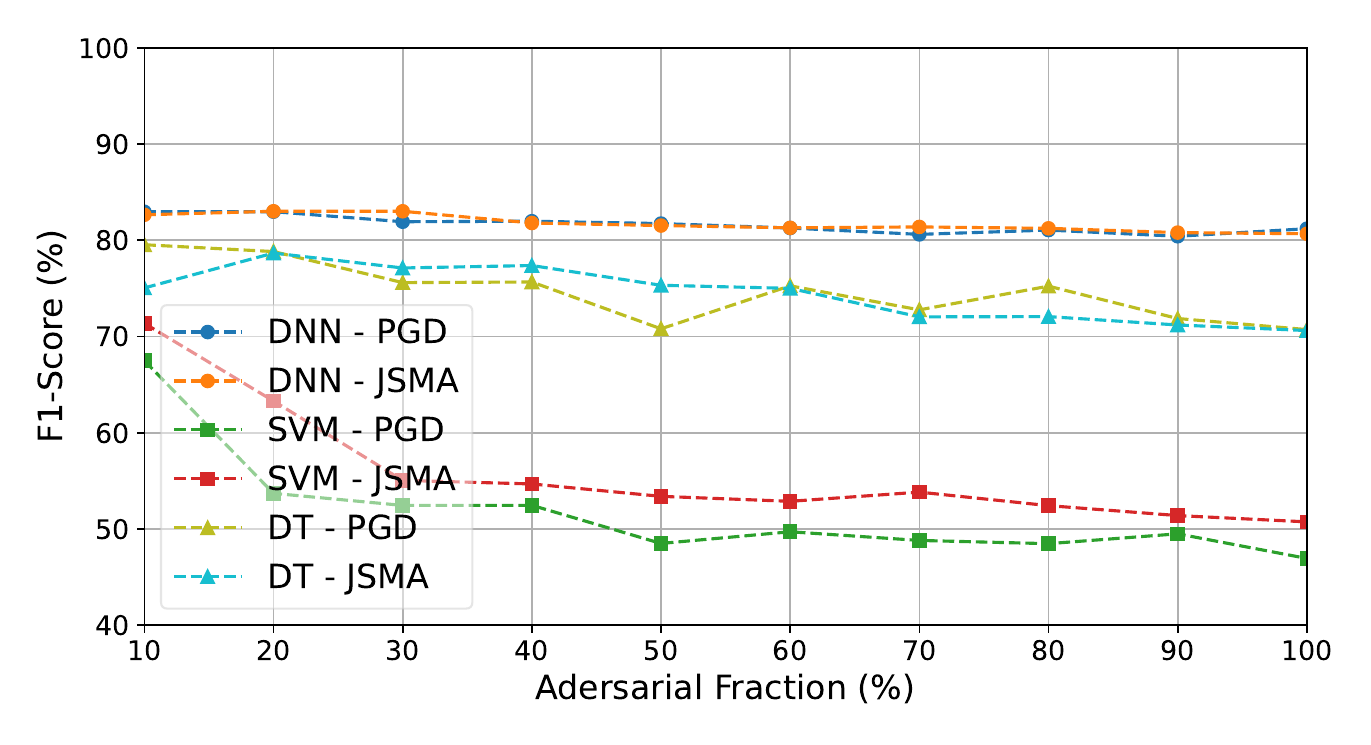} % Replace with your image file
        \vspace{-0.9em} % Adjust vertical space as needed
        \caption{RAMDA (DREBIN20)}
        \label{fig:image2}
    \end{subfigure}
     
    \vspace{0.3em}
    
    \begin{subfigure}[b]{0.200\textwidth}
        \centering
        \includegraphics[width=\textwidth]{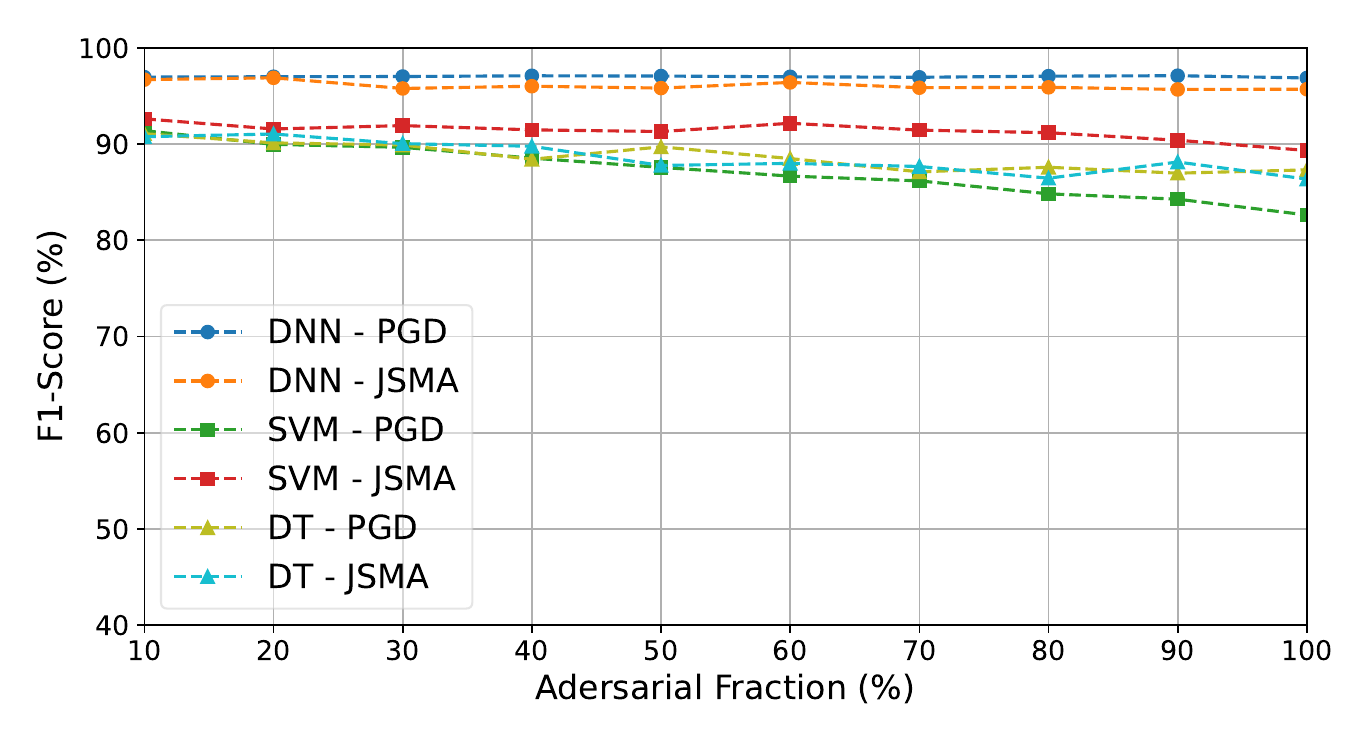} % Replace with your image file
        \vspace{-0.9em} % Adjust vertical space as needed
        \caption{DREBIN (APIGraph)}
        \label{fig:image1}
    \end{subfigure}
    % \hfill
    \hspace{0.5em} 
    \begin{subfigure}[b]{0.200\textwidth}
        \centering
        \includegraphics[width=\textwidth]{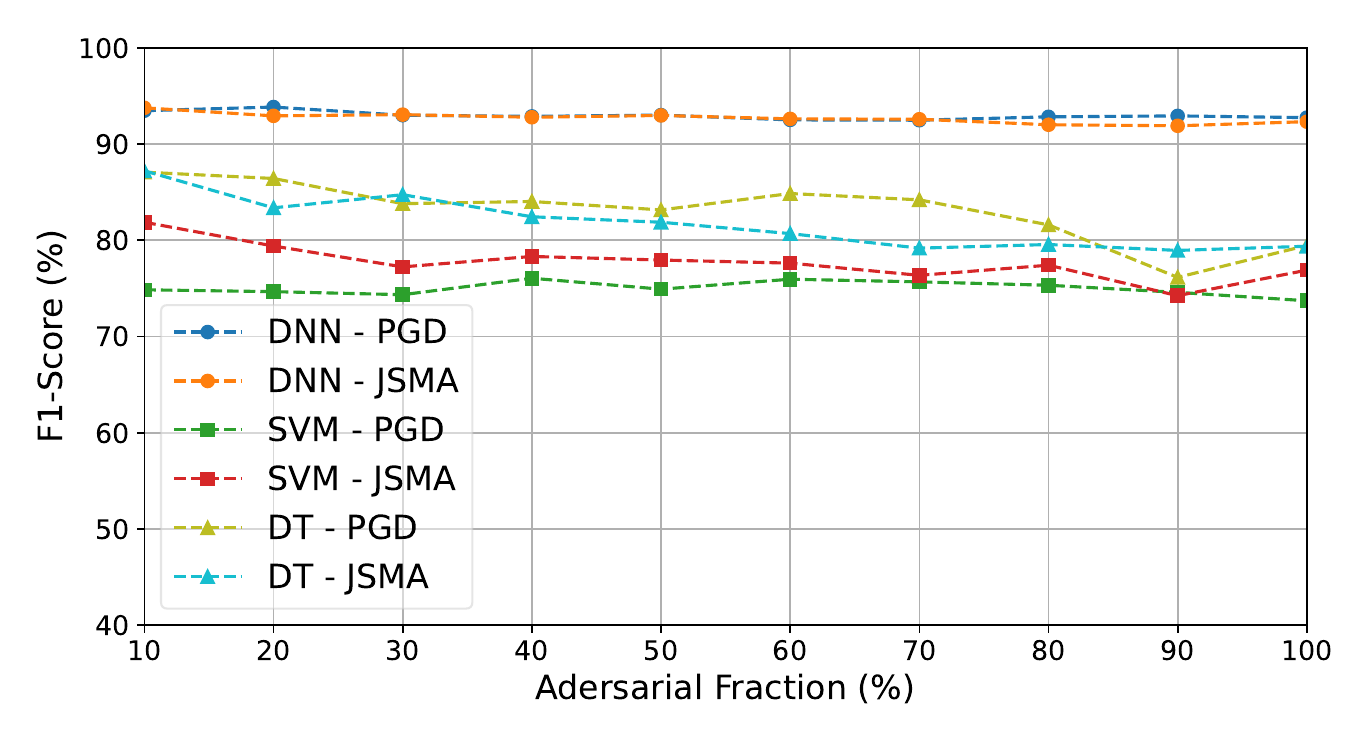} % Replace with your image file
        \vspace{-0.9em} % Adjust vertical space as needed
        \caption{RAMDA (APIGraph)}
        \label{fig:image2}
    \end{subfigure}  

    \caption{Clean performance of models trained on DREBIN and RAMDA representations of DREBIN20 and APIGraph datasets in terms of F1 score hardened with different fractions of AEs}
    \label{fig:clean_acc_ae_Rate}
\end{figure}

\subsubsection{Robust Optimization Settings: Variation of AE Fractions and AE Confidence Levels}
\label{subsubsec:varying_fraction_of_aes}
\begin{figure}[t!]
    \centering
    \vspace{0.3em}    
    \begin{subfigure}[b]{0.200\textwidth}
        \centering        \includegraphics[width=\textwidth]{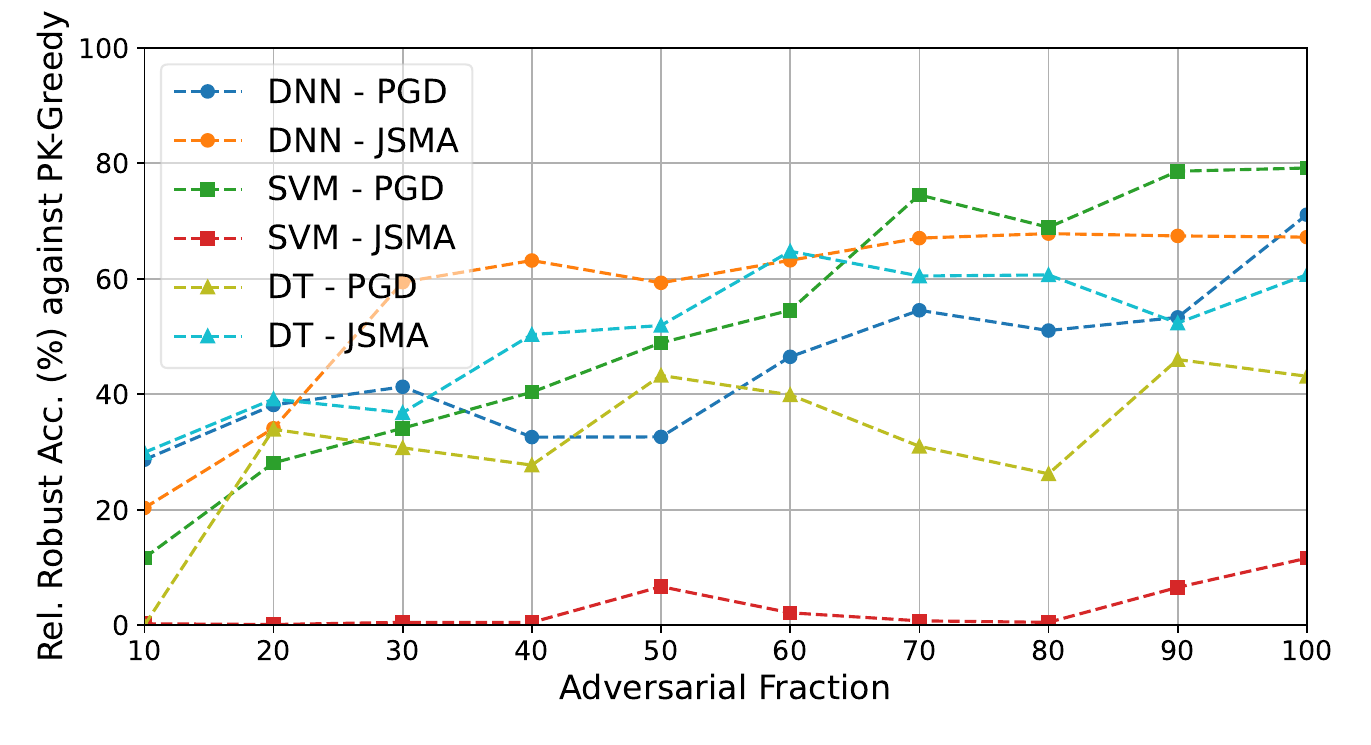} % Replace with your image file
        \vspace{-1.2em} % Adjust vertical space as needed
        \caption{DREBIN (DREBIN20)}
        \label{fig:image1}
    \end{subfigure}
    % \hfill
    \hspace{0.5em} 
    \begin{subfigure}[b]{0.200\textwidth}
        \centering
        \includegraphics[width=\textwidth]{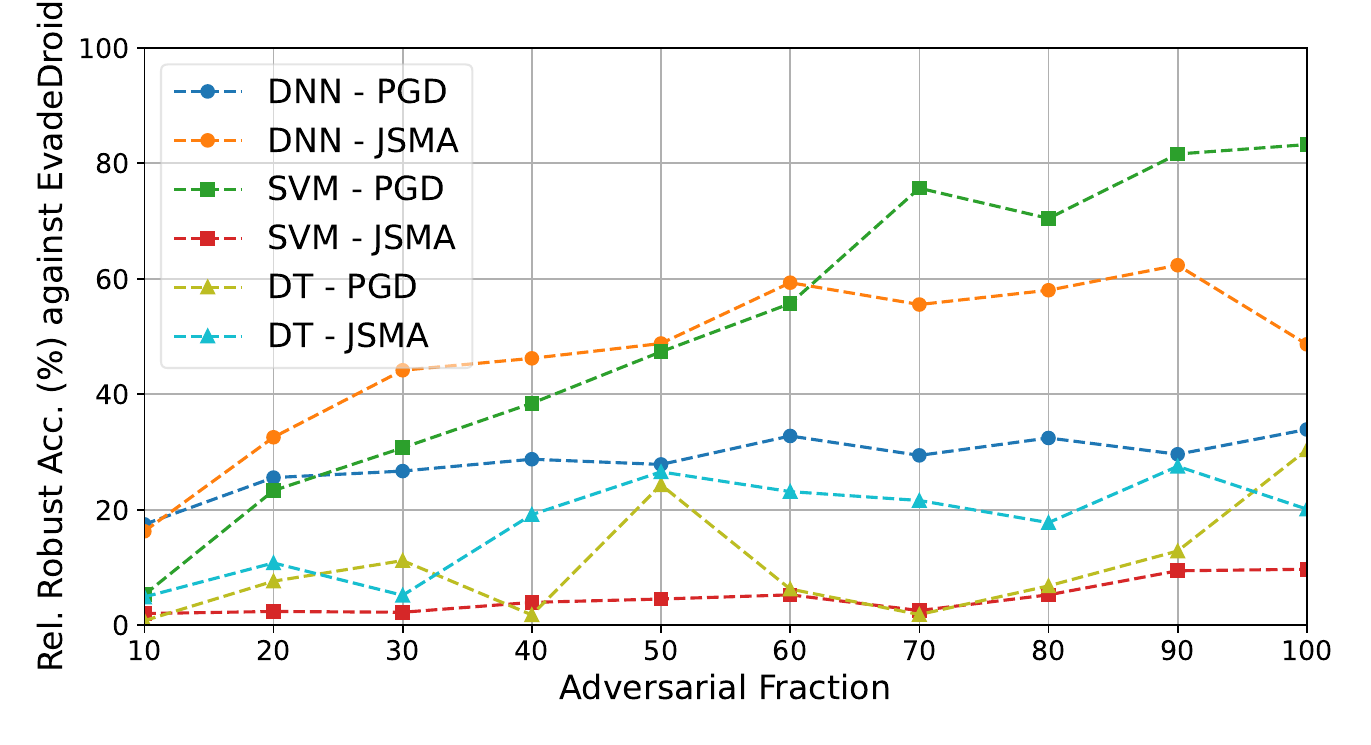} % Replace with your image file
        \vspace{-1.2em} % Adjust vertical space as needed
        \caption{DREBIN (DREBIN20)}
        \label{fig:image2}
    \end{subfigure}     
    \vspace{0.3em}    
    \begin{subfigure}[b]{0.200\textwidth}
        \centering
        \includegraphics[width=\textwidth]{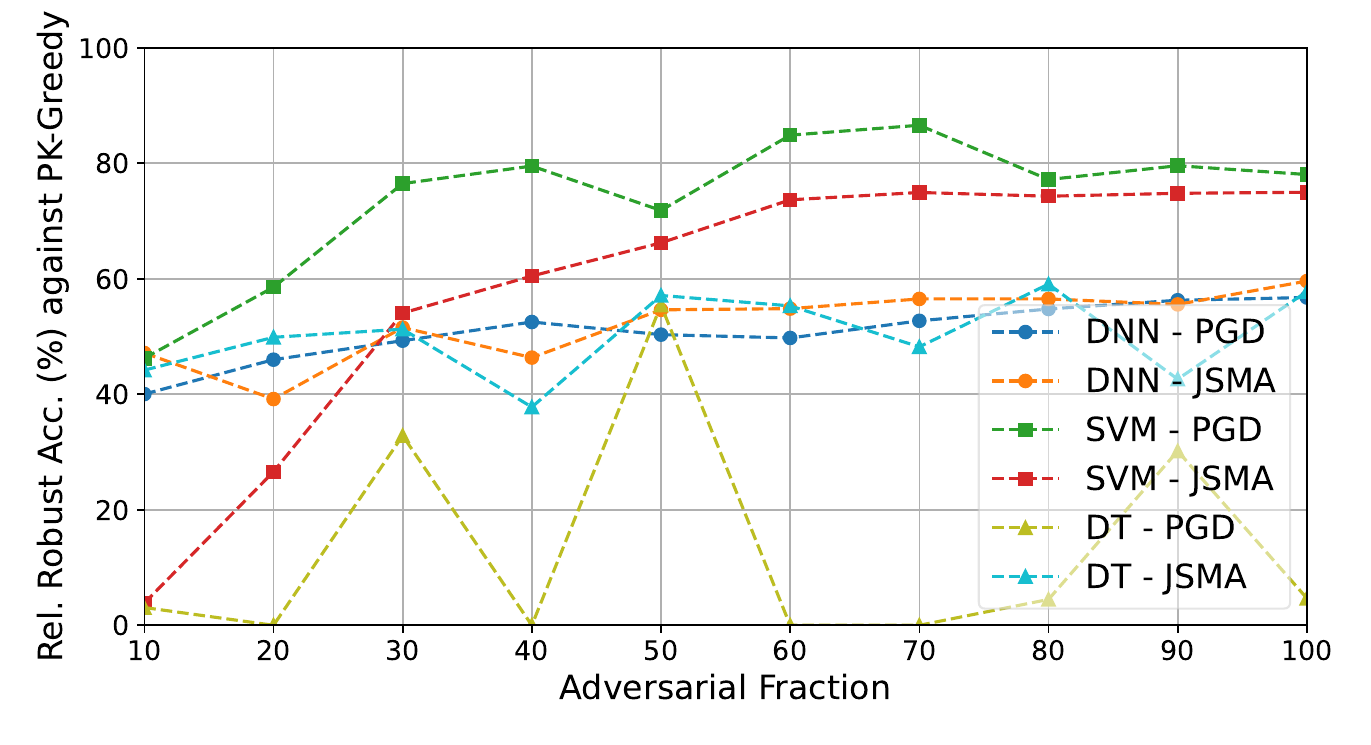} % Replace with your image file
        \vspace{-1.2em} % Adjust vertical space as needed
        \caption{RAMDA (DREBIN20)}
        \label{fig:image1}
    \end{subfigure}
    % \hfill
    \hspace{0.5em} 
    \begin{subfigure}[b]{0.200\textwidth}
        \centering
        \includegraphics[width=\textwidth]{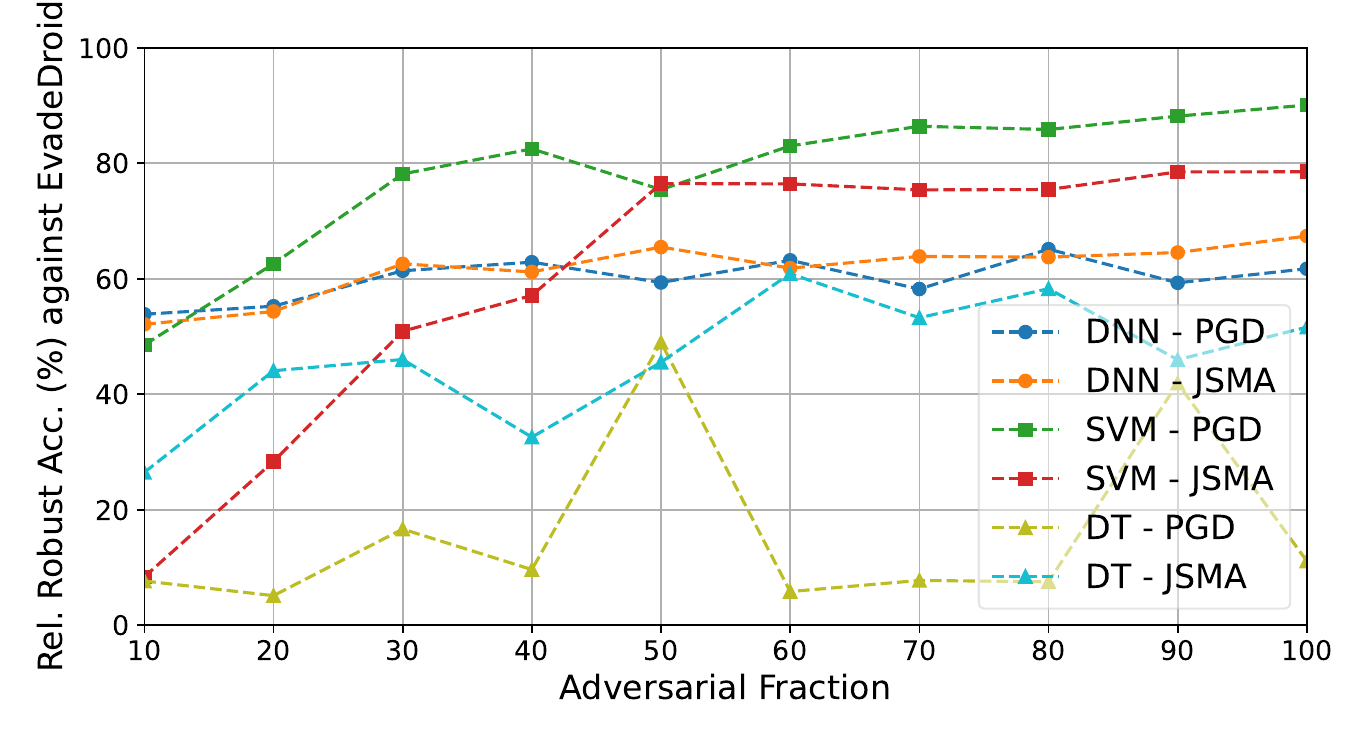} % Replace with your image file
        \vspace{-1.2em} % Adjust vertical space as needed
        \caption{RAMDA (DREBIN20)}
        \label{fig:image2}
    \end{subfigure} 
    \vspace{0.3em}    
     \begin{subfigure}[b]{0.200\textwidth}
        \centering        \includegraphics[width=\textwidth]{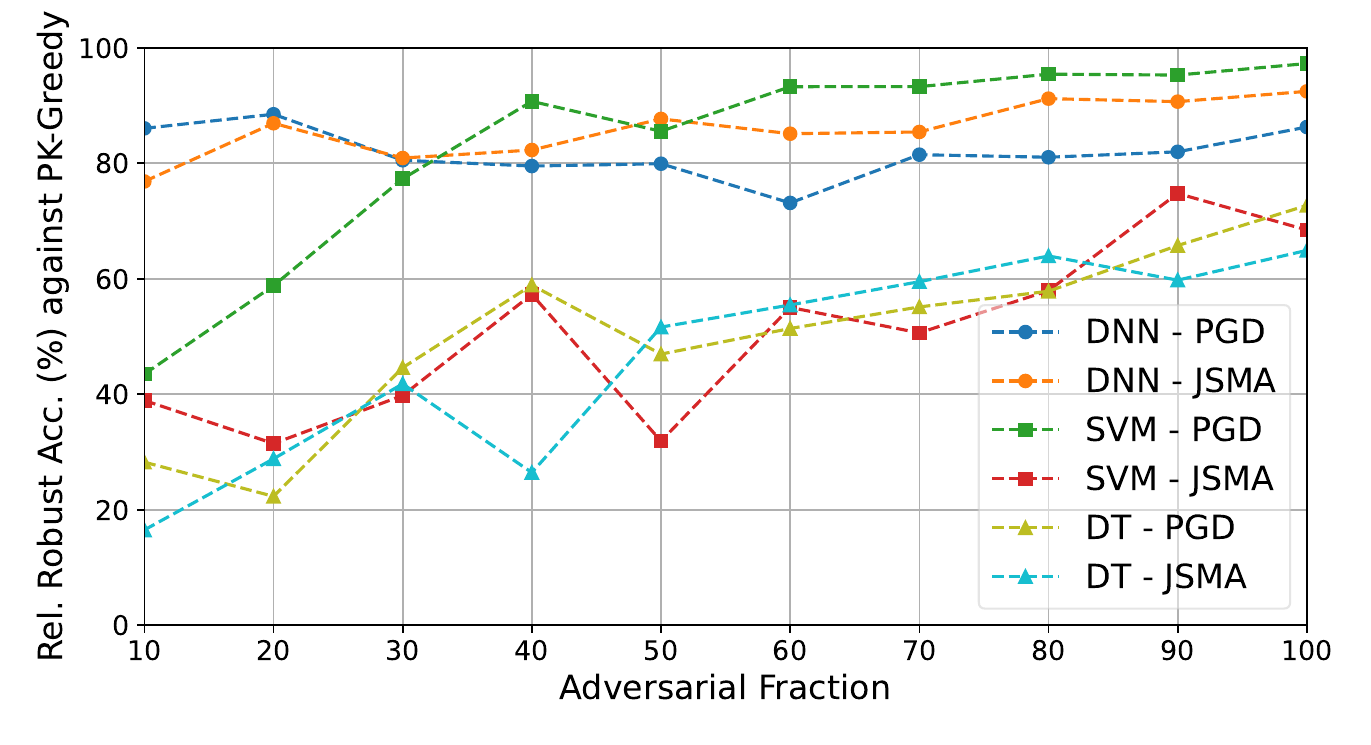} % Replace with your image file
        \vspace{-1.2em} % Adjust vertical space as needed
        \caption{DREBIN (APIGraph)}
        \label{fig:image1}
    \end{subfigure}
    % \hfill
    \hspace{0.5em} 
    \begin{subfigure}[b]{0.200\textwidth}
        \centering
        \includegraphics[width=\textwidth]{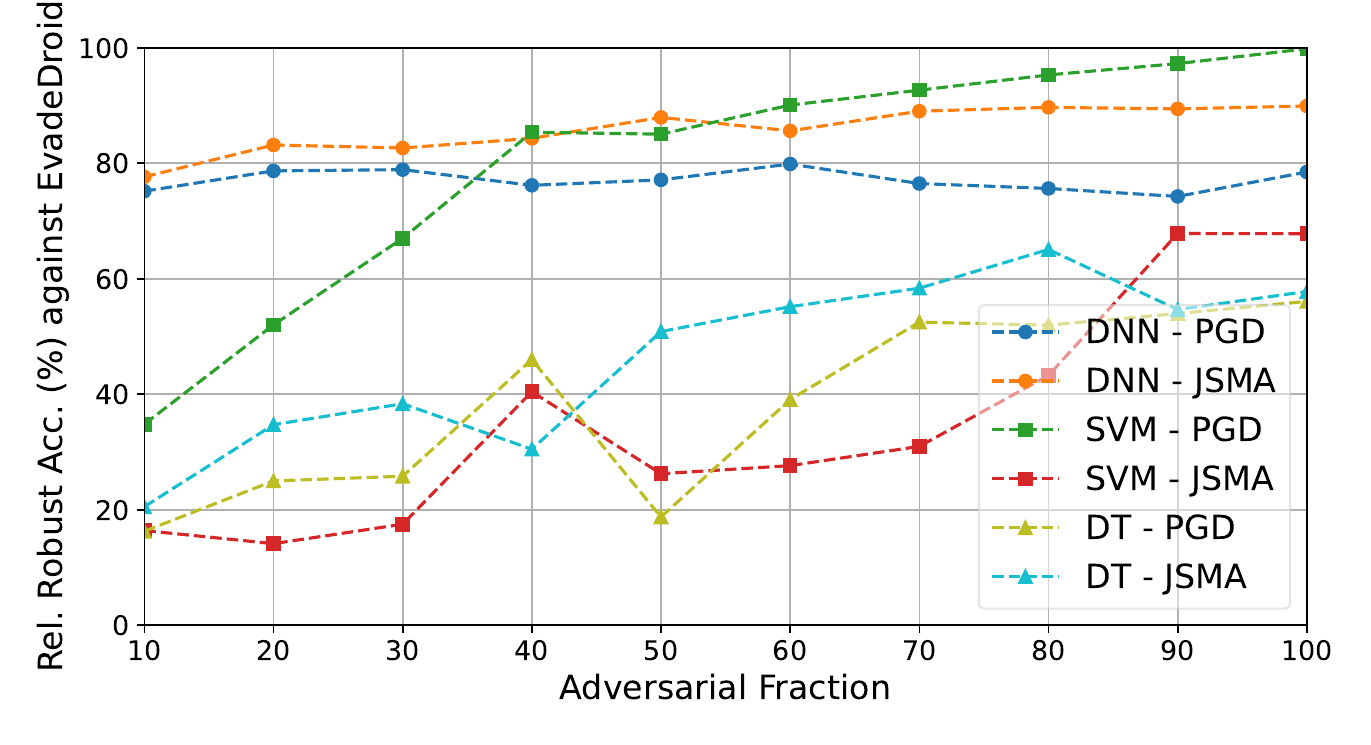} % Replace with your image file
        \vspace{-1.2em} % Adjust vertical space as needed
        \caption{DREBIN (APIGraph)}
        \label{fig:image2}
    \end{subfigure}     
    \vspace{0.3em}    
    \begin{subfigure}[b]{0.200\textwidth}
        \centering
        \includegraphics[width=\textwidth]{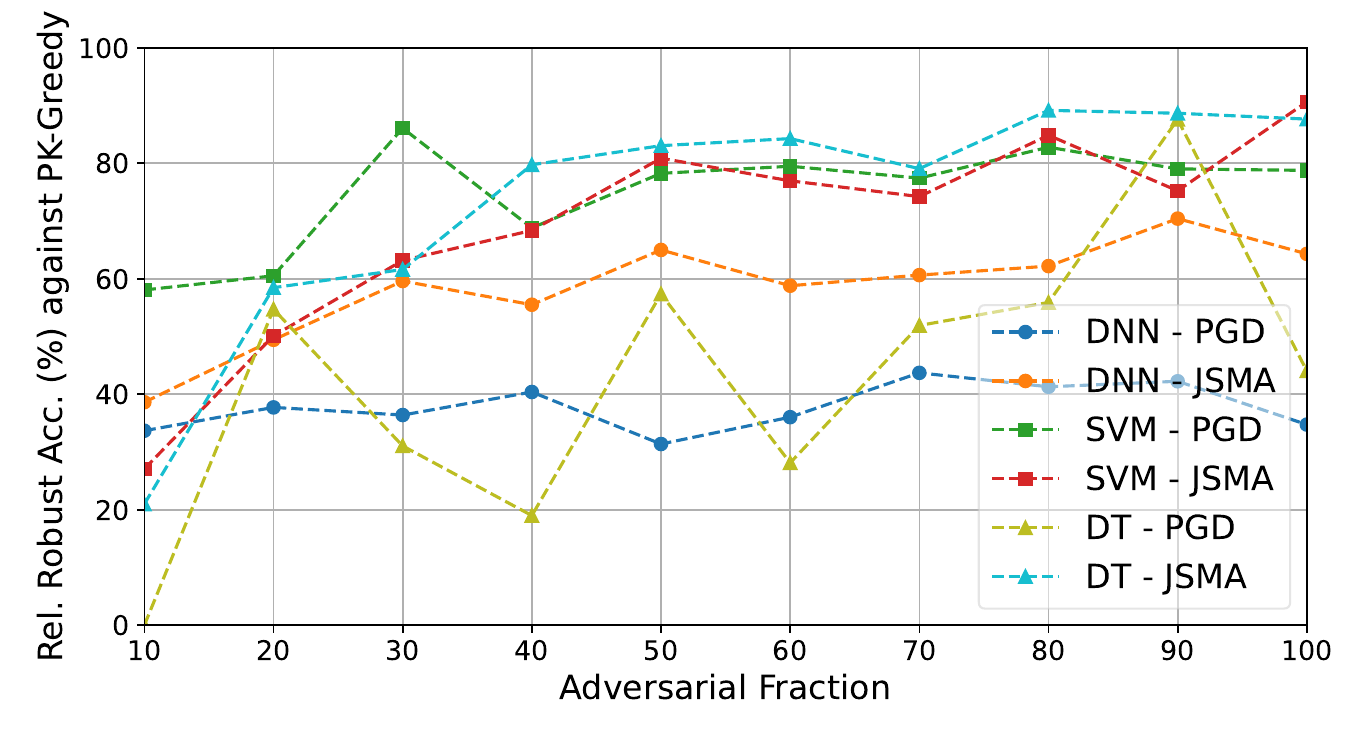} % Replace with your image file
        \vspace{-1.2em} % Adjust vertical space as needed
        \caption{RAMDA (APIGraph)}
        \label{fig:image1}
    \end{subfigure}
    % \hfill
    \hspace{0.5em} 
    \begin{subfigure}[b]{0.200\textwidth}
        \centering
        \includegraphics[width=\textwidth]{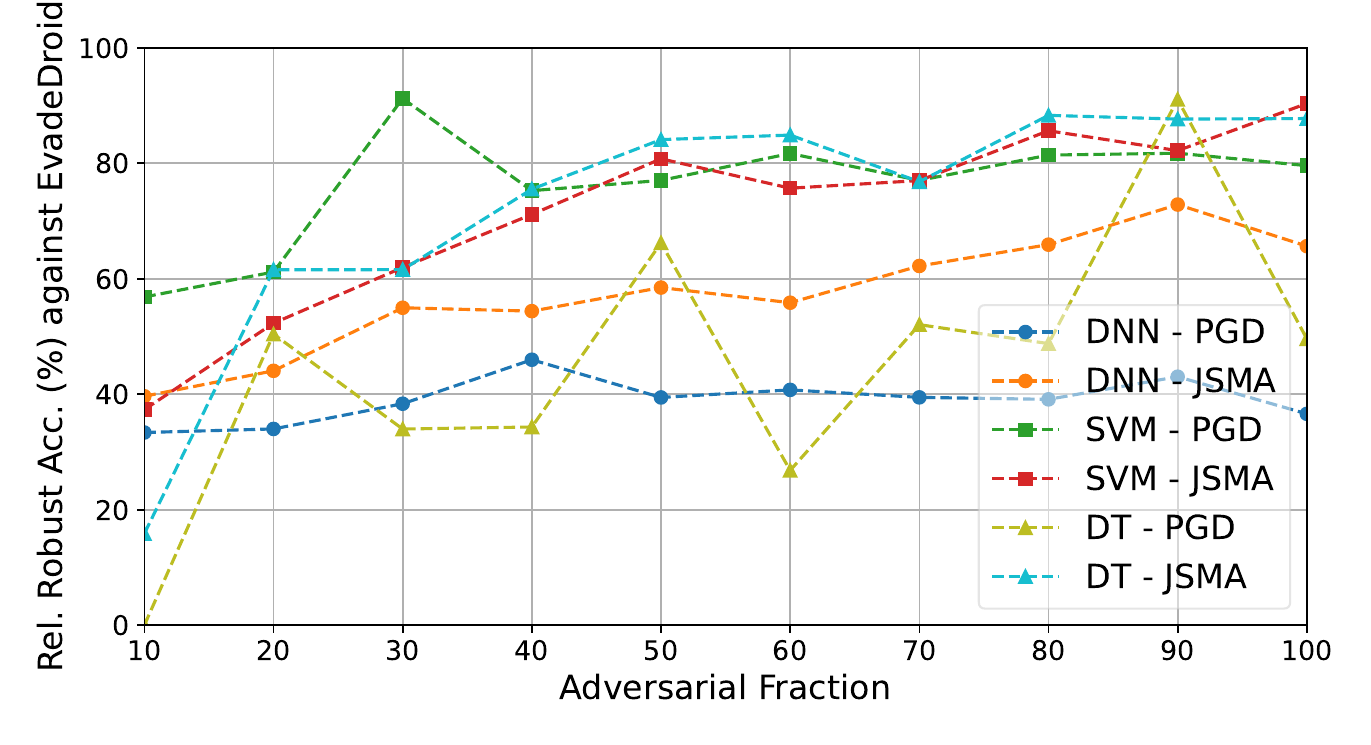} % Replace with your image file
        \vspace{-1.2em} % Adjust vertical space as needed
        \caption{RAMDA (APIGraph)}
        \label{fig:image2}
    \end{subfigure}   
    
    \caption{Relative robust accuracy gained from AT for hardened models trained on (a, b) DREBIN and (c, d) RAMDA representations of DREBIN20, and (e, f) DREBIN and (g, h) RAMDA representations of APIGraph against PK-Greedy and EvadeDroid, under varying AE fractions during training.}
    %\vspace{-1.5em}
    \label{fig:real_robust_acc_ae_rate}
\end{figure}

While increasing the number of AEs can uncover more blind spots when malware samples are uniformly distributed near the decision boundary, it may also impair clean performance. This experiment evaluates the impact of the AE fraction in AT (varied from 10\% to 100\% in 10\% increments, with a fixed $\epsilon=50$) on both clean performance and robust accuracy, following a design similar to \S\ref{subsubsec:perturbation_bound_and_the_confidence_of_aes}. The clean performance shown in Figure~\ref{fig:clean_acc_ae_Rate} and the relative robust accuracy depicted in Figure~\ref{fig:real_robust_acc_ae_rate} again highlight the classifier's pivotal role. Non-linear models like DNNs adapt their boundaries to AEs with minimal loss in clean accuracy. Conversely, the linear SVM shows a marked increase in robust accuracy---exceeding 60\% in the DREBIN representation, especially when hardened by PGD---but this gain is offset by a significant drop in clean performance as the AE fraction increases. Notably, this exploration suggests that the low robustness of non-linear models under large perturbation bounds, identified in \S\ref{subsubsec:perturbation_bound_and_the_confidence_of_aes}, can be mitigated by using more AEs during training, as DNN robustness often improves with a higher adversarial fraction.

% \vspace{0.1em}
\subsubsection{Robust Optimization Settings: Domain Constraints}
\label{subsubsec:domain_constraints}

\begin{table}[b!]
\footnotesize
\caption{Parameter settings for domain constraints exploration in terms of adversarial rate $\alpha$ and perturbation bound $\epsilon$. PG and ED define PK-Greedy and EvadeDroid, respectively.}
\label{table:domain_constraints_parameters}
\begin{center}
\begin{tabular}{l|c|rr|rr}
  \toprule
  \multirow{2}{*}{\textbf{Model}} 
      & \multirow{2}{*}{\textbf{$\alpha$}} 
          & \multicolumn{2}{c|}{\textbf{$\epsilon$ (DREBIN)}} & \multicolumn{2}{c}{\textbf{$\epsilon$ (RAMDA)}}\\
  \cline{3-6}
  & & \textbf{PG} & \textbf{ED} & \textbf{PG} & \textbf{ED} \\  
  \midrule
    DNN & 50 & 50 & 80 & 90 & 50\\
    SVM & 50 & 100 & 65 & 25 & 65\\
    DT & 50 & 70 & 85 & 75 & 30\\    
  \bottomrule
\end{tabular}
\end{center}
% \vspace{-1.5em}
\end{table}

% The domain constraints can guide AT to focus on regions vulnerable to realistic evasion attacks. To examine the impact of these constraints on AT's success, we used realizable AEs generated by bounded PK-Greedy and bounded EvadeDroid in our AT process. Since employing problem-space adversarial attacks for robust optimization to solve the inner maximization problem significantly increases training time, we only consider the DREBIN20 dataset in our evaluations. Table~\ref{table:domain_constraints_parameters} shows the settings for the robust optimization considered in this experiment. Our criterion for selecting the perturbation bound $\epsilon$ is to ensure that PK-Greedy and EvadeDroid achieve maximum success rates in fooling the evaluated models.
To guide AT toward regions vulnerable to realistic evasion attacks, we incorporated domain constraints using realizable AEs from bounded PK-Greedy and EvadeDroid. However, because robust optimization with problem-space attacks is computationally expensive, we limited this evaluation to the DREBIN20 dataset. The perturbation bound ($\epsilon$) was chosen to maximize the attack success rate of both PK-Greedy and EvadeDroid, and the specific robust optimization settings are detailed in Table~\ref{table:domain_constraints_parameters}.

\begin{table}[t]
\footnotesize
\caption{ Clean performance (F1 Score \%) of models trained on DREBIN and RAMDA representations of DREBIN20, hardened with unrealistic or realistic evasion attacks (PG = PK-Greedy, ED = EvadeDroid). F1 scores for vanilla DNN, SVM, and DT on DREBIN are 88.9, 86.1, and 81.4, respectively, and on RAMDA are 82.8, 71.3, and 79.5, respectively.
% Clean performance of different models trained on DREBIN and RAMDA representations of DREBIN20, hardened with either unrealistic or realistic evasion attacks in terms of F1 Score (\%). PG and ED define PK-Greedy and EvadeDroid, respectively. F1 scores of vanilla 
%  DNN, SVM, and DT trained on DREBIN representation are 88.9, 86.1, and 81.4, respectively. F1 scores of vanilla 
%  DNN, SVM, and DT trained on RAMDA representation are 82.8, 71.3, and 79.5, respectively.
 }
\label{table:clean_acc_domain_constraints}
% \begin{tabular}{p{2.2cm}|p{0.5cm}p{0.5cm}|p{1cm}p{1cm}}
\begin{center}
\resizebox{\columnwidth}{!}{
\begin{tabular}{l|rrrr|rrrr}
  \toprule
  \multirow{2}{*}{\textbf{Model}} 
      & \multicolumn{4}{c|}{\textbf{DREBIN}} 
          & \multicolumn{4}{c}{\textbf{RAMDA}} \\  \cline{2-9}
  & $\epsilon$ & \textbf{PGD} & \textbf{JSMA} & \textbf{PG} & $\epsilon$ & \textbf{PGD} & \textbf{JSMA} & \textbf{PG} \\  \midrule
    DNN & 50 & 89.7 & 89.0 & 89.1 & 90 & 81.5 & 81.3 & 82.6\\
    SVM & 100 & 72.6 & 82.5 & 85.1 & 25 & 51.8 & 54.6 & 70.9\\
    DT & 70 & 79.4 & 80.8 & 82.1 & 75 & 63.8 & 72.3 & 77.5\\
    \midrule
    & $\epsilon$ & \textbf{PGD} & \textbf{JSMA} & \textbf{ED} & $\epsilon$ & \textbf{PGD} & \textbf{JSMA} & \textbf{ED} \\  \midrule
    DNN & 80 & 88.9 & 88.8 & 90.4 & 50 & 81.7 & 81.5 & 82.3\\
    SVM & 65 & 76.2 & 82.5 & 85.5 & 65 & 51.0 & 53.1 & 71.2\\
    DT & 85 & 78.9 & 82.0 & 83.6 & 30 & 70.7 & 75.4 & 78.6\\
  \bottomrule
\end{tabular}
}
\end{center}
\vspace{-1.5em}
\end{table}

% Figure~\ref{fig:domain_constraints} illustrates the robust accuracy of various malware classifiers hardened with unrealistic and realistic evasion attacks. Our observations for DREBIN indicate that realistic attacks often provide robustness against similar attacks. For instance, as shown in Figure~\ref{fig:domain_constraints} (DREBIN-DNN, subfigure~a), while utilizing PK-Greedy achieves relatively high adversarial robustness against PK-Greedy (63.6\% robust accuracy), its robustness against EvadeDroid is very low (e.g., 0.5\% robust accuracy). However, employing unrealistic evasion attacks such as JSMA can strengthen DNN models against both realistic evasion attacks, achieving 59.3\% robust accuracy against PK-Greedy and 48.6\% against EvadeDroid. We observe a similar trend for other classifiers, but the robustness achieved using PK-Greedy and EvadeDroid in AT is significantly lower for linear SVM and DT compared to DNN. Overall, PK-Greedy and EvadeDroid are less effective at hardening malware classifiers trained on DREBIN (a sparse, high-dimensional discrete feature space), particularly for models using linear SVM and DT. However, these results highlight that domain constraints, which are implicitly defined through the use of realizable AEs in AT, have a more significant impact on the success of AT in dense, low-dimensional discrete feature spaces. For instance, applying EvadeDroid to harden the linear SVM trained on RAMDA resulted in 41.8\% robustness against PK-Greedy, while the same model trained on DREBIN only achieved 0.5\%.

\begin{figure}[t]
    \centering    
    \hfill
    \begin{subfigure}[b]{0.48\textwidth}
        \centering
        \includegraphics[width=\textwidth]{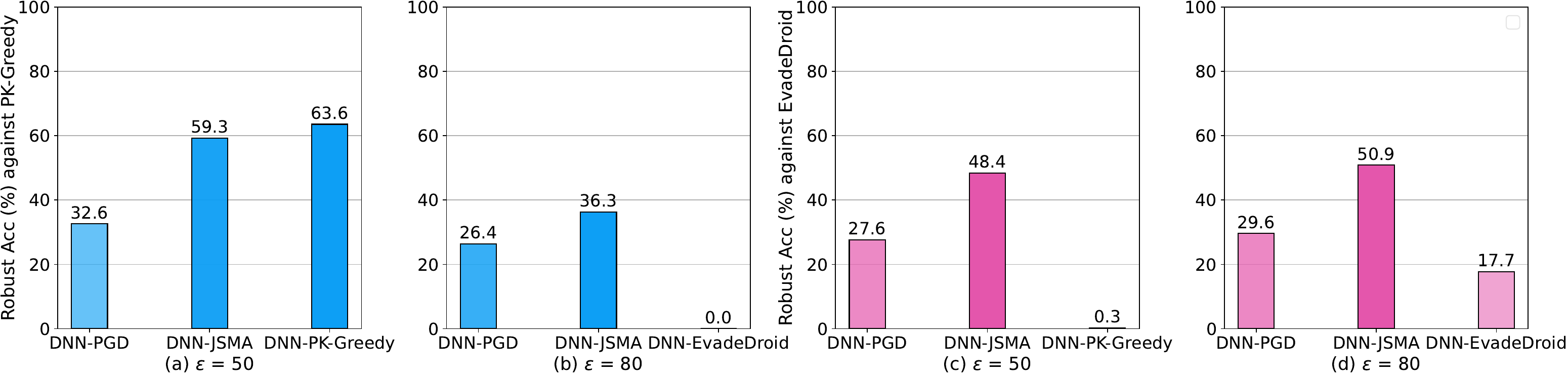} % Replace with your image file
        \vspace{-0.9em}
        \caption*{DNN models trained on DREBIN}
        \label{fig:image1}
    \end{subfigure}
   
    \hfill
    \begin{subfigure}[b]{0.48\textwidth}
        \centering
        \includegraphics[width=\textwidth]{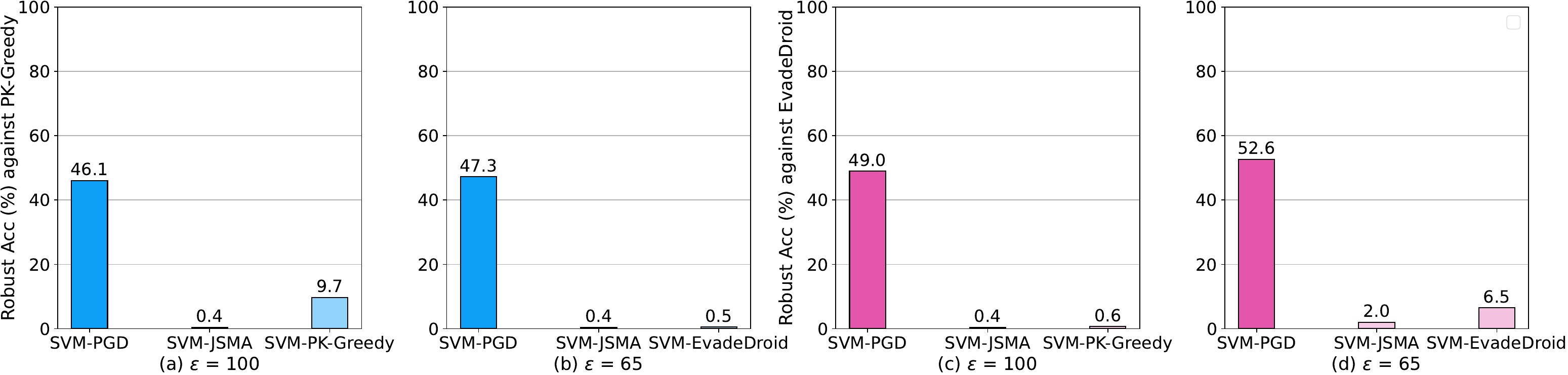} % Replace with your image file
        \vspace{-0.9em}
        \caption*{Linear SVM models trained on DREBIN}
        \label{fig:image2}
    \end{subfigure}
    
    \hfill
    \begin{subfigure}[b]{0.48\textwidth}
        \centering
        \includegraphics[width=\textwidth]{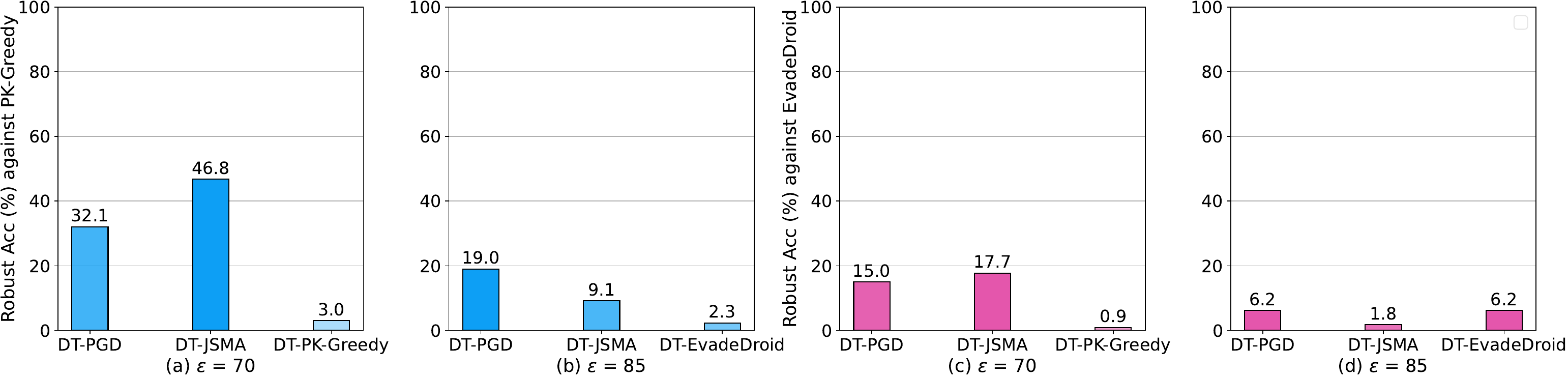} % Replace with your image file
        \vspace{-0.9em}
        \caption*{DT models trained on DREBIN}
        \label{fig:image3}
    \end{subfigure}
    
    \hfill
    \begin{subfigure}[b]{0.48\textwidth}
        \centering
        \includegraphics[width=\textwidth]{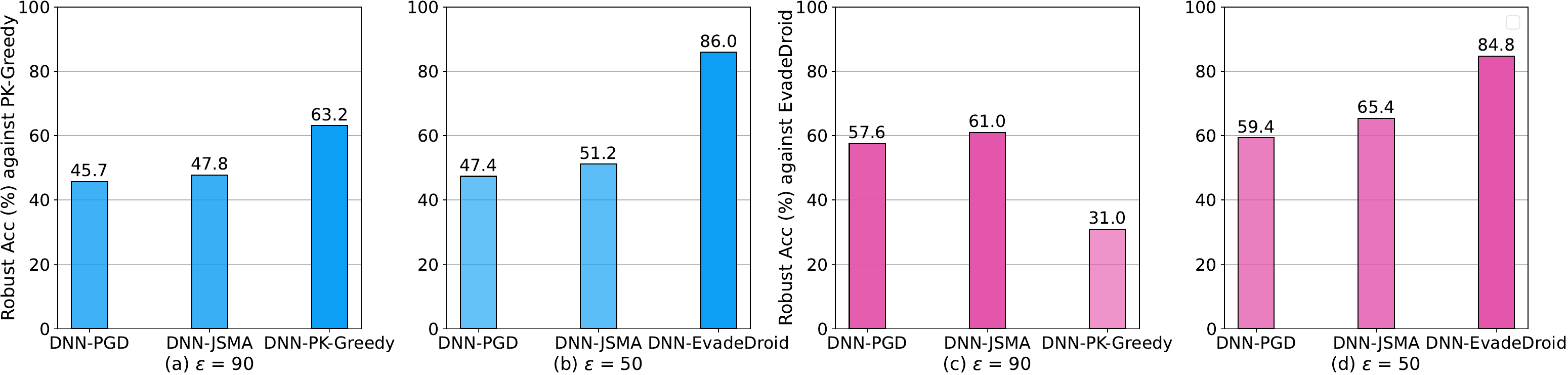} % Replace with your image file
        \vspace{-0.9em}
        \caption*{DNN models trained on RAMDA}
        \label{fig:image3}
    \end{subfigure}
   
    \hfill
    \begin{subfigure}[b]{0.48\textwidth}
        \centering
        \includegraphics[width=\textwidth]{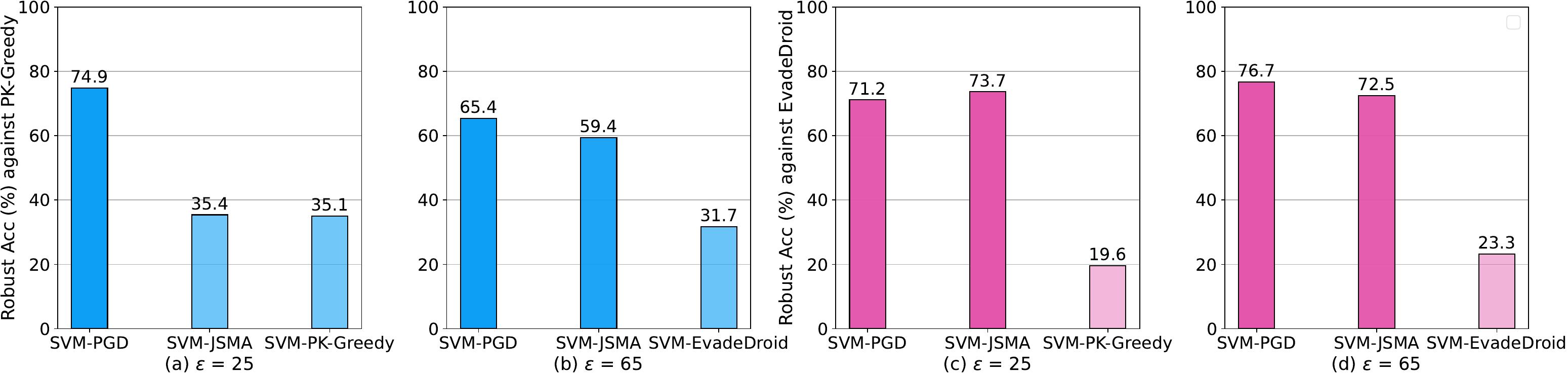} % Replace with your image file
        \vspace{-0.9em}
        \caption*{Linear SVM models trained on RAMDA}
        \label{fig:image3}
    \end{subfigure}    
    \hfill
    \begin{subfigure}[b]{0.48\textwidth}
        \centering
        \includegraphics[width=\textwidth]{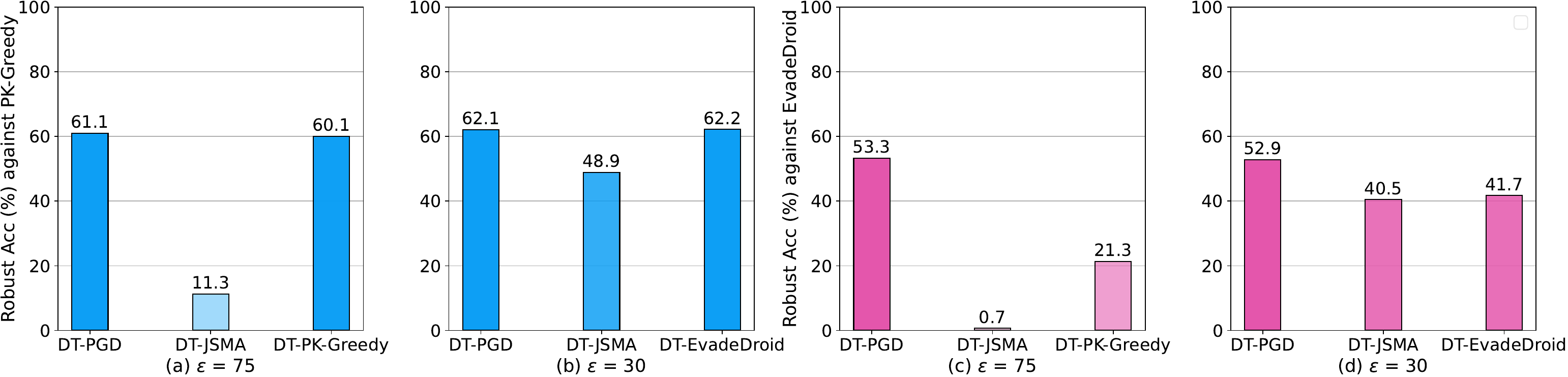} % Replace with your image file
        \vspace{-0.9em}
        \caption*{DT models trained on RAMDA}
        \label{fig:image3}
    \end{subfigure}
    \caption{Robust accuracy, gained from AT, of DNN, SVM, and DT models trained on DREBIN and RAMDA representations of DREBIN20, hardened with either unrealistic or realistic evasion attacks against realistic evasion attacks. The perturbation and attack bounds, both represented by $\epsilon$, are identical in each subfigure.}
    % \vspace{-2em}
    \label{fig:domain_constraints}
\end{figure}

Figure~\ref{fig:domain_constraints} illustrates the robust accuracy of various malware classifiers hardened against unrealistic and realistic evasion attacks. For DREBIN, our observations indicate that realistic attacks often confer robustness primarily against similar attack types. For example, as shown in Figure~\ref{fig:domain_constraints} (DREBIN-DNN, subfigure~a), while using PK-Greedy for AT achieves high robustness against PK-Greedy itself (63.6\%), its robustness against EvadeDroid is markedly lower (e.g., 0.3\%). Conversely, employing unrealistic evasion attacks like JSMA can strengthen DNN models more broadly, yielding 59.3\% robust accuracy against PK-Greedy and 48.4\% against EvadeDroid. A similar trend is observed for other classifiers; however, the robustness attained using PK-Greedy and EvadeDroid in AT is significantly lower for linear SVM and DT models compared to DNN. Overall, PK-Greedy and EvadeDroid are less effective at hardening malware classifiers trained on DREBIN---a sparse, high-dimensional discrete feature space---particularly for linear SVM and DT. These results underscore that domain constraints, which are implicitly defined through the use of realizable AEs during training, have a more substantial impact on the success of AT in dense, low-dimensional discrete feature spaces. For instance, applying EvadeDroid to harden the linear SVM trained on RAMDA resulted in 31.7\% robustness against PK-Greedy, whereas the same model trained on DREBIN achieved only 0.5\%.

% Furthermore, Figures~\ref{fig:domain_constraints}~(RAMDA-DNN, subfigure a and d) and \ref{fig:domain_constraints}~(RAMDA-DT, subfigure a and d) show that utilizing EvadeDroid to harden DNN and DT models trained on RAMDA provides high adversarial robustness against both PK-Greedy and EvadeDroid. This suggests that exploring the low-dimensional feature space blindly is more effective than following gradients, which are potentially biased towards finding certain vulnerable regions rather than all vulnerable regions. In other words, in the RAMDA, which is a low-dimensional feature space, robustness achieved by EvadeDroid, which works in ZK settings, is more transferable than PK-Greedy, which works in PK settings.

Furthermore, Figures~\ref{fig:domain_constraints} (RAMDA-DNN, subfigures a and d) and \ref{fig:domain_constraints} (RAMDA-DT, subfigures a and d) show that using EvadeDroid to harden DNN and DT models trained on RAMDA yields strong adversarial robustness against both PK-Greedy and EvadeDroid attacks. This suggests that, in some cases, blindly exploring the low-dimensional, discrete feature space can be more effective than following gradients, which may be biased toward only certain vulnerable regions rather than the entire space. In other words, within RAMDA's low-dimensional discrete space, the robustness achieved by EvadeDroid, which operates in a ZK setting, is more transferable than that achieved by PK-Greedy, which assumes PK setting.

Our observations in Figures~\ref{fig:domain_constraints} also indicate that PGD and JSMA can potentially be effective in hardening malware classifiers; however, their effectiveness must be weighed against their impacts on clean performance. Table~\ref{table:clean_acc_domain_constraints} shows notable drops in F1 Score, especially when PGD and JSMA are used for hardening linear SVM and DT, while PK-Greedy and EvadeDroid maintain clean performance. The decline in clean performance with unrealistic attacks may stem from AEs distorting class boundaries, causing artificial overlaps. This shift forces the model to focus on unfeasible regions (i.e., the areas where realizable AEs cannot be placed), hindering its ability to generalize and leading to more misclassifications on clean data. The use of realizable AEs in AT avoids these negative effects by guiding the optimizer to explore feasible and genuinely vulnerable regions.

\begin{figure}
    \centering
    \vspace{0.3em}
    \begin{subfigure}[b]{0.105\textwidth}
        \centering
        \includegraphics[width=\textwidth]{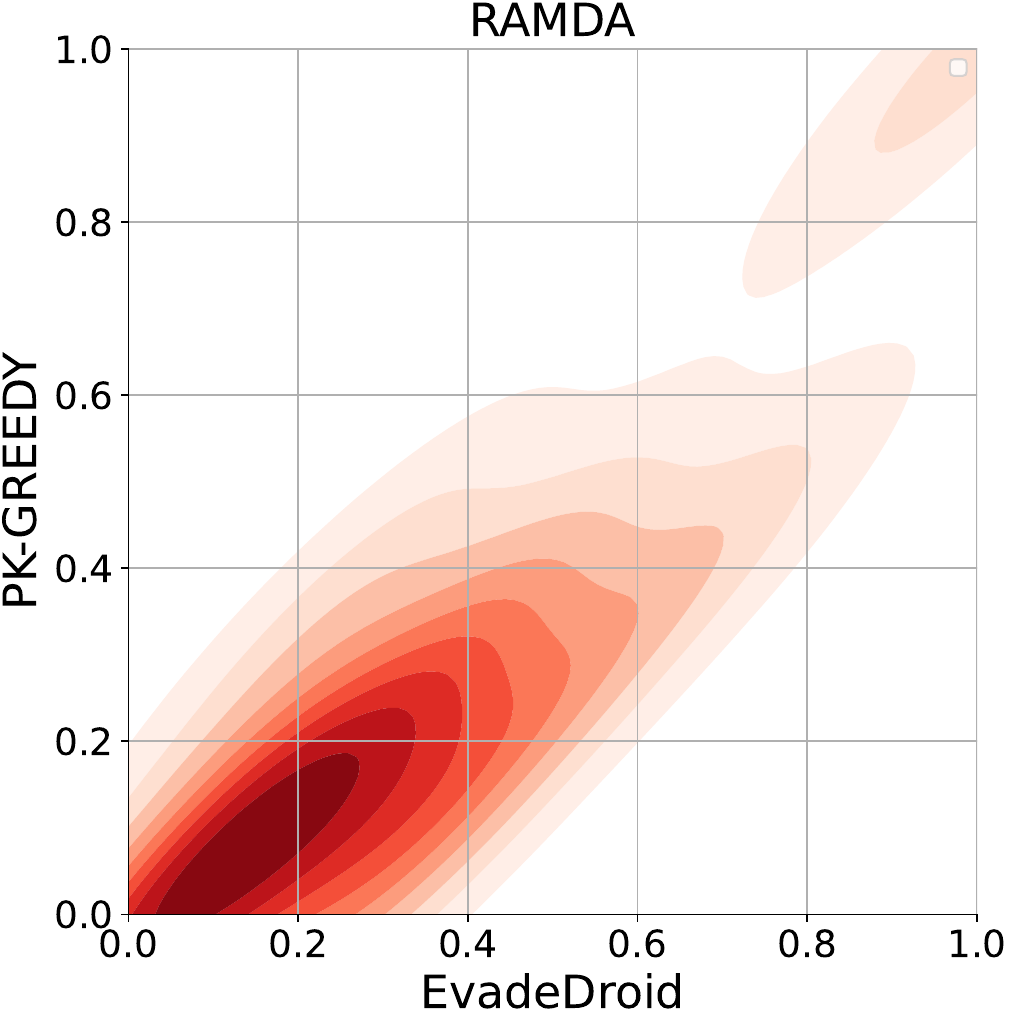}
        \vspace{-0.9em}
        \caption{}
        \label{fig:image1}
    \end{subfigure}
    \hspace{0.2em}
    \begin{subfigure}[b]{0.105\textwidth}
        \centering
        \includegraphics[width=\textwidth]{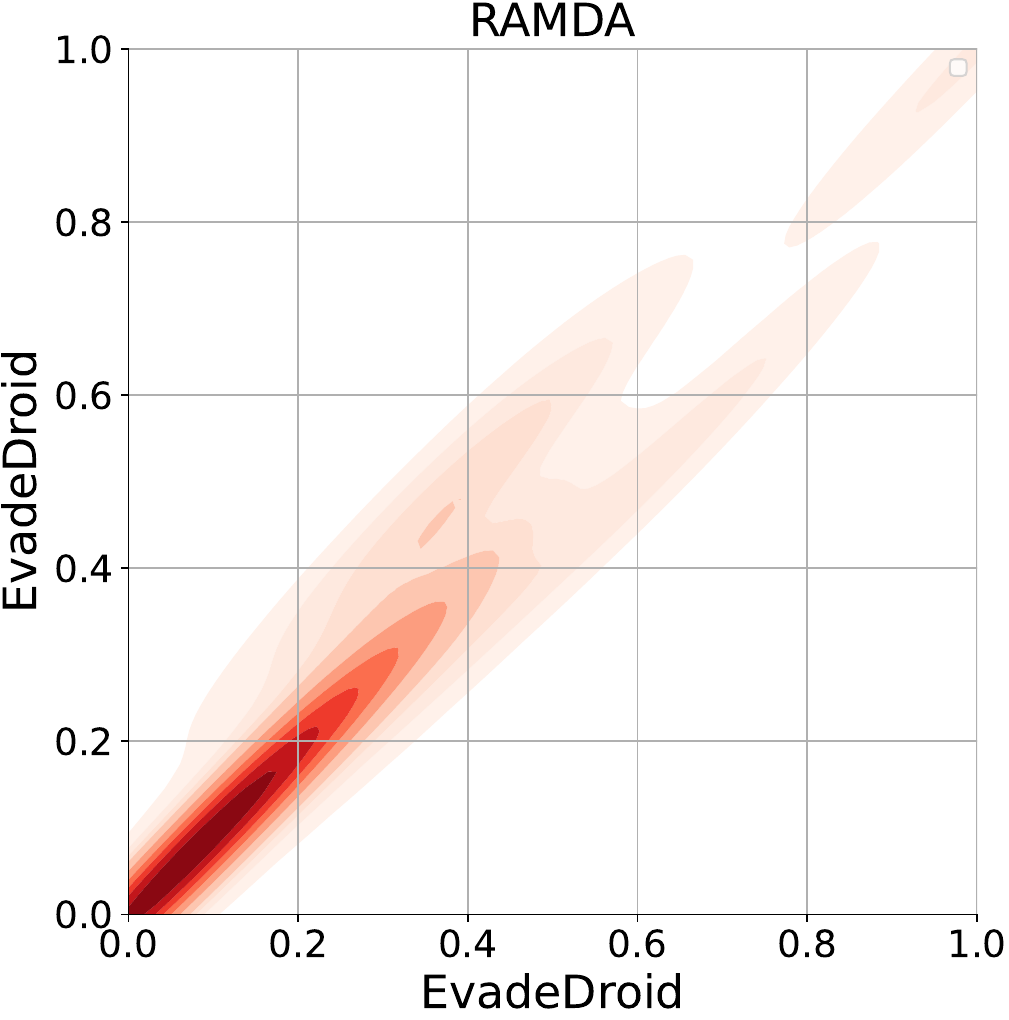}
        \vspace{-0.9em}
        \caption{}
        \label{fig:image2}
    \end{subfigure}
    \hspace{0.2em}
    \begin{subfigure}[b]{0.105\textwidth}
        \centering
        \includegraphics[width=\textwidth]{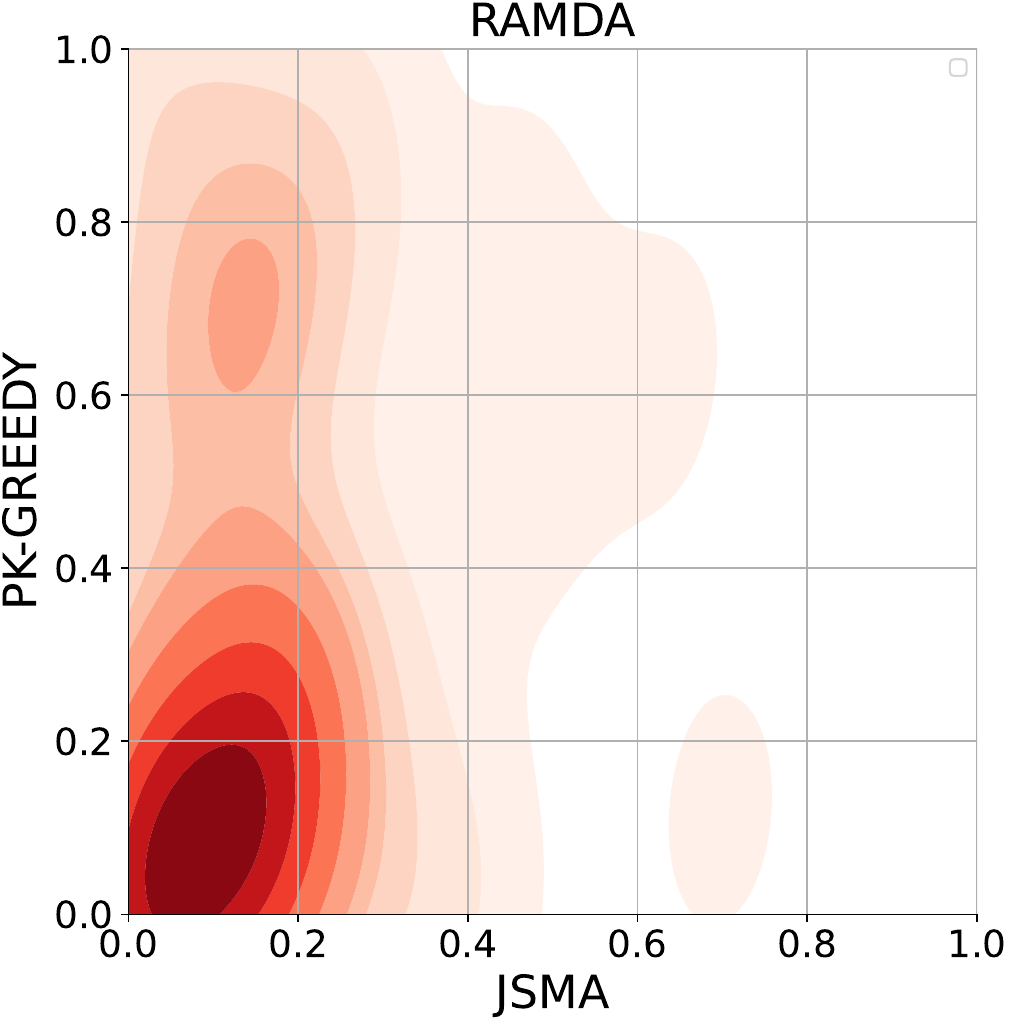}
        \vspace{-0.9em}
        \caption{}
        \label{fig:image3}
    \end{subfigure}
    \hspace{0.2em}
    \begin{subfigure}[b]{0.105\textwidth}
        \centering
        \includegraphics[width=\textwidth]{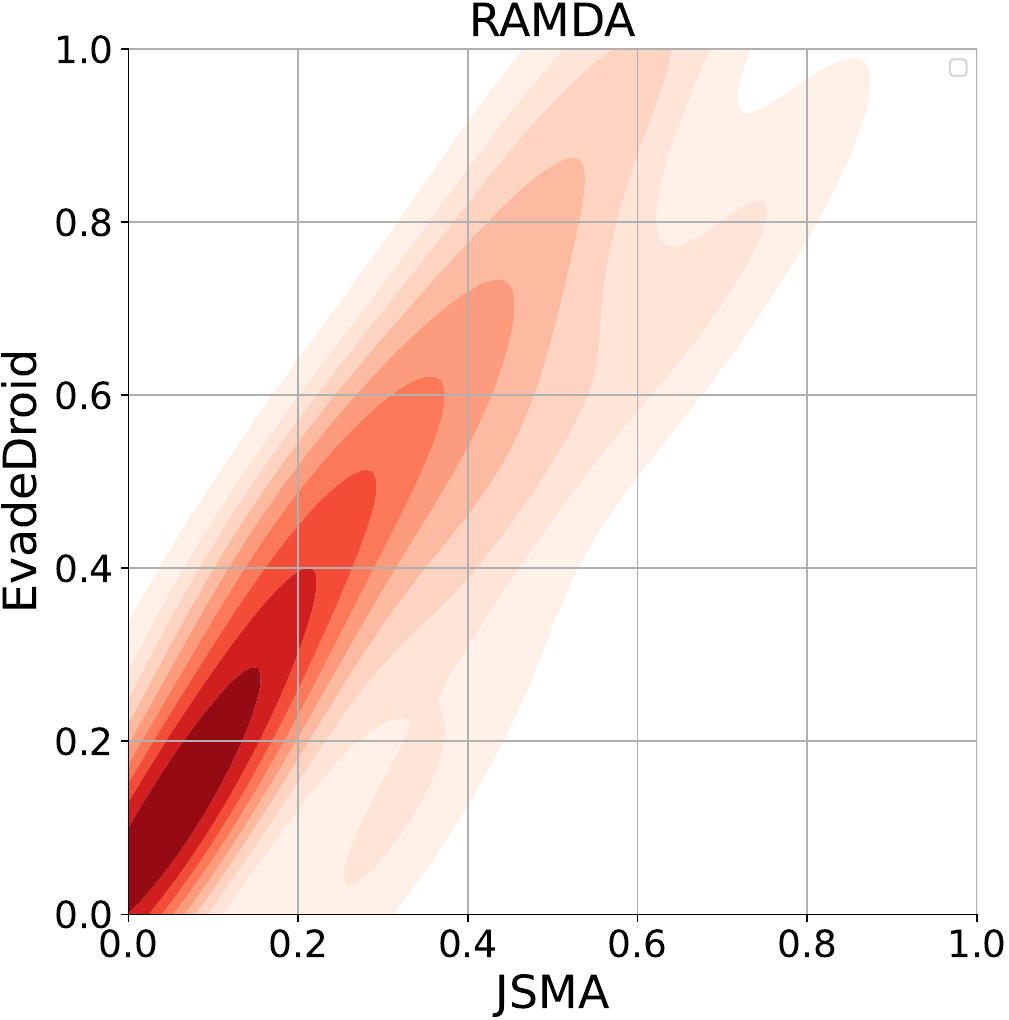}
        \vspace{-0.9em}
        \caption{}
        \label{fig:image4}
    \end{subfigure}
    \vspace{0.3em}
\caption{Joint distribution plots which compare the importance of features used both during the hardening phase (x-axis) and the attack phase (y-axis), referring to the scenario described in~\autoref{fig:domain_constraints}.}
\vspace{-1em}
\label{fig:joint_distribution_plots}
\end{figure}

% Figure X shows the higher adversarial robustness of models trained on RAMDA in comparison to DREBIN which confirms the AT works better for the models built on low-dimensional feature space as AT has a higher chance to uncover blind spots. To further investigate the size of feature space, we consider DREBIN with varied dimensions ...

% To investigate the impact of domain constraints on AT, we use PK-Greedy and EvadeDroid in the robust optimization instead of PGD and JSMA to generate realizable AEs. Indeed, it mean 

% \hb{This subsection investigates RQ1 and RQ2. Note that we must first clarify all the settings related to the experimental designs provided in \S\ref{subsec:characterizing_the_effectiveness_of_AT} (e.g., considering perturbation budgets from 0 to 100 with a step size of 5 for evaluating the impact of perturbation budgets), and then report and explain the results.}
% \jc{We should include data and plots for the rollback effect}
\subsubsection{Properties of AEs}
\label{subsubsec:properties_of_aes}
This section investigates \hyperlink{RQ2}{RQ2}. This research question is similar to the one explored in~\cite{dyrmishi2023empirical}. Intuitively, different methodologies for crafting AEs inherently produce distinct characteristics in the generated samples, as they exploit varying sets of features and algorithms. This distinction is illustrated in \autoref{fig:conf_DNN_without} of Appendix~\ref{app:confidence_of_AEs}, where it is evident that the confidence levels of AEs created by two different gradient-based strategies vary significantly. To delve deeper into AEs and their influence on the effectiveness of AT, we employ 
% the two metrics discussed 
diverse tools mentioned in
in \S\ref{subsec:analysis}.\\ \indent
% The level of complexity increases if we consider constrained strategies to generate adversarial points, as problem-space-compliant techniques. 
We first utilize the joint feature importance metric introduced in~\ref{subsec:joint_feature_importance} to determine the significance of features for both the defense and attack sides. Specifically, we analyze some interesting results reported in \S\ref{subsubsec:domain_constraints}. As shown in Figures~\ref{fig:domain_constraints}-b and~\ref{fig:domain_constraints}-d for DNN models trained on RAMDA, the adversarial robustness achieved by using EvadeDroid in hardening DNN trained on RAMDA is highly transferable, providing high adversarial robustness against not only EvadeDroid but also PK-Greedy. Figures~\ref{fig:joint_distribution_plots}-a and~\ref{fig:joint_distribution_plots}-b clarify this, demonstrating that some overlapping regions are highly important for both the defender and the attacker.
% EvadeDroid used in hardening and PK-Greedy used in attacking in (Figure~\ref{fig:joint_distribution_plots}a), and for both EvadeDroid used in hardening and EvadeDroid used in attacking in (Figure~\ref{fig:joint_distribution_plots}b).
Note that in each plot, the contours highlight regions where the importance of common features is greater than in areas outside the contours. The contours near the top right indicate that the common features are mostly important for both the defender and the attacker, with darker contours representing a higher feature overlap. Indeed Figures~\ref{fig:joint_distribution_plots}-a and~\ref{fig:joint_distribution_plots}-b demonstrate that EvadeDroid can generate some AEs during training that highlight features often targeted by both EvadeDroid and PK-Greedy during attacks. This helps the DNN model guard against adversarial changes that might affect these features. Figures~\ref{fig:domain_constraints}-b and~\ref{fig:domain_constraints}-d for DNN models trained on RAMDA also illustrate a relatively high adversarial robustness achieved by hardening DNN with JSMA, an unrealistic attack. We visualize similar joint importance feature plots (Figures~\ref{fig:joint_distribution_plots}-c and \ref{fig:joint_distribution_plots}-d) which, like the previous analysis, demonstrate that some overlapping regions are highly important for both defense and attack sides.\\ \indent
% JSMA used in hardening and PK-Greedy used in attacking (Figure~\ref{fig:joint_distribution_plots}c), and for both JSMA used in hardening and EvadeDroid used in attacking (Figure~\ref{fig:joint_distribution_plots}d).
% Next, 
Then, we examine the potential connection between the roughness of the decision boundary, measured using the decision-function roughness metric introduced in \S\ref{subsec:decision_function_roughness}, and adversarial robustness. Here, we consider several models hardened by PGD on RAMDA with $\epsilon$ values ranging from 50 to 60. As shown in Figure~\ref{fig:real_robust_acc_perturbation_budgets_vs_robustness_bounded_attack}-c, this range is particularly interesting because, at $\epsilon=55$, DT exhibits a significant drop in adversarial robustness against PK-Greedy, while DNN and SVM models remain more stable. Table~\ref{table:decision_roughness} presents the decision-function roughness $\gamma$ and robust accuracy for these models. Our observations indicate that increasing $\epsilon$ can affect adversarial robustness, particularly in DT. For instance, DT's robust accuracy drops significantly from 58.1\% to 18.8\% when $\epsilon$ changes from 50 to 55, whereas the $\gamma$ changes only slightly by -0.03. Furthermore, as shown for the DT model, increasing or decreasing $\epsilon$ does not necessarily correlate with changes in $\gamma$. Lastly, while the results for DNN and SVM suggest that lower $\gamma$ can lead to better adversarial robustness, this is not consistently the case, as evidenced by the DT results contradicting this hypothesis.

\begin{table}[t]
\footnotesize
\caption{Analyzing various models trained on RAMDA in terms of decision-function roughness $\gamma$ and robust accuracy R\_Acc (\%) against PK-Greedy. These models are hardened by PGD.}
\label{table:decision_roughness}
% \begin{tabular}{p{2.2cm}|p{0.5cm}p{0.5cm}|p{1cm}p{1cm}}
\begin{center}
\begin{tabular}{l|rr|rr|rr}
  \toprule
  \multirow{2}{*}{\textbf{Model}} 
      & \multicolumn{2}{c|}{\textbf{$\epsilon=50$}} 
          & \multicolumn{2}{c|}{\textbf{$\epsilon=55$}} & \multicolumn{2}{c}{\textbf{$\epsilon=60$}}\\  \cline{2-7}
  & \textbf{$\gamma$} & \textbf{R\_Acc} & \textbf{$\gamma$} & \textbf{R\_Acc} & \textbf{$\gamma$} & \textbf{R\_Acc} \\  \midrule
    DNN & 0.82 & 55.7 & 0.82 & 58.6& 0.84 & 57.8\\
    SVM & 0.26 & 74.6 & 0.31 & 76.0& 0.26 & 77.6\\
    DT & 0.16 & 58.1 & 0.13 & 18.8& 0.17 & 49.2\\    
  \bottomrule
\end{tabular}
\end{center}
\vspace{-1.5em}
\end{table}

\begin{figure}[b!]
    \centering
    \vspace{0.3em}
    \begin{subfigure}[b]{0.172\textwidth}
        \centering        \includegraphics[width=\textwidth]{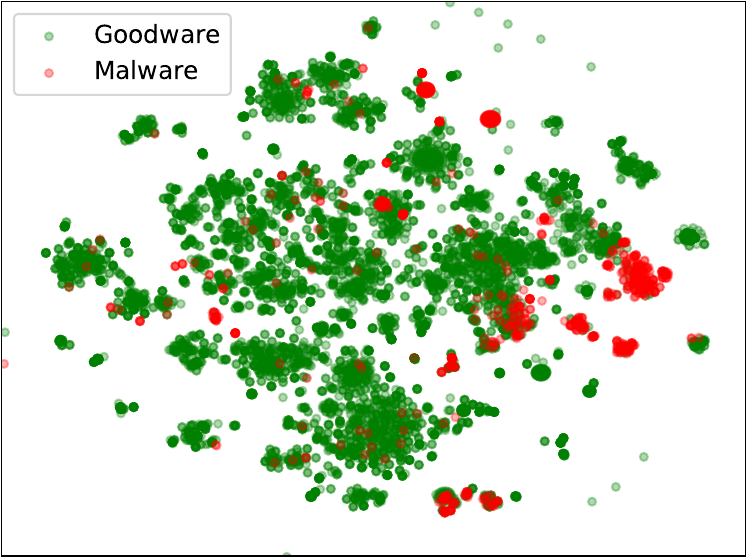} % Replace with your image file
        \vspace{-0.9em} % Adjust vertical space as needed
        \caption{DREBIN (DREBIN20)}
        \label{fig:image1}
    \end{subfigure}
    % \hfill
    \hspace{0.5em} 
    \begin{subfigure}[b]{0.172\textwidth}
        \centering
        \includegraphics[width=\textwidth]{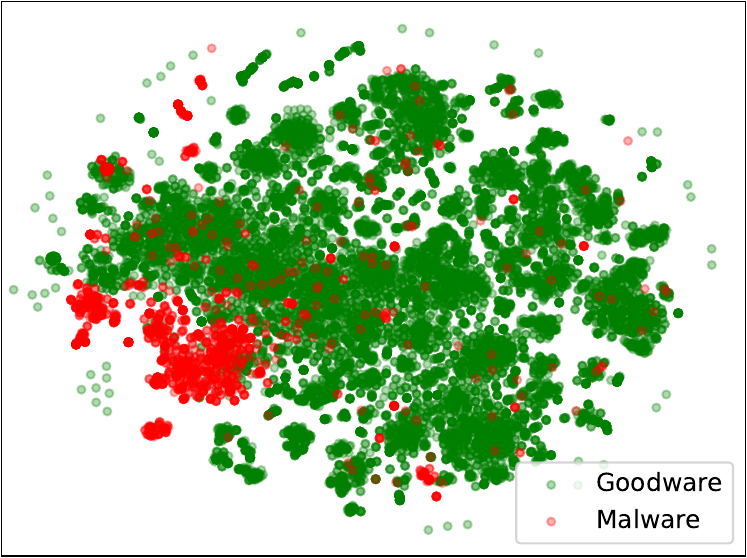} % Replace with your image file
        \vspace{-0.9em} % Adjust vertical space as needed
        \caption{DREBIN (APIGraph)}
        \label{fig:image2}
    \end{subfigure}
     
    \vspace{0.3em}
    
    \begin{subfigure}[b]{0.172\textwidth}
        \centering
        \includegraphics[width=\textwidth]{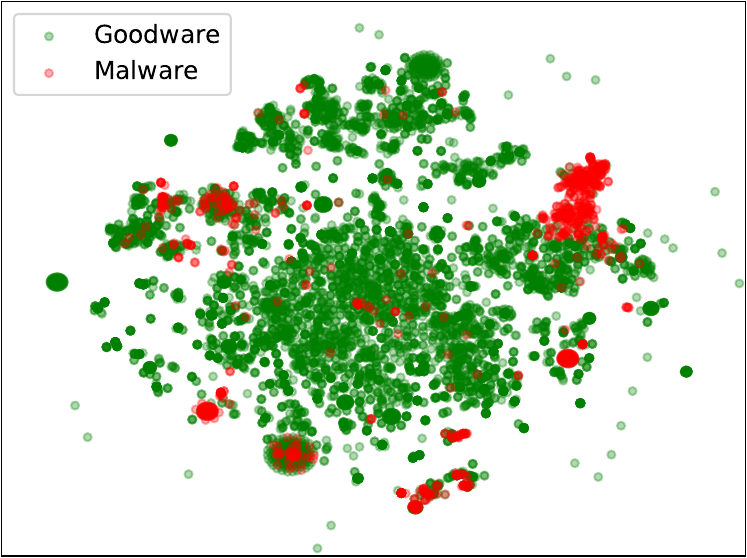} % Replace with your image file
        \vspace{-0.9em} % Adjust vertical space as needed
        \caption{RAMDA (DREBIN20)}
        \label{fig:image1}
    \end{subfigure}
    % \hfill
    \hspace{0.5em} 
    \begin{subfigure}[b]{0.172\textwidth}
        \centering
        \includegraphics[width=\textwidth]{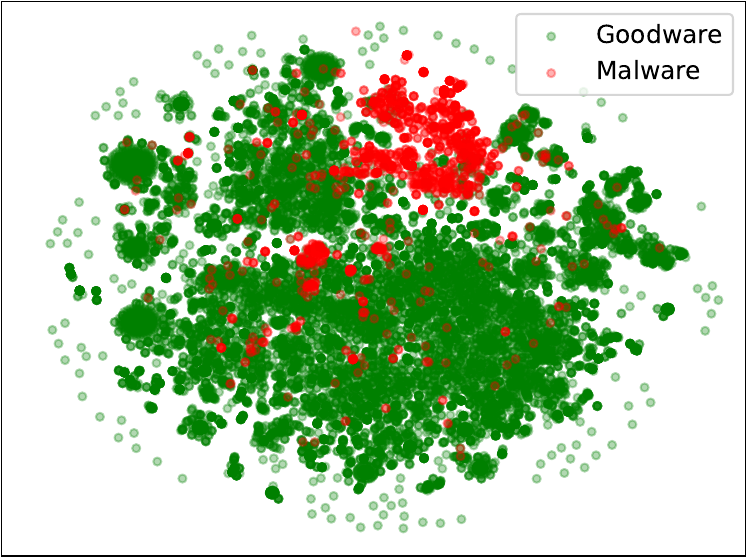} % Replace with your image file
        \vspace{-0.9em} % Adjust vertical space as needed
        \caption{RAMDA (APIGraph)}
        \label{fig:image2}
    \end{subfigure}
    \caption{t-SNE plots of goodware and malware from DREBIN20 and APIGraph, represented with DREBIN and RAMDA features.}
    \vspace{-1.5em}
    \label{fig:t_sne}
\end{figure}

Next, as shown in the results presented in \S\ref{subsubsec:perturbation_bound_and_the_confidence_of_aes} and \S\ref{subsubsec:varying_fraction_of_aes}, hardened models trained on the APIGraph dataset often achieve higher clean and robust accuracy compared to those trained on DREBIN20. 
To understand this performance gap, we use t-SNE visualizations of both datasets. As shown in Figure~\ref{fig:t_sne}, APIGraph plots---under both DREBIN and RAMDA features---demonstrate clearer separation between malware and goodware than DREBIN20, likely contributing to the improved performance. Additionally, prior work~\cite{zheng2025learning} indicates that APIGraph experiences less distribution drift, supporting its suitability for training effective and stable malware classifiers.

Finally, we assess the \textit{hardening cost} of AT, which depends largely on the computational cost of generating AEs, to evaluate its practical feasibility. AE generation time varies widely depending on the attack type. 
Unrealistic attacks like PGD and JSMA are fast (0.002--0.09 seconds per AE, depending on the perturbation bound), while realistic ones like PK-Greedy and EvadeDroid take over 200 seconds~\cite{bostani2024evadedroid,bostani2022domain}, making AT impractical. To mitigate this, we follow~\cite{Pierazzi2020Intriguing} and generate realizable AEs in the feature space by first identifying feature-level perturbations corresponding to all problem-space transformations, cutting AE generation time to about one second. 
% AT is relatively fast with unrealistic evasion attacks like PGD or JSMA, as they generate each AE in 0.001–0.01 seconds, depending on the perturbation bound. However, when using realistic evasion attacks such as PK-Greedy and EvadeDroid, AE generation exceeds 200 seconds on average, rendering AT impractical. To address this, we adopt the approach proposed in~\cite{Pierazzi2020Intriguing}, which generates realizable AEs in the feature space rather than the problem space by first identifying the feature-level perturbations that correspond to each problem-space transformation employed by PK-Greedy and EvadeDroid. This approach reduces AE generation time to about one second.
While this significantly improves efficiency, AT with realistic evasion attacks remains costly, as generating one AE per second is still too slow for AT methods that require generating many AEs in each epoch of robust optimization. For example, training a DNN with PK-Greedy can take around 15 hours per epoch.

\subsubsection{Computational Constraints} This study systematically analyzes how various factors affect the effectiveness of AT in malware classification. We trained 1K+ models and ran 3K+ evaluations, which took more than six months despite significant hardware resources (see Appendix~\ref{app_sub:computational_resources}). 
% This study systematically examines the impact of various factors on the effectiveness of AT in hardening malware classifiers. To this end, we trained over 1K models in total to conduct more than 3K evaluations. Despite allocating significant hardware resources (refer to the Computational Resources section in Appendix~\ref{app:settings}), the experiments took over six months to complete. 
Specifically, training times ranged from 4 to 140 hours per model, depending on the settings, while clean performance and robustness evaluations took 15 and 780 seconds per model, respectively. It is important to note that a full exploration of all possible combinations--datasets, feature representations, classifiers, and robust optimization settings (e.g., perturbation bounds and adversarial fractions)--would require nearly 27K models (see Appendix~\ref{app:exhaustive_exploration}). Given that training just 1K models took over six months, pursuing an exhaustive training regimen is computationally infeasible. To balance practicality and meaningful insights, we adopt a structured approach, varying key parameters while holding others constant. For instance, in \S\ref{subsubsec:perturbation_bound_and_the_confidence_of_aes}, we examine perturbation bounds and adversarial confidence across datasets, features, and classifiers, fixing the adversarial fraction and using unrealistic attacks, reducing the model count to 480 for efficient evaluation.
% in \S\ref{subsubsec:perturbation_bound_and_the_confidence_of_aes}, we analyze perturbation bounds and adversarial confidence across datasets, features, and classifiers, while fixing the adversarial fraction and using unrealistic attacks, reducing the model count to 480 for meaningful yet efficient evaluation.
% we analyze perturbation bounds and adversarial confidence across different datasets, feature representations, and classifiers while fixing adversarial fraction and relaxing domain constraints via unrealistic attacks. This reduced the model count to 480, ensuring meaningful observations without excessive computational cost.

% \vspace{-.5em}
\subsubsection{Key Findings}
\label{subsec:key_findings}
Our results shed light on the trade-offs between clean performance and adversarial robustness when using AT to strengthen malware classifiers, advocating for structured evaluations over exhaustive training. Our study challenges conclusions drawn in previous influential research: Specifically, \S\ref{subsubsec:domain_constraints} questions the conclusions of~\cite{cortellazzi2024intriguingpropertiesadversarialml, dyrmishi2023empirical}, arguing that realistic evasion attacks do not always enhance robustness, as their effectiveness is conditional on multiple interacting factors. Additionally, while~\cite{cortellazzi2024intriguingpropertiesadversarialml, lucas2023adversarial} report that unrealistic evasion attacks degrade clean performance, we show that this effect varies by classifier, particularly in deep, non-linear models, where AT with unrealistic attacks does not necessarily harm clean performance. However, when realistic attacks are used, our results align with~\cite{cortellazzi2024intriguingpropertiesadversarialml, lucas2023adversarial}, supporting the claim that such settings can maintain clean accuracy. Our findings in \S\ref{subsubsec:perturbation_bound_and_the_confidence_of_aes} and \S\ref{subsubsec:varying_fraction_of_aes} also challenge the assumption made in~\cite{li2023pad}, demonstrating that generating highly confident AEs in the inner loop of adversarially robust optimization is not always necessary. In fact, depending on the specific settings, using low-confidence AEs---settling for a local rather than a global solution---can sometimes lead to better robustness. Finally, our results in \S\ref{subsubsec:varying_fraction_of_aes} challenge the observations reported in~\cite{dyrmishi2023empirical,lucas2023adversarial} regarding the adversarial fraction, showing that AT can benefit from increasing the budget of AEs generated by unrealistic attacks.

In addition to challenging relevant influential studies, our systematic evaluation reveals novel synergies between model architecture and feature-space structure, highlighting that practitioners must make intelligent, aligned choices across the entire malware classifier pipeline to build effective detectors. Drawing on our extensive empirical investigation, whose complete results are summarized in Table~\ref{table:summary_of_key_findings} of Appendix~\ref{app:overview_of_results}, we identify four key findings to inform and guide future research on AT in the malware domain.

\noindent\textbf{Takeaway 1 -- Architecture dictates how AT fails or adapts.}
\label{takeaway1}
Our analysis (\S\ref{subsubsec:perturbation_bound_and_the_confidence_of_aes}, \S\ref{subsubsec:varying_fraction_of_aes}) shows that in discrete feature spaces, AT affects models differently: linear and shallow classifiers often lose clean accuracy at larger perturbation bounds, whereas deep non-linear models maintain accuracy while gaining robustness.

\noindent\textbf{Recommendations:}
\vspace{-0.4em}
\begin{itemize}
\item Avoid AT on linear models unless either small perturbation bounds or low-confidence AEs are used to reduce clean performance loss.
\item Prefer deep non-linear models, which balance robustness and accuracy.
\end{itemize}

\noindent\textbf{Takeaway 2 -- AE ratio tunes the robustness-accuracy tension.}
Our results (\S\ref{subsubsec:varying_fraction_of_aes}) confirm that the proportion of AEs used in AT critically governs the clean-robustness trade-off. Too few AEs yield weak robustness, while too many---especially in linear models---severely degrade clean accuracy.

\noindent\textbf{Recommendations:}
\vspace{-0.4em}
\begin{itemize}
\item Balance the AE fraction with model capacity and perturbation strength: start small fraction and gradually increase it while monitoring the clean-robust trade-off. Deep non-linear models tolerate higher AE ratios under larger perturbations, whereas linear and shallow models need smaller fractions to preserve clean performance.
\item Avoid full-batch AT in linear models, which significantly degrades clean performance.
% \item Validate AE fraction empirically by monitoring both clean and adversarial performance.
\end{itemize}

\noindent\textbf{Takeaway 3 -- Dense, low-dimensional spaces enable effective AT with aligned models and attacks.}
Empirical results (\S\ref{subsubsec:perturbation_bound_and_the_confidence_of_aes}, \S\ref{subsubsec:domain_constraints}) indicate that dense, low-dimensional discrete features provide better robustness, even against different attacks, than sparse, high-dimensional ones. However, achieving robustness without degrading clean performance requires aligning the model architecture and attack realism.

\noindent\textbf{Recommendations:}
\vspace{-0.4em}
\begin{itemize}
\item Favor dense, low-dimensional representations to expose large vulnerabilities for AT in discrete spaces.
\item Use realizable AEs for deep non-linear models in dense discrete spaces to enhance robustness while preserving clean performance.
\item Avoid unrealizable AEs for linear models in any discrete space to prevent clean performance loss.
\item Use unrealizable AEs only with shallow or deep non-linear models in sparse spaces, where robustness can be gained without major loss in clean performance.
\end{itemize}

% \noindent\textbf{Takeaway 5} (based on observations in \S\ref{subsubsec:properties_of_aes}):
% In discrete feature spaces, adversarial robustness depends more on semantic overlap between attacker and defender priorities than on decision boundary smoothness. Robustness improves when classifiers rely on features also targeted by attacks, especially in datasets like APIGraph, where class separability enhances this alignment.

% \noindent\textbf{Practical Recommendations:} \begin{itemize} 
% \item \textit{Do not rely on decision boundary smoothness alone.} Smoothing the classifier's decision surface in discrete spaces does not reliably improve robustness, contrary to trends observed in continuous domains~\cite{eghbal2024rethinking}. 
% \item \textit{Prioritize alignment between defensive and offensive signals.} Robust models tend to emerge when training emphasizes features commonly exploited by attackers, creating shared ground that is harder to bypass. 
% \item \textit{Assess feature importance from both perspectives.} During evaluation, analyze whether critical features for classification are also targeted by AEs; high overlap indicates stronger inherent robustness. 
% \item \textit{Consider the dataset's structural properties.} Prefer datasets where malicious and benign samples exhibit clear separability, as such structure facilitates learning of meaningful, robust features. 
% \end{itemize}

\noindent\textbf{Takeaway 4 -- Robustness in discrete spaces relies on feature alignment, not decision boundary smoothness.}
Our analysis (\S\ref{subsubsec:properties_of_aes}) shows that in discrete spaces, robustness improves when classifiers focus on features also targeted by attacks, particularly in datasets like APIGraph with high class separability. Contrary to~\cite{eghbal2024rethinking}, decision boundary roughness does not necessarily imply vulnerability, so smooth boundaries alone do not show robustness.

\noindent\textbf{Recommendations:}
\vspace{-0.4em}
\begin{itemize}
\item Do not rely solely on decision boundary smoothness to ensure robustness; it is insufficient in discrete spaces.
% \item Emphasize features likely exploited by attacks during training to create aligned, robust classifiers.
\item Assess whether AT covers critical features that overlap with features targeted by potential evasion attacks; higher overlap indicates stronger robustness.
\item Prefer datasets with clear class separability, as this facilitates learning robust, meaningful features.
\end{itemize}

\section{Related Work}
\vspace{-0.7em}

\noindent\textbf{Adversarial Training.} Although the study by Goodfellow et al.~\cite{Goodfellow2014Explaining} is recognized as the first to demonstrate that the inclusion of AEs during the training phase can enhance the robustness of ML models to evasion attacks, the robust optimization formulation proposed by Madry et al.~\cite{madry2017towards} for AT marks a turning point in this area. Over the past few years, numerous studies have focused on robust optimization to enhance the adversarial robustness of ML models. For instance, Zhang et al.~\cite{zhang2019limitations} demonstrated the limitations of robust optimization in ensuring robustness when test points slightly deviate from the training set distribution. Zhang et al.~\cite{zhang2019limitations} revealed that the effectiveness of AT is closely tied to the proximity of test data to the training data manifold.
% Zimmerman et al.\cite{zimmermann2022increasing} proposed a unit test to ensure the reliability of adversarial robustness evaluations. 
%Tram\'{e}r et al.~\cite{tramer2019adversarial} argued that AT is typically tailored to a single perturbation type rather than multiple perturbation types. 
Levi and Kontorovich~\cite{levi2023splitting} introduced an AT approach where perturbed examples from each class are regarded as distinct categories by dividing each class into \textit{clean} and \textit{adversarial}.
% They also introduced the "blind-spot attack," illustrating that adversarial examples tend to cluster in low-density areas within the training data distribution.
%Losch et al.~\cite{losch2024adversarial} demonstrated that generating AEs for only half of the training data can nearly match the baseline performance of vanilla AT.
% , even on large datasets. 
Zhang et al.~\cite{zhang2019theoretically} proposed a new defense method to balance adversarial robustness and accuracy by decomposing the robust error into classification and boundary errors. 
%Ilyas et al.~\cite{ilyas2019adversarial} contended that AEs stem from non-robust features in the data, mentioning that simply strengthening models against these examples may not resolve the issue of relying on non-robust features for prediction.

\noindent\textbf{Adversarial Training for Malware Detection.}
% AT, particularly robust optimization, has been less extensively studied in malware detection compared to the image domain. 
% In the realm of malware detection, 
AT has been regarded as the most prevalent defense mechanism for strengthening ML-based detectors. In the last few years, several studies~\cite{dyrmishi2023empirical, wang2021advandmal, chen2021using, wang2020mdea, rathore2021robust, huang2019malware, anderson2018learning, yang2017malware, grosse2017adversarial, zhang2021enhanced} have explored adversarial retraining~\cite{Goodfellow2014Explaining} to improve the robustness of malware detectors. Most of these studies~\cite{wang2021advandmal, chen2021using, wang2020mdea, rathore2021robust, huang2019malware, anderson2018learning, yang2017malware, grosse2017adversarial, zhang2021enhanced} have primarily focused on proposing new evasion attacks for generating AEs needed in adversarial retraining. For instance, Grosse et al.~\cite{grosse2017adversarial} adapted the JSMA~\cite{Papernot2016Limitations} to generate highly effective AEs.\\ \indent
One of the major concerns frequently observed in studies exploring AT is that the hardened models were not tested against realistic evasion attacks. Specifically, 
% while these studies discussed how semantics can be preserved in generating AEs, 
they did not clarify whether their attacks meet all domain constraints~\cite{Pierazzi2020Intriguing}.
For example, the AEs used in~\cite{li2023pad, xu2023ofei, rathore2021robust, li2020adversarial, li2021framework, grosse2017adversarial} may lack robustness to preprocessing, as preprocessing operators can potentially remove features added to the Manifest file of Android apps~\cite{Pierazzi2020Intriguing}. Additionally, bytes that are appended into non-executable areas of Portable Executable (PE) files through the attacks described in~\cite{lucas2023adversarial, wang2020mdea, anderson2018learning} might be discarded by preprocessing before classification~\cite{zhao2022gradient}. It is worth noting that the adversarial attacks used in some studies, such as~\cite{dyrmishi2023empirical,huang2018visual}, may not adequately reflect the robustness of detectors against adversaries, as they did not evaluate the adversarial robustness against some attacks conducted in PK settings, potentially allowing adversaries to create strong attacks.\\ \indent
% these attacks do not work under PK settings, potentially allowing adversaries to create strong attacks.
% Moreover, as previously discussed, while many studies have considered AT an effective defense to harden ML-based malware detectors, it has not been extensively explored because either AT is not the primary focus of most studies~\cite{lucas2021malware,xu2023ofei,rathore2021robust,khormali2019copycat,huang2019malware,yang2017malware,grosse2017adversarial} or only exploring key factors have been neglected~\cite{wang2021advandmal
% ,chen2021using,zhang2021enhanced,wang2020mdea,li2020adversaria,li2021framework,huang2018visual,al2018adversaria,anderson2018learning}. There exist only a few studies that have focused on AT  investigating AT. For instance, 
Another notable concern is the lack of in-depth examination of adversarial robustness in most studies since exploring AT was not their primary focus. In recent years, only a few studies within the field of malware have dedicated their research to investigating AT.
% Another notable concern in relevant studies is that in the malware domain research there was no specific focus on investigating AT impact, despite the widespread use of it for enhancing the adversarial robustness of malware detection.
Two significant explorations have been conducted in~\cite{dyrmishi2023empirical} and~\cite{lucas2023adversarial}. In fact, Dyrmishi et al.~\cite{dyrmishi2023empirical,cortellazzi2024intriguingpropertiesadversarialml} examined the influence of domain constraints on AT by exploring realizable AEs generated in the problem space. Their findings demonstrated that in the malware domain, models hardened with unrealizable AEs exhibit less robustness against realistic evasion attacks compared to models strengthened with realizable AEs. Their observations also showed that clean performance is slightly affected after AT.
% , even when unrealizable AEs are used in AT.
Lucas et al.~\cite{lucas2023adversarial} enhanced the efficiency of various problem-space evasion attacks in generating AEs, making AT practical for raw-binary malware detectors. 
%Cortellazzi et al. \cite{cortellazzi2024intriguingpropertiesadversarialml} shows that in their considered setting, realistic AT has a greater impact on the robustness of the hardened model compared to unrealistic adversarial samples, for both the considered classifier.
% Additionally, they demonstrated that in most cases, using a lower-effort version of a specific attack in AT can improve the detectors' robustness against that attack. 
Furthermore, unlike the observations reported in~\cite{dyrmishi2023empirical}, they found that using unrealistic attacks in AT can provide appropriate adversarial robustness; however, it significantly degrades clean performance. In addition to these two studies,~\cite{bostani2022domain,doan2023feature,huang2018visual} specifically explored robust optimization for AT. Similar to~\cite{dyrmishi2023empirical}, Bostani et al.~\cite{bostani2022domain} studied the impact of domain constraints on AT. However, they explored realizable AEs generated in the feature space to overcome the limitations of utilizing problem-space AEs in AT. Doan et al.~\cite{doan2023feature}
% performed adversarial learning in the feature space rather than the problem space, similar to\cite{bostani2022domain}. They 
introduced a new adversarial learning objective based on Bayesian inference to capture the distribution of models, leading to improved robustness.\\ \indent %Huang et al.~\cite{huang2018visual} investigated adversarial and natural variants in a model's decision space by introducing methods to compare a model's loss behavior against adversarial variants.\\ \indent
% created during training and those from other sources. 
% They confirmed that the flatness of the loss landscape, associated with generalization in naturally trained models, also applies to adversarially hardened models.
Finally, the lack of extensive investigation into the impact of influential factors on AT, especially classifiers and feature representations is another noteworthy concern in relevant studies. Most of the studies~\cite{dyrmishi2023empirical,li2023pad,doan2023feature,lucas2023adversarial,wang2021advandmal,lucas2021malware,chen2021using,labaca2021realizable,xu2023ofei,zhang2021enhanced,wang2020mdea,li2020adversarial,li2021framework,khormali2019copycat,huang2019malware,huang2018visual,al2018adversarial,anderson2018learning,yang2017malware,grosse2017adversarial} have primarily focused on using AT to reinforce a single type of classifier, typically relying on DNNs for malware detection; however, it remains uncertain how effective their explored AT methods are in fortifying other types of classifiers. Furthermore, to the best of our knowledge, there has not been a thorough investigation into how different representations, such as high-dimensional and low-dimensional feature spaces, impact the effectiveness of AT in the context of malware detection.

\section{Discussion}
\vspace{-1em}
\label{sec:discussion}
% Our analysis demonstrates that multiple variables (e.g., classifiers, feature representations, and robust optimization settings) significantly influence the effectiveness of hardening a target ML-based malware detection through AT.
% For instance, the method employed to create AEs is also crucial, as our research indicates that for a model to exhibit robustness, there needs to be some overlap between the areas of the input space covered by the adversarial hardening approach and those exploited by an attacker. Our findings indicate that there is no perfect general formula that exists, but that each configuration must be evaluated individually and meticulously to ensure the maximum effectiveness of the hardening process. This careful evaluation is crucial to avoid configurations that could potentially harm the learning process. For instance, inappropriate choices in feature representation or learning algorithms may lead to suboptimal robustness and a drop in clean performance. Furthermore, it is essential to tailor the AT approach to the specific characteristics of datasets and the threat models. By doing so, one can enhance the model's robustness without compromising its overall performance. This highlights the importance of comprehensive and context-specific evaluations in developing AT strategies.

Using our framework reveals that multiple factors jointly influence the effectiveness of hardening malware classifiers through AT. There is no universal recipe for achieving robustness---careful evaluation is essential to avoid configurations that may inadvertently hinder learning. For instance, inappropriate choices of learning algorithms can result in suboptimal robustness and degraded clean performance. Therefore, tailoring AT configurations to the characteristics of the data, feature representations, and classifiers is critical for developing balanced and reliable defenses. Rubik also enables the identification of the following questionable evaluation assumptions that may lead to unreliable conclusions.

\noindent\textbf{Hypothesis~1 -- Robust accuracy is enough.}
\label{hypothesis1}
When comparing the impact of AT under different settings, robust accuracy is often used~\cite{doan2023feature,bostani2022domain} to measure robustness improvement. However, our empirical analysis indicates that apparent improvements can be misleading when the corresponding vanilla models have different baseline levels of robustness.

\noindent\textbf{Recommendation.}
To fairly compare the impact of AT across models trained under different settings (e.g., a DNN trained on DREBIN versus RAMDA), the relative robustness defined in eq.~\ref{eq:relative_robustness} can be used. This metric accounts for each model's available robustness margin and provides a more reliable basis for comparison.

\noindent\textbf{Hypothesis~2 -- PK attacks represent the worst case.}
PK evasion attacks might be assumed to represent the worst-case scenario for uncovering adversarial vulnerabilities. However, unlike in computer vision, where ZK settings are rarely justified~\cite{carlini2019evaluating}, in the malware domain, a PK attack may not be the worst case, since its effectiveness also depends on problem-space transformations. For example, as shown in Figure~\ref{fig:real_robust_acc_ae_rate}-a and \ref{fig:real_robust_acc_ae_rate}-b, in most cases, DNN models hardened with PGD exhibit lower robustness against EvadeDroid than against PK-Greedy, indicating that EvadeDroid---despite operating under a ZK setting---surpasses PK-Greedy, which assumes PK access.

\noindent\textbf{Recommendation.} Adversarial robustness should be evaluated under multiple threat models, including both PK and ZK attacks, since robustness to one does not necessarily imply robustness to the other, as each may exploit different vulnerabilities arising from distinct problem-space transformations.
% \\ \indent

\noindent\textbf{Hypothesis~3 - Repeating experiments ensures reliable robustness.}
A common assumption in evaluating robustness is that repeating experiments and reporting aggregated metrics isolates the effect of factors such as AT perturbation bounds. While this reduces stochastic variability, it is not scalable for large-scale evaluations and can still be influenced by randomness. For example, our preliminary evaluation shows that the F1 score of the same model trained with identical bounds can vary between runs.

\noindent\textbf{Recommendation.}
For large-scale studies, a practical solution is to fix random seeds for all stochastic operations, ensuring deterministic training under identical hardware and software and providing a reproducible way to isolate the factor's effect.

\noindent\textbf{Hypothesis~4 - Functionality preservation guarantees realizable adversarial malware.}
It is often assumed that problem-space attacks preserving malicious functionality are sufficient to evaluate malware classifier robustness~\cite{demetrio2021functionality, demontis2017yes}. However, such attacks may still be unrealistic if they violate other constraints, such as robustness to preprocessing. For example, the attack proposed in~\cite{grosse2017adversarial} fails to show true robustness improvement, as defenders can filter manipulated samples before classification.

\noindent\textbf{Recommendation.} When evaluating adversarial robustness, it is necessary to consider evasion attacks that are realistic by satisfying entire domain constraints in the problem space~\cite{Pierazzi2020Intriguing}. 

% \hb{By emphasizing computational constraints, we encourage future research to build on this framework, exploring targeted adversarial settings to refine the balance between clean accuracy and robustness in malware detection.}
\vspace{-0.7em}
\subsection{Limitations} 
\vspace{-0.4em}
% Our research has focused on three classifiers that we consider representative of broader categories of learning algorithms: DNN, linear SVM, and DT. While these models are illustrative and can represent both linear and non-linear classifiers, it is crucial to note that the target learning algorithm significantly influences the effectiveness of AT. 
% % Due to the exponential complexity, we could not explore all possible algorithms.
% Moreover, our study has focused on discrete representation, which is predominant in the malware domain. However, altering the feature space would drastically influence attackers' capabilities, thereby impacting the AT process. This underscores the need to consider various feature representations to further explore the adversarial robustness of models against adversarial attacks. Expanding on this, different learning algorithms and feature representations require tailored AT approaches. For instance, while linear SVMs are sensitive to perturbations, DNNs can adapt to perturbations. DTs, on the other hand, are influenced by the choice of features and the depth of the tree, making their adversarial hardening unique. 

Our research focused on classifiers suitable for exploring a range of models, from deep non-linear to linear models. However, further exploration is needed, as the target learning algorithms significantly impact the effectiveness of AT. Moreover, we concentrated on discrete feature representations, common in malware classification, but since altering the feature representations could greatly affect attacker capabilities and the AT process, exploring more diverse feature representations is crucial for a deeper understanding of clean performance and adversarial robustness. Expanding the exploration of AT for malware classification by adding more classifiers and feature representations can offer deeper insights, but is impractical within a single study due to its complexity. Adding just one classifier or feature representation can increase the number of models in our study from $\approx1\text{K}$ to $\approx1.5\text{K}$, with training and evaluation potentially taking over a year. This highlights the trade-off between comprehensiveness and feasibility in large-scale AT experiments.
% \hb{Expanding the scope of AT for malware classification by incorporating additional feature representations and classifiers could provide deeper insights, but the sheer explosion in complexity makes such an endeavor impractical for a single study. Even a single added feature representation or classifier would significantly increase the number of models to train--scaling from ~1K to ~1.5K--leading to a staggering increase in computational demands. The training process alone could stretch beyond a year, not even accounting for the extensive evaluation required afterward. This highlights the trade-off between comprehensiveness and feasibility when designing large-scale AT experiments.}
\vspace{-0.7em}
\subsection{Future work}
\vspace{-0.7em}
% In line with the outlined limitations, future work could focus on exploring new feature representations, such as graph-based or continuous feature spaces, to comprehensively understand their impact. Different feature representations can significantly affect attackers since they alter the set of possible problem-space transformations that indirectly impact the adversarial training process. Moreover, investigating additional learning algorithms can further enhance the depth of the research, providing deeper insights into the strengths and weaknesses of each approach. Finally, while we have considered two realistic problem-space attacks, a 
% % white-box 
% PK gradient-based attack, and a 
% % black-box 
% ZK decision-based attack, future research can explore more attack strategies and threat models, such as transfer-based evasion attacks. This includes comparing their effectiveness and transferability with other techniques, both for attacking and defending. Such comparisons can highlight potential vulnerabilities and guide the development of more resilient models.
We underscore the real-world computational demands of large-scale AT to inspire future advancements. Our framework offers researchers a platform to explore customized adversarial settings, optimizing the balance between clean accuracy and robustness in malware detection. Given the limitations of prior studies, which often draw conclusions from narrow and varied settings, we encourage researchers to build on our framework to further explore the key factors highlighted in our study, especially for platforms other than Android, such as Windows. Such investigations can help identify broad trends and establish clear benchmarks across diverse experimental, operational, and threat model contexts.
% \hb{By emphasizing the real-world computational challenges of large-scale adversarial training, we aim to inspire future research to build upon our framework. Each researcher can explore new adversarial settings tailored to their interests, refining the delicate balance between clean accuracy and robustness in malware detection.}

% \subsection{\hb{Recommendations}}
% \hb{Based on our analysis and key findings, we offer the following recommendations to assist practitioners in deploying more effective malware classifiers:}

% \begin{itemize}
%     \item \hb{\textbf{Begin with feature selection.} When preparing the training set, reduce feature space dimensions as much as possible without impacting the classifier's performance on clean data. This can help the model learn more robust features, ultimately improving adversarial robustness.}
% \end{itemize}

% Understanding the impact of these variables is crucial for developing robust adversarial training methods. For instance, different feature representations can significantly alter the set of possible transformations and affect the adversarial training process. Similarly, experimenting with various learning algorithms can provide deeper insights into the strengths and weaknesses of each approach.

% Moreover, expanding the scope of attack strategies and threat models will provide a more comprehensive evaluation of the adversarial robustness. Comparing the effectiveness and transferability of different attack techniques can highlight potential vulnerabilities and guide the development of more resilient models.
\section{Conclusion}
\vspace{-0.7em}
We present Rubik, a unified framework that delineates the principal dimensions of AT for malware detection and their associated factors. Through a systematic evaluation of Rubik for Android malware, we explore the interplay of these factors. This evaluation challenges conclusions from prior work, provides new insights, rectifies misconceptions, and proposes best practices for enhancing AT. Our results demonstrate that a complex interdependence of characteristics dictates AT's success, underscoring the necessity of a tailored approach for building robust models.

\vspace{-0.7em}
\section*{Ethical Considerations}
\vspace{-0.7em}
This study did not involve human subjects and was conducted in full compliance with applicable laws and institutional ethical standards.

Nevertheless, we provide some considerations on risks and mitigations we put in place. This research investigates adversarial training to advance understanding of the robustness of malware detection systems. Since our methodology involves generating adversarial malware variants for training and evaluation, it is conducted strictly within a defensive security context. However, as adversarial techniques could in principle be misused, we have carefully assessed the associated ethical implications. To this end, we follow the principles of the Menlo Report~\cite{kenneally2012menlo}---beneficence, justice, respect for persons, and respect for law and public interest---along with relevant community guidelines. To mitigate potential risks, we implemented the following safeguards:
\begin{itemize}
\item All experiments were performed in isolated virtual machine environments that did not execute any malicious code nor generated any malicious traffic.
\item Only publicly-released research datasets were utilized for this study.
\item The released artifacts operates exclusively on feature-level abstractions, and cannot produce functional malware.
\end{itemize}

\section*{LLM Usage Considerations}
\vspace{-0.7em}
LLMs were used exclusively for editorial purposes in this manuscript, specifically to rephrase certain paragraphs and proofread.

\bibliographystyle{unsrt}
\bibliography{ref}

% \appendix
\appendices
% \renewcommand{\thefigure}{\Alph{section}.\arabic{figure}}
% \renewcommand{\thetable}{\Alph{section}.\arabic{table}}
% % Force appendix sections to use letters instead of numbers
% \renewcommand{\thesection}{\Alph{section}}
% % \section{Background}
% \label{app:background}
% \input{_NDSS25_submission/03-background}
% % \renewcommand{\thesection}{B}
% \renewcommand{\thesubsection}{B.\arabic{subsection}}
\appsection{Experimental Settings}
\label{app:settings}
\vspace{-0.5em}
\subsection{Data}
\vspace{-1em}
To empirically examine how the \textit{distribution} and \textit{volume} of data influence AT's efficacy, we employ two datasets: DREBIN20~\cite{Pierazzi2020Intriguing} and APIGraph~\cite{zhang2020enhancing}, which differ in size and labeling criteria.
DREBIN20 comprises $\approx150K$ Android applications (135,708 \emph{goodware} and 15,765 \emph{malware}) from 2016--2018, labeled using VirusTotal (VT) thresholds: zero detections for goodware, four or more for malware~\cite{Miller:Reviewer}.
APIGraph contains $\approx323K$ applications (290,505 goodware and 32,089 malware) from 2012--2018, using VT thresholds of 0 and 15, respectively~\cite{xu2019droidevolver}.
Both datasets follow Tesseract's methodology~\cite{Tesseract}, maintaining a realistic $\approx10\%$ malware ratio and chronological ordering to preserve temporal and spatial consistency.
We apply stratified sampling (67\% training, 33\% testing) and reserve 10\% of the training data for validation.
To ensure a fair evaluation and minimize distribution shift, we begin with vanilla models that perform well on clean data and design the datasets to mitigate temporal and spatial biases, thereby reducing concept drift before applying AT.

\vspace{-0.5em}
\subsection{Feature representations}
% In our study, we focus on two primary data representations to examine the role of \textit{dimensionality} and \textit{sparsity}: DREBIN~\cite{Arp:Drebin} (sparse, high-dimensional) and RAMDA~\cite{li2021robust} (dense, low-dimensional).
% These were chosen for their contrasting characteristics, which make them suitable for analyzing how representation affects security hardening.
% The DREBIN representation~\cite{Arp:Drebin} is defined by a high-dimensional, sparse feature space with about 1.5M features, offering a comprehensive yet complex view of data that captures nuanced behavioral patterns.
% Given this high dimensionality, we select the top $10K$ most distinctive features, following prior recommendations~\cite{demontis2017yes}.
% In contrast, RAMDA~\cite{li2021robust} employs a dense, low-dimensional space of 379 features.
% This compact form simplifies modeling by reducing complexity and improving dataset manageability. The choice of these representations is crucial because, as noted in~\cite{zhang2019limitations}, the data distribution strongly influences AT.
% Lower-dimensional spaces, such as RAMDA's, can enhance the fidelity of AT samples to the true data distribution.
\vspace{-1em}
In our study, we focus on two primary discrete data representations to examine the role of \textit{dimensionality} and \textit{sparsity}: DREBIN~\cite{Arp:Drebin} (sparse, high-dimensional) and RAMDA~\cite{li2021robust} (dense, low-dimensional).
Their contrasting characteristics make them ideal for analyzing how representation affects the hardening process.
The DREBIN representation~\cite{Arp:Drebin}---a high-dimensional, sparse space of ~1.5M features--- offers a comprehensive yet complex view that captures nuanced behaviors.
Given this high dimensionality, we select the top $10K$ most distinctive features, following prior recommendations~\cite{demontis2017yes}.
In contrast, RAMDA~\cite{li2021robust} employs a dense, low-dimensional space of 379 features.
This compact form simplifies modeling by reducing complexity and improving manageability. The choice of these representations is crucial because, as noted in~\cite{zhang2019limitations}, the data distribution strongly influences AT.
Lower-dimensional spaces, such as RAMDA's, may enhance the fidelity of AT samples to the true data distribution.
\vspace{-0.5em}
\subsection{Classifiers}
\vspace{-1em}
We employ three models---DNN (deep non-linear), DT (shallow non-linear), and linear SVM---to investigate how different learning algorithms are influenced by AT. Their range from non-linear to linear makes them well-suited for assessing the \textit{model flexibility} factor in our framework.

\vspace{-0.5em}
\subsection{Threat Model}
\vspace{-1em}
In order to conduct a thorough investigation, we consider a comprehensive threat modeling scenario that examines both defender and attacker capabilities. As an attacker, our goal is to fool the classifier and evade it successfully, while as a defender, we aim to make the model as robust as possible against multiple threat actors. Our analysis focuses on four key factors:

% \noindent\textbf{(a) Adversarial Confidence}: This factor captures the model's confidence in classifying an input as malicious or benign, playing a critical role in both attack success and model robustness. Attackers strive to craft adversarial malware that bypasses detection with high confidence, whereas defenders seek to lower this confidence and mitigate related threats.

\noindent\textbf{(a) Adversarial Confidence}: This factor measures the model's confidence in its classification, which is important to both attack and defense. Attackers aim to create malware that evades detection with high confidence, while defenders work to lower this confidence to mitigate threats.

% \noindent \textbf{(b) Perturbation Bound}: This defines the maximum number of feature changes allowed for a target sample. For example, given a binary vector and a perturbation bound of 5, it means that, to craft an AE, the modification limit is set to 5 features. This parameter is essential for both attackers and defenders as it directly affects sample confidence and feasibility. From the attacker's perspective, a lower bound limits modifiable features and may reduce evasion success, whereas higher bounds increase evasion likelihood but also the risk of detection or functionality alteration through behavioral analysis or manual inspection.
\noindent \textbf{(b) Perturbation Bound}: This factor defines the maximum number of feature changes allowed for a target sample. For example, given a binary vector and a perturbation bound of 5, the modification limit for crafting an AE is 5 features. This parameter is essential for both attackers and defenders as it directly affects confidence and feasibility. From the attacker's perspective, a lower bound limits modifiable features and may reduce evasion success, whereas a higher bound increases not only the evasion likelihood but also the risk of detection or functionality loss.

% \noindent \textbf{(c) Adversarial Fraction}: This factor denotes the percentage of AEs used during training. For instance, an adversarial fraction of 5\% means only 5\% of malware samples per batch are used for AT. While this parameter is primarily determined by the defender, it indirectly reflects the attacker's influence. A lower adversarial fraction provides less exposure to adversarial behavior, leaving the model potentially more vulnerable. From the attacker's perspective, this parameter impacts success rates, as models trained with limited AE exposure may generalize poorly to unseen attacks.
\noindent \textbf{(c) Adversarial Fraction}: This factor denotes the percentage of AEs used during training. For instance, a 5\% fraction means 5\% of malware samples per batch are used for AT. While primarily determined by the defender, it can indirectly reflect the attacker's influence. A lower fraction provides less exposure to adversarial behavior, potentially leaving the model more vulnerable. From the attacker's perspective, this impacts success rates, as models trained with limited AE exposure may generalize poorly to unseen attacks.

\noindent \textbf{(d) Domain Constraints}: In practice, attackers must ensure modifications preserve malicious functionality, are robust to removal, and remain plausible. These constraints substantially limit the space of allowable perturbations, especially for realistic evasion attacks. Defenders may enforce or relax such constraints during training, producing strategies that may or may not reflect real attack behavior; from the defender's perspective, realistic evasion is therefore confined to feasible regions shaped by the domain constraints they aim to capture or approximate.

By framing threat-model capabilities via adversarial confidence, perturbation bounds, adversarial fraction, and domain constraints, we can comprehensively assess attacker-defender dynamics. This framing clarifies how attackers optimize strategies within constraints and how defenders can adapt training to improve resilience. When generating AEs to either strengthen or deceive malware classifiers, the following factors are also considered for evasion attacks:

\noindent \textbf{Attacker Knowledge Assumptions}: This paper considers both Perfect Knowledge (PK) and Zero Knowledge (ZK) threat models. In the PK setting, the attacker has full access to the target model's architecture, parameters, and feature space. Conversely, the ZK setting permits only black-box query access to the model's output.

% \noindent \textbf{Attack Strategies}: In evaluating hardened malware classifiers, our primary focus is on realistic, problem-space evasion attacks. For PK scenarios, we utilize the PK-Greedy attack from Pierazzi et al.~\cite{Pierazzi2020Intriguing}, which also qualifies as an adaptive attack. PK-Greedy operates in white-box settings and adapts to the target model by identifying the most influential benign features the hardened classifier highlights. It dynamically adjusts to model characteristics and, via a greedy search, sorts candidate problem-space transformations and selects the most effective one each iteration, further reflecting its adaptive nature. For ZK scenarios, we employ EvadeDroid~\cite{bostani2024evadedroid}, a realistic, decision-based problem-space attack. Both PK-Greedy and EvadeDroid are designed to produce realizable AEs by enforcing domain constraints, ensuring generated malware remains valid and functional, making them suitable for real-world adversarial evaluation. In some experiments, PK-Greedy and EvadeDroid are also applied in AT to strengthen detectors. Additionally, we include PGD~\cite{madry2017towards} and JSMA~\cite{Papernot2016Limitations} in AT---both unrealistic feature-space attacks using gradients to directly generate AEs in the feature space.
\noindent \textbf{Attack Strategies}: We evaluate hardened malware classifiers using realistic, problem-space evasion attacks. For PK settings, we use the adaptive PK-Greedy attack~\cite{Pierazzi2020Intriguing}, which identifies the most influential benign features and uses a greedy search to select effective, realizable transformations. For ZK settings, we employ EvadeDroid~\cite{bostani2024evadedroid}, which is a decision-based attack. Both attacks enforce domain constraints to ensure the generated AEs are realizable, making them suitable for real-world evaluation. We also use these attacks for AT to strengthen detectors. Additionally, we include the unrealistic feature-space attacks PGD~\cite{madry2017towards} and JSMA~\cite{Papernot2016Limitations} in our AT, which use gradients to generate AEs directly in the feature space.
\vspace{-0.5em}
\subsection{Computational Resources}
\vspace{-1em}
\label{app_sub:computational_resources}
All our experiments were conducted on a dedicated instance equipped with an NVIDIA A100 80GB GPU, a 32-core AMD EPYC Milan processor @2.6 GHz, 128GB of RAM, and a 2.5TB SSD.
\vspace{-0.7em}
\appsection{Implementation Details}
\label{app:implementation_details}
\vspace{-0.7em}
Our study used learning algorithms implemented in PyTorch: linear SVM, DT, and DNN. To fully control the training process, we approximated the linear SVM as a single-layer neural network using PyTorch~\cite{paszke2017automatic}. We evaluated the SVM model with scikit-learn's \texttt{LinearSVC} class~\cite{scikit-learn}, which uses the LIBLINEAR library~\cite{fan2008liblinear}, to validate our implementation. For the DREBIN representation, we set the hyperparameter $C=1$~\cite{Arp:Drebin}, while preliminary evaluations suggested $C=4$ for better performance on RAMDA. For DT, we implemented the approach from~\cite{frosst2017distilling} and tuned it with Optuna~\cite{akiba2019optuna}, adjusting the following hyperparameters:
\begin{verbatim}
max_depth': 5,
'output_dim': 2,
'momentum': 0.53,
'lmbda':0.47,
'learning_rate': 0.12,
'weight_decay':  5e-4,
\end{verbatim}
Our implementation of the DNN is based on the Multilayer Perceptron described in~\cite{grosse2017adversarial} for malware detection.

\begin{figure}[t!]
    \centering
    \includegraphics[width=0.5\columnwidth]{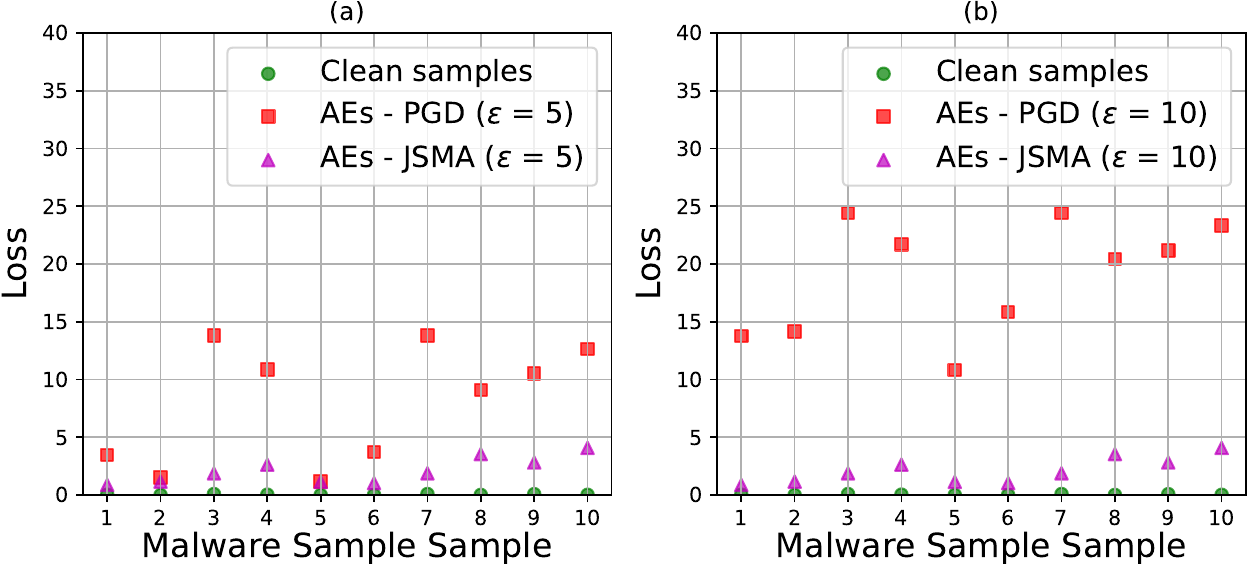}
    \caption{
    % Confidence.
   An example demonstrating the confidence level of different clean malware and adversarial malware samples in terms of loss. A larger loss indicates greater confidence in misclassification.}
    %\vspace{-1.5em}
    \label{fig:conf_DNN_without}
\end{figure}

Moreover, we implemented PGD and JSMA, with the former adapted for the Android malware domain according to~\cite{li2021framework}, and the latter directly adapted for use in our framework. Lastly, for PK-Greedy and EvadeDroid, we utilized the codes shared in their respective studies--\cite{Pierazzi2020Intriguing} and\cite{bostani2024evadedroid}, 
% ~\cite{bostani2022domain}, 
respectively.
\vspace{-0.7em}
\appsection{Confidence of AEs}
\label{app:confidence_of_AEs}
% To ascertain whether the feature-space attacks can generate AEs with varying levels of confidence, we conduct a preliminary evaluation of their performance against a vanilla DNN trained on the DREBIN representation. Drawing inspiration from \cite{pintor2022indicators}, we evaluated the loss (i.e., prediction error) of the vanilla DNN when classifying AEs generated by PGD or JSMA targeting the vanilla DNN. It is important to note that a higher loss corresponds to a higher evasion confidence. Figure~\ref{fig:conf_DNN_without} illustrates the confidence levels of 10 randomly selected malware samples from the DREBIN test set, comprising both clean and adversarial examples. It is evident that AEs inherently exhibit higher confidence than clean samples, with those generated by PGD showing higher confidence than those generated by JSMA. Furthermore, as $\epsilon$ increases, the confidence of PGD-generated AEs rises, whereas it remains relatively stable or changes only slightly for JSMA. It is noted that our empirical evaluation with multiple subsets aligns with the results in Figure~\ref{fig:conf_DNN_without}.
\vspace{-1em}
To determine if feature-space attacks generate AEs with varying confidence levels, we first evaluated their performance against a vanilla DNN trained on the DREBIN representation. Inspired by \cite{pintor2022indicators}, we assessed the prediction error of the DNN when classifying AEs from PGD and JSMA attacks. A higher loss signifies greater evasion confidence. Figure~\ref{fig:conf_DNN_without} shows the confidence levels for 10 random malware samples from the DREBIN20 test set, including both clean and adversarial examples. The results show that AEs inherently exhibit higher confidence than clean samples, with PGD-generated AEs being more confident than those from JSMA. Furthermore, as $\epsilon$ increases, confidence rises for PGD AEs but remains stable for JSMA. Our evaluation with multiple subsets confirms these findings.\\ \indent

\begin{figure}[t]
    \centering  
      
    \vspace{0.1em}
    \begin{subfigure}[b]{0.200\textwidth}
        \centering        \includegraphics[width=\textwidth]{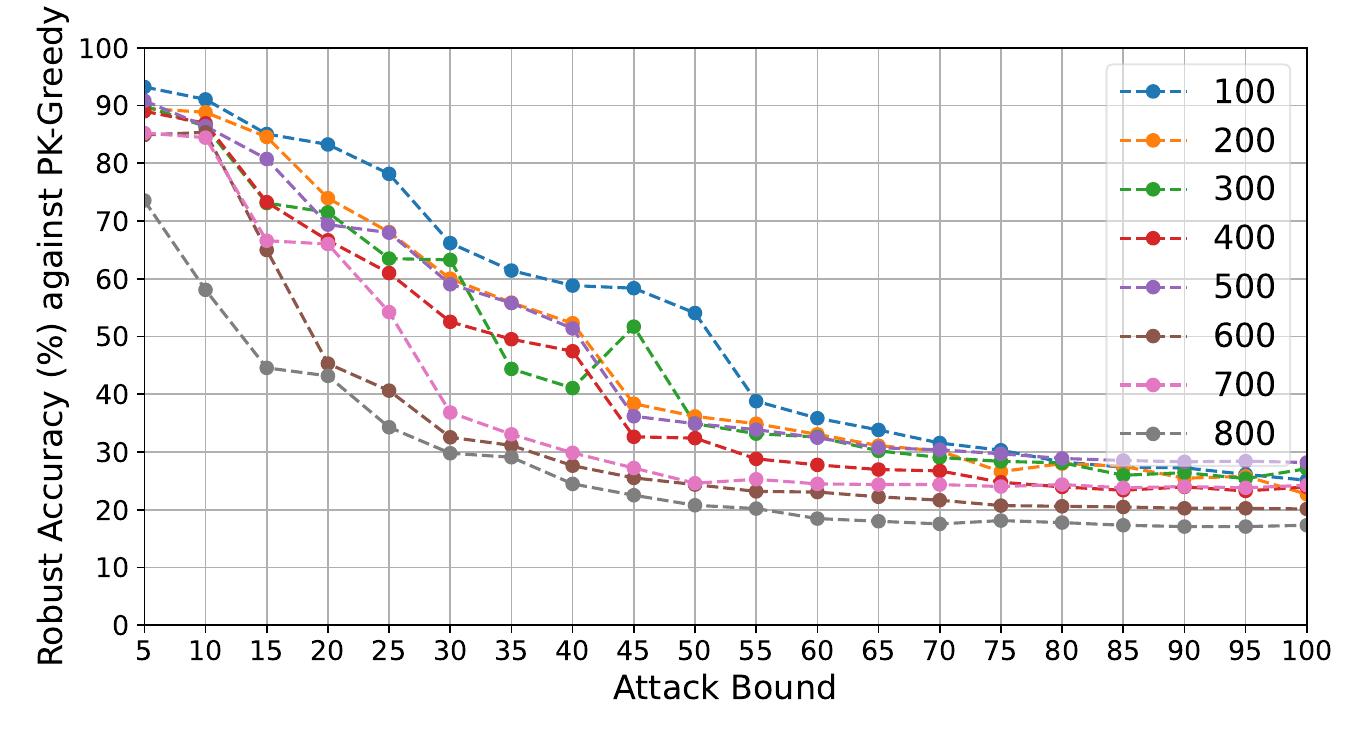} % Replace with your image file
        \vspace{-1.1em} % Adjust vertical space as needed
        \caption{DNN}
        \label{fig:image1}
    \end{subfigure}
    % \hfill
    \hspace{0.5em} 
    \begin{subfigure}[b]{0.200\textwidth}
        \centering
        \includegraphics[width=\textwidth]{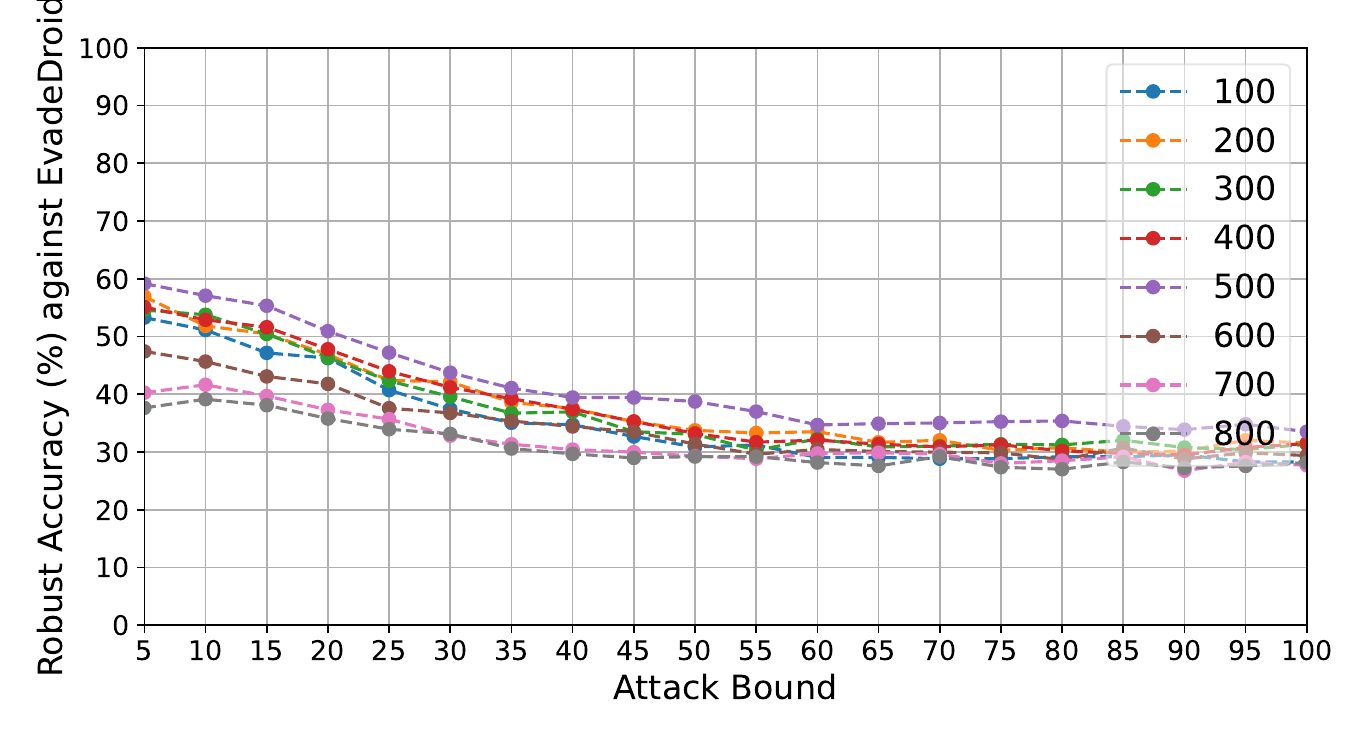} % Replace with your image file
        \vspace{-1.1em} % Adjust vertical space as needed
        \caption{DNN}
        \label{fig:image2}
    \end{subfigure} 

        \vspace{0.3em}
    \begin{subfigure}[b]{0.200\textwidth}
        \centering        \includegraphics[width=\textwidth]{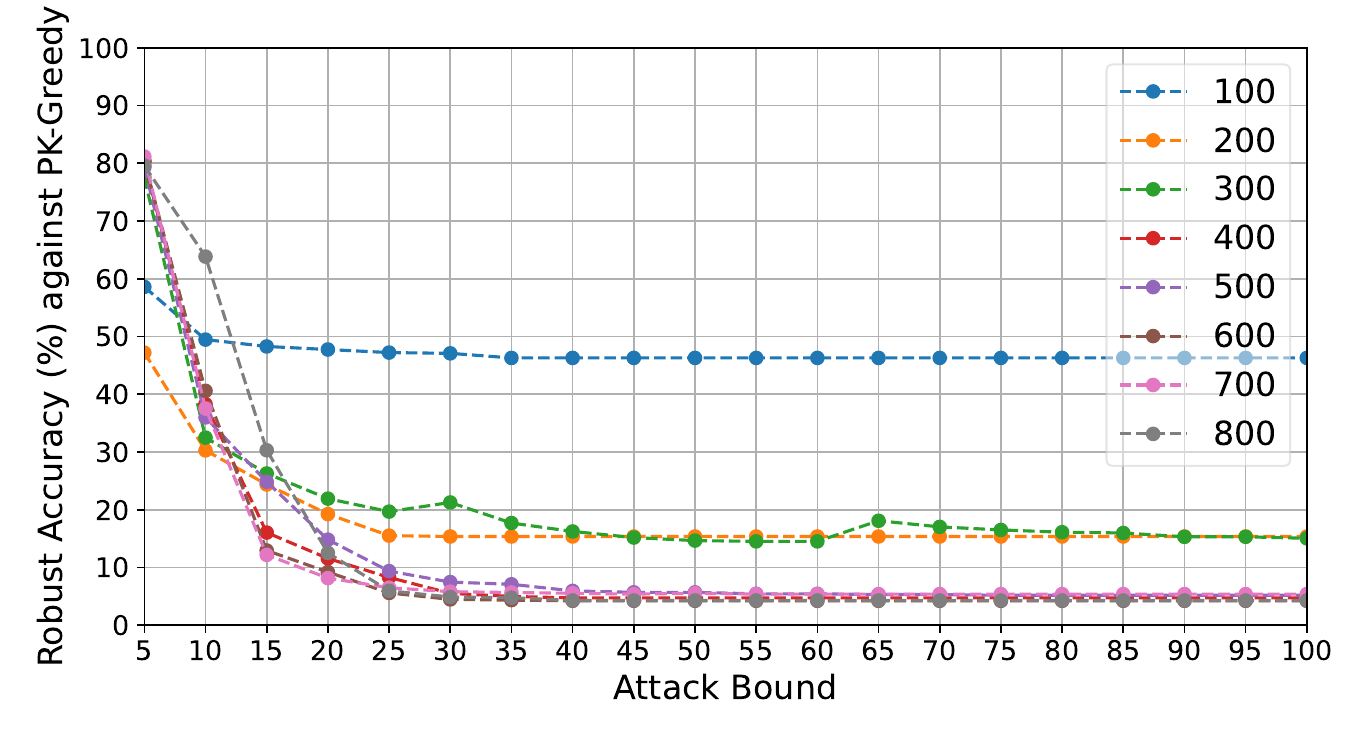} % Replace with your image file
        \vspace{-1.1em} % Adjust vertical space as needed
        \caption{SVM}
        \label{fig:image1}
    \end{subfigure}
    % \hfill
    \hspace{0.5em} 
    \begin{subfigure}[b]{0.200\textwidth}
        \centering
        \includegraphics[width=\textwidth]{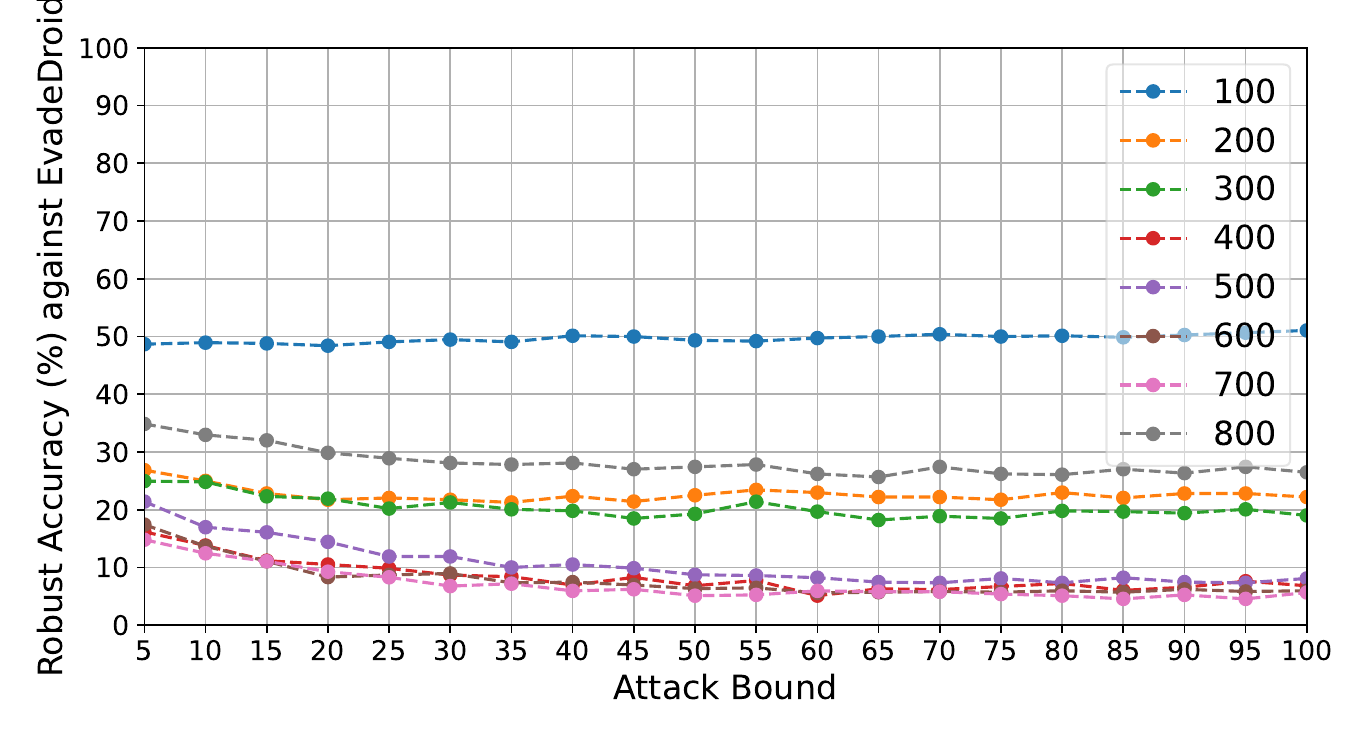} % Replace with your image file
        \vspace{-0.9em} % Adjust vertical space as needed
        \caption{SVM}
        \label{fig:image2}
    \end{subfigure}  
    % \vspace{0.3em}
    
    \begin{subfigure}[b]{0.200\textwidth}
        \centering
        \includegraphics[width=\textwidth]{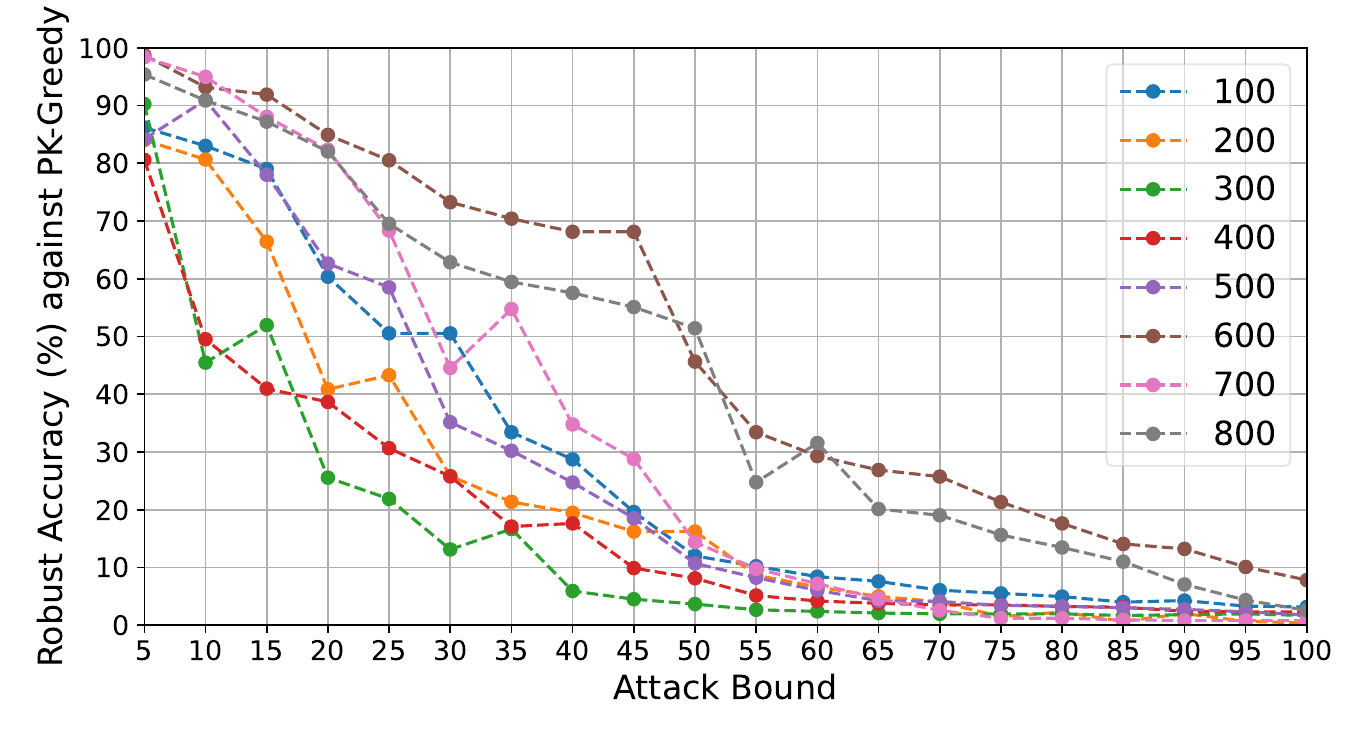} % Replace with your image file
        \vspace{-1.1em} % Adjust vertical space as needed
        \caption{DT}
        \label{fig:image1}
    \end{subfigure}
    % \hfill
    \hspace{0.5em} 
    \begin{subfigure}[b]{0.200\textwidth}
        \centering
        \includegraphics[width=\textwidth]{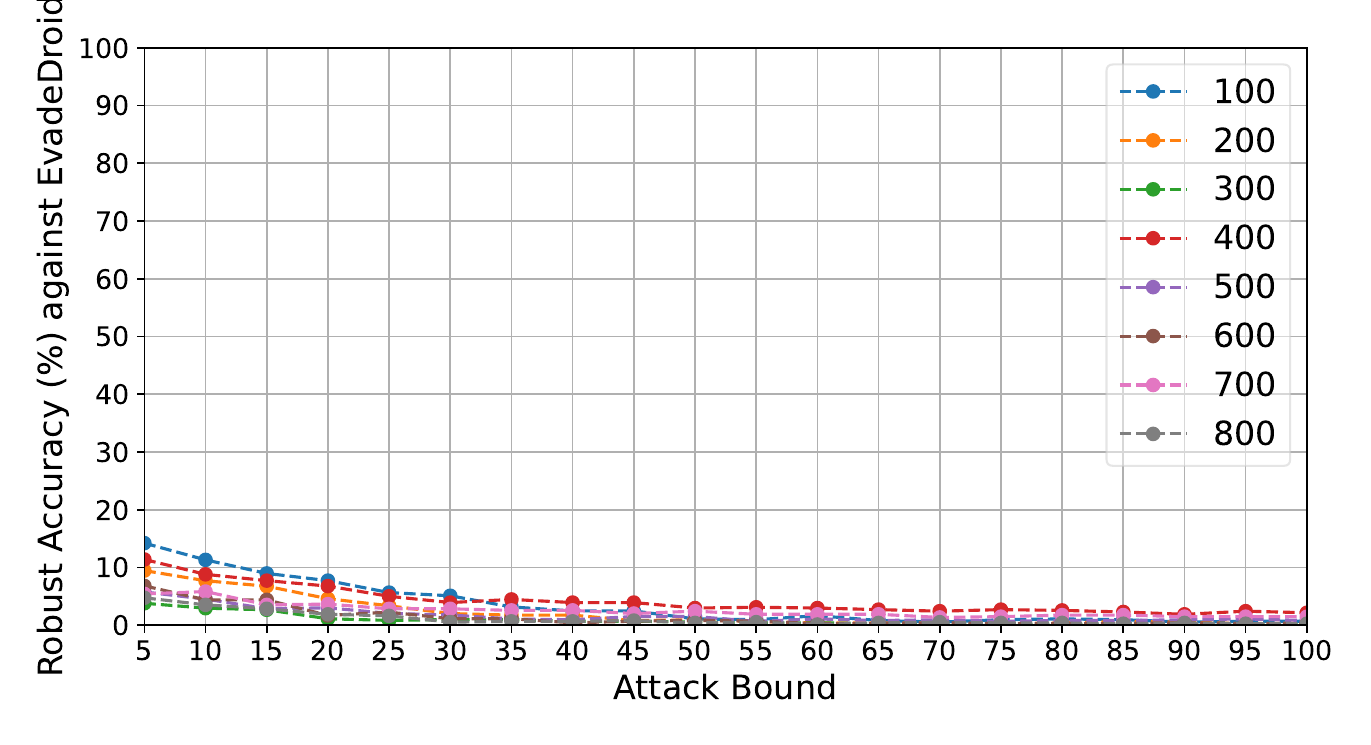} % Replace with your image file
        \vspace{-1.1em} % Adjust vertical space as needed
        \caption{DT}
        \label{fig:image2}
    \end{subfigure} 

    \caption{Robust accuracy of DNN, SVM, and DT trained on the DREBIN representation of the DREBIN20 dataset against PK-Greedy and EvadeDroid, with each line showing a model hardened with a specific perturbation bound.}
    \label{fig:big_perturbations_DNN_DT}
\end{figure}

\vspace{-2em}
\appsection{Large Perturbation Bound}
\label{app:big_perturbations}
\vspace{-1em}
This section extends the setup from \S\ref{subsubsec:perturbation_bound_and_the_confidence_of_aes} by increasing the defender's perturbation bound while keeping attacker capabilities fixed. Given DREBIN's high dimensionality, previous bounds may not reveal all vulnerabilities. We evaluate DNN, SVM, and DT models on DREBIN against PGD with bounds from 100 to 800 to see if larger perturbations expose more blind spots and improve AT. Figure~\ref{fig:big_perturbations_DNN_DT} highlights the learning algorithm's vital role, shown by the models' divergent performance against smaller perturbations. Linear SVMs gain no benefit from larger bounds against either attacker. DNN improves slightly against EvadeDroid, while DT gains against PK-Greedy, especially at lower bounds with PGD $\epsilon=600$. However, for large bounds ($\epsilon > 90$), results are consistent across models, indicating negligible or negative effects, as seen with linear SVM in Figure~\ref{fig:big_perturbations_DNN_DT}.

\appsection{Clean Performance Considering Different Adversarial Fractions}
\label{app:clean_acc}
\vspace{-0.7em}
This section includes the evaluation of experiments that focus on evaluating the clean performance of various models trained on DREBIN and RAMDA representations of both DREBIN20 and APIGraph datasets in terms of F1 score as shown in Figure~\ref{fig:clean_acc_ae_Rate}. The models are strengthened using either PGD or JSMA with different perturbation fractions of adversarial example rates.

% \appsection{Robust Performance Considering Different Adversarial Fractions}
% % AE rate}
% \label{app:at_rate}
% This section includes the evaluation of experiments that focus on exploring the 
% % robust accuracy
% robustness
% of various models
% trained on DREBIN and RAMDA representations of both DREBIN20 and APIGraph datasets in terms of relative robust accuracy as shown in Figure~\ref{fig:real_robust_acc_ae_rate}. The models are strengthened using either PGD or JSMA with different rates of AE during the training.

\vspace{-0.7em}
\appsection{Challenge of Exhaustive Exploration}
\label{app:exhaustive_exploration}
% Although we aimed to systematically evaluate the impact of various AT settings, conducting a comprehensive exploration across all experimental factors would require training a prohibitive number of models. Specifically, the total number of configurations is derived from the following combinations:
\vspace{-0.7em}
Although we aimed to evaluate AT settings systematically, an exhaustive exploration of all configurations would require a prohibitive number of models. Specifically, the total configurations are derived from:
\begin{itemize}
    \item \textbf{Datasets}: 2 (DREBIN20, APIGraph)
    \item \textbf{Feature Representations}: 2 (DREBIN, RAMDA)
    \item \textbf{Classifiers}: 3 (DNN, linear SVM, DT)
    \item \textbf{Perturbation Bounds}: 28 (20 for regular perturbations + 8 for very large perturbations)
    \item \textbf{Adversarial Fractions}: 10
    \item \textbf{Adversarial Confidence Levels}: 2 (low confidence (JSMA) and high confidence (PGD))
    \item \textbf{Domain Constraints}: 4 (PK-Greedy, EvadeDroid, PGD, JSMA)
\end{itemize}
% By systematically varying each factor independently, the total number of potential model configurations follows:
Systematically varying each factor yields:
\[
2 \times 2 \times 3 \times 28 \times 10 \times 2 \times 4 = 26,880
\]
% This calculation represents the full scope of model variations required for an exhaustive evaluation of AT across all dimensions. Given past training experience--where 1K+ models required over six months--evaluating 26,880 models would be computationally infeasible, requiring several years. Furthermore, expanding the current experimental setup by introducing additional variables into each designed experiment, such as considering various adversarial fractions in \S\ref{subsubsec:perturbation_bound_and_the_confidence_of_aes}, would significantly increase complexity. This would make the problem analogous to an NP-hard problem, rendering it intractable within practical time constraints.
Since training 1,000+ models previously took over six months, evaluating 26,880 would be computationally infeasible, requiring years. Adding variables, such as different adversarial fractions in \S\ref{subsubsec:perturbation_bound_and_the_confidence_of_aes}, would further increase complexity to an NP-hard scale, rendering the problem intractable.

\clearpage
\thispagestyle{empty}

\begin{landscape}

\appsection{Overview of Results}
\label{app:overview_of_results}
To simplify the review of the empirical results shown across various plots, we offer a high-level summary of the findings from our evaluations, as presented in Table~\ref{table:summary_of_key_findings}.

\noindent
\begin{center}
\footnotesize
\captionof{table}{Summary of Findings.}  % Use caption with "of" for non-floating
\label{table:summary_of_key_findings}
\tiny % Change the font size for the table

\begin{tabular}{l | l | l | l | l | l | l | l | l}
  \toprule
  \multirow{2}{*}{\textbf{Feature Space}} 
    & \multirow{2}{*}{\textbf{Attack used in AT}} 
    & \multirow{2}{*}{\textbf{\makecell{Attacker's\\ Knowledge}}}
    & \multirow{2}{*}{\textbf{Type of AEs in AT}}
    & \multirow{2}{*}{\textbf{Variable}}
    & \multicolumn{2}{c}{\textbf{Linear Model}} 
    & \multicolumn{2}{c}{\textbf{Non-linear Model}} \\  
  \cline{6-9}
    &   &   &   &  
    & \textbf{Clean Perf.}  & \textbf{Robust Acc.} 
    & \textbf{Clean Perf.}  & \textbf{Robust Acc.} \\  
  \midrule
  \multirow{3}{*}{\makecell[l]{Discrete high-dim.\\ sparse space}} 
    & \multirow{3}{*}{Feature-space, Unrealistic}  
    & \multirow{3}{*}{PK} 
    & \multirow{3}{*}{High-Conf. AEs} 
    & Increasing Pert. Bound 
    & - More loss     & - More gain 
    & - No impact (deep)        & - No trend \\
    &   &   &   &  
    &   &  
    & - Slight loss (shallow)        &   \\
  \cline{5-9} 
    &   &   &   & Very Large Bound 
    & Unexamined                    & No effect or worse 
    & Unexamined                    & No effect or worse \\
  \cline{5-9} 
    &   &   &   & Increasing AE Fraction 
    & - More loss      & - More gain 
    & - No impact (deep)        &  \\  
    &   &   &   &  
    &   &  
    & - Slight loss (shallow)        & - More gain \\
  \midrule
  \multirow{3}{*}{\makecell[l]{Discrete high-dim.\\ sparse space}} 
    & \multirow{3}{*}{Feature-space, Unrealistic}  
    & \multirow{3}{*}{PK} 
    & \multirow{3}{*}{Low-Conf. AEs} 
    & Increasing Pert. Bound 
    & - Slight loss               & - No trend 
    & - No impact (deep)     & - No trend \\
    &   &   &   &  
    &                              &  
    & - Slight loss (shallow)     & \makecell[l]{- Higher robust acc. compared to\\ the high-conf. AEs setting} \\
  \cline{5-9} 
    &   &   &   & Very Large Bound 
    & Unexamined                  & Unexamined 
    & Unexamined                  & Unexamined \\
  \cline{5-9} 
    &   &   &   & Increasing AE Fraction 
    & Sligh loss               & \makecell[l]{- More gain, but lower than\\ the high-conf. AEs setting} 
    & - Slight loss               & - More gain \\
  \midrule
  \multirow{3}{*}{\makecell[l]{Discrete low-dim.\\ dense space}} 
    & \multirow{3}{*}{Feature-space, Unrealistic}  
    & \multirow{3}{*}{PK} 
    & \multirow{3}{*}{High-Conf. AEs} 
    & Increasing Pert. Bound 
    & -Varied loss      & - No trend 
    & - No impact (deep)        & - No trend \\   
    &   &   &   &  
    &                                   &  
    & - Varied loss (shallow).          &  \makecell[l]{- Higher robust acc. compared to\\ the high-dim. space setting (shallow)} \\
  \cline{5-9} 
    &   &   &   & Very Large Bound 
    & Unexamined                      & Unexamined 
    & Unexamined                      & Unexamined \\ 
  \cline{5-9} 
    &   &   &   & Increasing AE Fraction 
    & - More loss                      & - More gain 
    & - No impact (deep)        & - More gain \\  
    &   &   &   &  
    &                                &  
    & - More loss (shallow)         &   \\
  \midrule
  \multirow{3}{*}{\makecell[l]{Discrete low-dim.\\ dense space}} 
    & \multirow{3}{*}{Feature-space, Unrealistic}  
    & \multirow{3}{*}{PK} 
    & \multirow{3}{*}{Low-Conf. AEs} 
    & Increasing Pert. Bound 
    & - More loss 
    & - More gain
    & - No impact (deep) 
    & - No trend \\  
    &   &   &   &  
    & \makecell[l]{- Higher clean perf. compared to\\ the high-conf. AEs setting}
    &  
    & - Varied loss (shallow) 
    & \makecell[l]{- Higher robust acc. compared to\\ the high-conf. AEs setting (deep)}  \\
  \cline{5-9} 
    &   &   &   & Very Large Bound 
    & Unexamined                  & Unexamined 
    & Unexamined                  & Unexamined \\ 
  \cline{5-9} 
    &   &   &   & Increasing AE Fraction 
    & - More loss 
    & - More gain 
    & - No impact (deep) 
    & - More gain\\ 
    &   &   &   &  
    & \makecell[l]{- Higher clean perf. compared to\\ the high-conf. AEs setting} & & - More loss (shallow) & \makecell[l]{- Higher robust acc. compared to\\ the high-conf. AEs setting}\\  
  \midrule
  \makecell[l]{Discrete high-dim.\\ sparse space}
    & Problem-space, Realistic 
    & PK 
    & Realizable AEs 
    & Forcing Domain Constraints 
    & \makecell[l]{- No impact}
    % vs.\\ significant loss in the\\ high-conf. AEs setting}
    & \makecell[l]{- Low robust acc. (only against\\ similar attack)}
    % \\ vs. moderate robust acc. in the high-conf.\\ AEs setting.}
    & \makecell[l]{- No impact}
    % like the\\ AEs setting}
    & \makecell[l]{- High robust acc. against\\ similar attack (deep)} \\ 
    &   &   &   &  
    &  
    &  
    &  
    & \makecell[l]{- Very low robust acc. (shallow)} \\
  \midrule
  \makecell[l]{Discrete high-dim.\\ sparse space}
    & Problem-space, Realistic 
    & ZK 
    & Realizable AEs 
    & Forcing Domain Constraints 
    & \makecell[l]{- No impact}
    % vs.\\ significant loss in the \\ high-conf. AEs setting}
    & \makecell[l]{- Low robust acc. (only against\\ similar attack)}
    % \\ vs. moderate robust acc. in the high-conf.\\ AEs setting.}
    & \makecell[l]{- No impact}
    % like\\ AEs setting}
    & \makecell[l]{- Low robust acc. against \\ similar attack, and very low\\ against dissimilar attack (deep)}\\
    % \\ vs. moderate robust acc. in the high-conf.\\ AEs setting (deep)} \\ 
    &   &   &   &  
    &  
    &  
    &  
    & \makecell[l]{- Very low robust acc. (shallow)} \\ 
    
    \midrule
  \makecell[l]{Discrete low-dim.\\ dense space}
    & Problem-space, Realistic 
    & PK 
    & Realizable AEs 
    & Forcing Domain Constraints 
    & \makecell[l]{- No impact}
    % vs.\\ significant loss in the\\ AEs setting}
    & \makecell[l]{- Moderate robust acc.}
    % vs. high robust\\ acc. in AEs setting.}
    & \makecell[l]{- No impact}
    % like\\ the AEs setting (deep)}
    & \makecell[l]{- High robust acc.\\ against similar attack (deep)} \\    
    &   &   &   &  
    &  
    &  
    &  
    & \makecell[l]{- High robust acc. (shallow)}
    % \\ like the robust acc. in\\ high-confidence AEs setting (shallow)}
    \\
    \midrule
  \makecell[l]{Discrete low-dim.\\ dense space}
    & Problem-space, Realistic 
    & ZK 
    & Realizable AEs 
    & Forcing Domain Constraints 
    & \makecell[l]{- No impact}
    % vs.\\ significant loss in the \\ AEs setting}
    & \makecell[l]{- Moderate robust acc.}
    % vs. high\\ robust acc. in nrealizable AEs setting.}
    & \makecell[l]{- No impact}
    % like\\ the AEs setting (deep)}
    & \makecell[l]{- Very high robust acc. (deep)} \\ 
    &   &   &   &  
    &  
    &  
    &  
    & \makecell[l]{- Moderate robust acc.\\ against similar attack\\ and high robust acc.\\ against dissimilar attack (shallow)}
    % lower than\\  high-confidence\\ AEs setting (shallow)}  \\
    \\
  \bottomrule
\end{tabular}

\end{center}

\end{landscape}

\clearpage

\end{document}